\documentclass[12pt]{article}
\usepackage{fullpage}
\usepackage[colorlinks=true,citecolor=blue]{hyperref}
\usepackage{url}
\usepackage{amsmath, amssymb}
\usepackage{amsfonts}
\usepackage{amsthm}
\usepackage{tikz}
\usepackage{CJKutf8}
\usepackage{graphicx}

\usetikzlibrary{shapes.geometric, arrows.meta}

\usepackage{listings}
\theoremstyle{definition}
\newtheorem{problem}{Problem}

\title{Winning Gold at IMO 2025 with a Model-Agnostic Verification-and-Refinement Pipeline\footnote{Code available at: \url{https://github.com/lyang36/IMO25}}}
\author{Yichen Huang (黄溢辰)\thanks{\href{mailto:huangtbcmh@gmail.com}{huangtbcmh@gmail.com}} \and Lin F. Yang (杨林)\thanks{\href{mailto:linyang@ee.ucla.edu}{linyang@ee.ucla.edu}, Department of Electrical and Computer Engineering, and Department of Computer Science, UCLA.}}
\date{\today}

\begin{document}
\begin{CJK}{UTF8}{gbsn}
\maketitle
\end{CJK}
\begin{abstract}
The International Mathematical Olympiad (IMO) is widely regarded as the world championship of high-school mathematics. IMO problems are renowned for their difficulty and novelty, demanding deep insight, creativity, and rigor. Although large language models perform well on many mathematical benchmarks, they often struggle with Olympiad-level problems. Using carefully designed prompts, we construct a model-agnostic, verification-and-refinement pipeline. We demonstrate its effectiveness on the recent IMO 2025, avoiding data contamination for models released before the competition. Equipped with any of the three leading models---Gemini 2.5 Pro, Grok-4, or GPT-5---our pipeline correctly solved 5 out of the 6 problems ($\approx85.7\%$ accuracy). This is in sharp contrast to their baseline accuracies: 31.6\% (Gemini 2.5 Pro), 21.4\% (Grok-4), and 38.1\% (GPT-5), obtained by selecting the best of 32 candidate solutions. The substantial improvement underscores that the path to advanced AI reasoning requires not only developing more powerful base models but also designing effective methodologies to harness their full potential for complex tasks.
\end{abstract}

\section{Introduction}

The International Mathematical Olympiad (IMO) is an annual competition widely regarded as the world championship of high-school mathematics. Established in Romania in 1959 with just seven participating countries, it has since expanded to include over 100 nations, each represented by a team of up to six of their most talented pre-university students. The competition is an extremely challenging test of creativity and sustained concentration: over two consecutive days, contestants are given two 4.5-hour sessions to solve three problems per session, drawn from the fields of algebra, combinatorics, geometry, and number theory~\cite{imo_official}.

Qualifying for the IMO is itself extremely challenging. In the United States, for instance, a student must advance through a rigorous series of national competitions of increasing difficulty, from the American Mathematics Competitions (AMC) to the American Invitational Mathematics Examination (AIME), and finally to the USA Mathematical Olympiad (USAMO). Top performers at USAMO are invited to compete for the six spots on the U.S. national team. Most other countries have similarly stringent selection processes, ensuring that the IMO convenes the world's most talented pre-university students. A gold medal at the IMO is an extraordinary achievement, awarded to only the top twelfth of the contestants. Consequently, the IMO serves as the preeminent stage where future leaders in mathematics demonstrate their exceptional talent, and success at the IMO has a significant correlation with the Fields Medal, the highest honor in mathematics. Of the 34 Fields medalists awarded since 1990, 11---including renowned mathematicians Terence Tao, Maryam Mirzakhani, and Grigori Perelman---are prior IMO gold medalists.\footnote{Compiled by the authors by checking the IMO records of all Fields medalists awarded 1990-2022. IMO data was sourced from the official website.} Furthermore, the probability that an IMO gold medalist will become a Fields medalist is 50 times larger than the corresponding probability for a PhD graduate from a top-10 mathematics program \cite{AG20}.

Advanced mathematical reasoning is a hallmark of intelligence and the foundation of science and technology. Consequently, automated mathematical reasoning has become a major frontier in artificial intelligence (AI). The rapid advancement of Large Language Models (LLMs) has enabled them to master mathematical benchmarks of increasing difficulty \cite{ahn2024large, wang2025survey}. This progress has been enabled by inference-time methods such as Chain-of-Thought, which improves performance on complex tasks by breaking them down into a sequence of intermediate reasoning steps~\cite{wei2022chain}. Early datasets GSM8K~\cite{cobbe2021gsm8k} and MATH~\cite{hendrycks2021math}, which test grade-school and high-school mathematics, respectively, have been largely solved. The performance of leading models, such as Gemini 2.5 Pro~\cite{gemini2025blog}, Grok-4, and GPT-5, is also approaching saturation on the AIME, a significantly more challenging competition benchmark. However, AIME problems are not required to be entirely novel: 8 out of the 30 problems in AIME 2025 were identified as having close analogs in online sources available prior to the event \cite{BDP+25}. This allows models to achieve high performance partly through sophisticated pattern recognition and adaptation of existing solutions rather than completely original reasoning.

The remarkable success of LLMs on these benchmarks has pushed the frontier of AI mathematical reasoning to the next tier: Olympiad-level problems~\cite{he2024olympiadbench}. This represents a shift not merely in difficulty, but in the very nature of the task. Whereas the AIME requires only a final numerical answer, the USAMO and IMO demand a complete and rigorous proof. In mathematics, an answer without a rigorous proof is merely a conjecture; it is the proof that promotes a conjecture to a theorem. Furthermore, IMO problems are systematically selected for novelty: the selection process is designed to filter out any candidate problem that is too similar to a known problem~\cite{IMOF, Kim23}. Thus, solving IMO problems requires original insights and multi-step creative reasoning, rather than pattern recognition and retrieval from training data. These three pillars---the renowned difficulty, the demand for rigor, and the strict criterion of problem novelty---establish the IMO as a grand challenge and the preeminent benchmark for assessing the genuine mathematical reasoning capability of LLMs. The demand for logically sound arguments, in particular, exposes a critical weakness in current LLMs~\cite{mahdavi2025brains}. Recent evaluations on the USAMO 2025~\cite{petrov2025proof} and IMO 2025~\cite{matharena2025} show that state-of-the-art models struggle to generate sound, rigorous proofs, often committing logical fallacies or using superficial heuristics, and consequently fail to win even a bronze medal.

Last year, Google DeepMind announced a breakthrough: an AI system that achieved a silver-medal performance at the IMO 2024 \cite{deepmind2024alphaproof}. Their approach used AlphaGeometry 2 \cite{trinh2024solving, CTO+15}, a specialized solver for geometry, and AlphaProof for algebra and number theory problems. Notably, AlphaProof generates proofs in the formal language Lean. The primary advantage of this formal approach is guaranteed correctness: a proof successfully verified by the Lean proof assistant is irrefutably sound. However, this guarantee comes at the cost of human readability. Proofs in formal languages are often verbose and cumbersome, and require specialized training to understand, making them inaccessible to most mathematicians. Our work, in contrast, is situated entirely within the natural language paradigm. Our approach produces human-readable proofs, akin to those in mathematical journals and textbooks. This is crucial for enabling effective human-AI collaboration, where mathematicians can understand, critique, and build upon an AI's reasoning. While generating machine-verifiable proofs is a vital goal, our natural language approach tackles the complementary challenge of creating an AI that can reason and communicate like a human mathematician, thereby making its insights easily accessible to the scientific community.

In this paper, we construct a model-agnostic verification-and-refinement pipeline and demonstrate its effectiveness across three leading large language models: Gemini 2.5 Pro, Grok-4, and GPT-5. When equipped with any of these models, our pipeline solved 5 out of the 6 problems from the IMO 2025, achieving an accuracy of approximately 85.7\%. This result stands in sharp contrast to the models' baseline performance. An independent evaluation \cite{matharena2025} by MathArena employed a best-of-32 post-selection strategy: for each problem, 32 solutions are generated, and the model itself selects the most promising one for human grading. Even with this performance-boosting inference-time method, the reported accuracies were only 31.6\% for Gemini 2.5 Pro, 21.4\% for Grok-4, and 38.1\% for GPT-5. A pervasive and fundamental challenge in the evaluation of LLMs is data contamination, where test problems are included in a model's training data, inflating its performance metrics~\cite{zhou2024dont}. Our use of the recent IMO 2025 problems helps mitigate this issue. Since Gemini 2.5 Pro and Grok-4 were released before the competition, they were evaluated on a pristine testbed. While GPT-5 was released after the competition, raising the possibility of data contamination, the comparison to the best-of-32 baseline remains fair, as any contamination would affect both their and our evaluations. The substantial performance gain---from a baseline of 38.1\% to our 85.7\%---isolates the contribution of our pipeline, confirming its effectiveness regardless of potential data contamination. Our results demonstrate that strong existing LLMs already possess powerful mathematical reasoning capabilities, but that a verification-and-refinement pipeline is essential for converting their latent capabilities into rigorous mathematical proofs.

To further validate the generalizability and robustness of our pipeline, its performance was independently assessed on a different and challenging benchmark: the 2025 International Mathematics Competition for University Students (IMC). The IMC is a prestigious annual contest that includes topics from undergraduate curricula. Thus, it requires a broader and more advanced mathematical knowledge base than the IMO. MathArena evaluated what they termed the ``Gemini agent''---our pipeline with Gemini 2.5 Pro as the base model. The agent achieved 94.5\% accuracy~\cite{imc}, ranked \#3 among 434 human participants. By contrast, the base model alone only scored 57.7\% and ranked \#92. This third-party validation demonstrates our pipeline's effectiveness in a more knowledge-intensive domain and on a pristine, uncontaminated dataset, as the competition occurred after the public release of our code.

Our work builds upon a growing body of research aimed at enhancing the reasoning capabilities of LLMs through verification and iterative refinement. Foundational works~\cite{KBM23, MTG+23} have pioneered the framework of this approach, where a model generates an output, receives feedback, and then refines its work. In the mathematical domain, this approach has been adapted into various methods designed to verify and improve the logical steps of a solution~\cite{WZX+23, PIP+24, WLS+24}. Another line of research focuses on generating and repairing proofs in formal languages to guarantee correctness~\cite{FRRB23}. While our pipeline is built on these core ideas of iterative refinement, our contribution is to construct a model-agnostic, inference-time framework with carefully designed prompts that specialize this process for the extreme rigor and novelty demanded by Olympiad mathematics. Our robust verifier design directly addresses the challenge of generating high-quality feedback, a known bottleneck for self-correction methods~\cite{KZZ+24}. By applying our pipeline to state-of-the-art LLMs, we demonstrate a level of performance on the IMO 2025---a grand-challenge benchmark of significantly greater difficulty than those addressed in prior studies---that was previously unattainable.

Other teams, including OpenAI~\cite{wei2025openai}, Google DeepMind~\cite{Deepmind}, and ByteDance~\cite{bytedance2025seed}, announced strong performance of their AI systems on the IMO 2025 problems after the event.

\section{Methods}

\subsection{Pipeline}

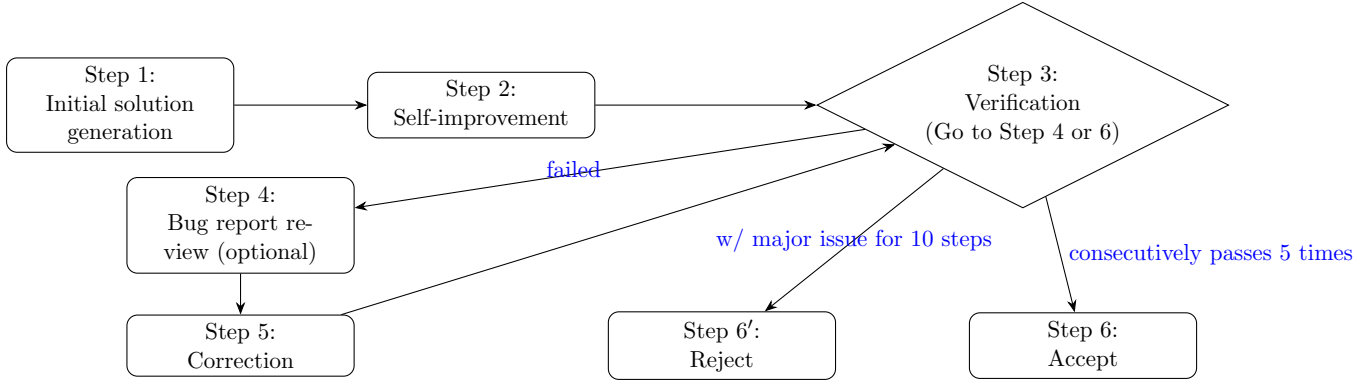
\begin{figure}[ht]
\centering
\begin{tikzpicture}[
  scale=0.8,
  transform shape,
  every node/.style={font=\small, align=center},
  process/.style={rectangle, draw, rounded corners, minimum width=3.5cm, minimum height=1cm, text width=3.5cm},
  decision/.style={diamond, draw, aspect=2, inner sep=1pt, minimum height=1.5cm, text width=4cm},
  ->, >=Stealth
  ]

\node[process] (step1) at (0,0) {Step 1:\\ Initial solution generation};
\node[process] (step2) at (6,0) {Step 2:\\ Self-improvement};
\node[decision] (step3) at (15, 0) {Step 3:\\ Verification\\ (Go to Step 4 or 6)};
\node[process] (step4) at (2, -2) {Step 4:\\ Bug report review (optional)};
\node[process] (step5) at (2, -4) {Step 5:\\ Correction};
\node[process] (step6) at (16, -4) {Step 6:\\ Accept};
\node[process] (step7) at (10,-4) {Step 6$'$:\\ Reject};

\draw[->] (step1) -- (step2);
\draw[->] (step2) -- (step3);
\draw[->] (step3) -- (step4) node[midway, left] {\textcolor{blue}{failed}};
\draw[->] (step3) -- (step6) node[midway, right] {\textcolor{blue}{consecutively passes 5 times}};
\draw[->] (step3) -- (step7) node[midway] {\textcolor{blue}{w/ major issue for 10 steps}};
\draw[->] (step4) -- (step5);
\draw[->] (step5) -- (step3);

\end{tikzpicture}
\caption{Flow diagram of our pipeline. See the main text for detailed explanations of each step.}
\label{fig:flow-diagram}
\end{figure}

At a high level, our pipeline proceeds as follows (illustrated in Figure \ref{fig:flow-diagram}):

\begin{itemize}
    \item Step 1: Initial solution generation with the prompt in Section~\ref{prompt:step1};
    \item Step 2: Self-improvement;
    \item Step 3: Verifying the solution with the prompt in Section~\ref{prompt:verification} and generating a bug report; go to Step 4 or Step 6 (see below for explanations);
    \item Step 4: Review of the bug report (optional);
    \item Step 5: Correcting or improving the solution based on the bug report; go to Step 3;
    \item Step 6: Accept or Reject.
\end{itemize}

We run the procedure some number of times (in parallel or in serial, independently) in order to obtain a correct solution. We hope that the model either outputs a correct solution or reports that it failed to find one.

\subsection{Solver}
The solver prompt in Section~\ref{prompt:step1} for Step 1 is designed to emphasize rigor rather than focus on finding the final answer and thus matches the theme of IMO. We have randomly selected some outputs of this step and found that the overall quality of the solutions is pretty low. This is consistent with very recent findings of Ref.~\cite{matharena2025}.

In Step 2, the model is prompted to review and try to improve its work. General-purpose LLMs are not tailored to solving exceptionally challenging mathematical problems in a single pass. A significant constraint is their finite reasoning budget allocated for a single query. This limitation is most transparently illustrated by Gemini 2.5 Pro, which allows for explicit control over its reasoning budget via the number of thinking tokens. Note that thinking is quite token consuming: Even a trivial fact might take a few thousand tokens for the model to prove. The maximum number of thinking tokens for Gemini 2.5 Pro is 32,768, which is not enough for solving a typical IMO problem. We observe that in Step 1, the model almost always uses up this budget. Consequently, it lacks the capacity to fully solve the problem in one go. While Grok-4 and GPT-5 offer no or less granular control over their reasoning effort---it cannot be adjusted for Grok-4 and can only be set categorically for GPT-5---they operate under similar, albeit less explicit, constraints. This is why we break down the problem-solving process into steps. Step 2 effectively injects another budget of reasoning tokens, allowing the model to review and continue its work. We keep monitoring the entire process and consistently observe that the outputs have been noticeably improved during Step 2.

Next we will use the verifier to make iterative improvement and decide whether to accept an improved solution.

\subsection{Verifier}

The verifier plays an important role in our pipeline. Its functionality is to carefully review a solution step by step and find out issues (if any). We emphasize mathematical rigor and classify issues into critical errors and justification gaps. Critical errors are something that is demonstratively false or with clear logical fallacies, while justification gaps can be major or minor. A major justification gap that cannot be repaired would crash an entire proof, while minor justification gaps may not even be well defined: A minor gap could sometimes be viewed as concise argument.

In Step 3, we use the verifier to generate a bug report for each solution outputted in Step 2. The bug report contains a list of issues classified as critical errors or justification gaps. For each issue, an explanation is required. The bug report will serve as useful information for the model to improve the solution, either fixing errors or filling gaps. Step 4 (optional) is to carefully review each issue in the bug report. If the verifier makes a mistake and reports an issue which is not really an issue, the issue would be deleted from the bug report. Thus, Step 4 increases the reliability of the bug report. In Step 5, the model tries to improve the solution based on the bug report. We iterate Steps 3-5 a sufficient number of times until we decide to accept or decline a solution. We accept a solution if it robustly passes the verification process and decline a solution if there are always critical errors or major justification gaps during the iterations.

We observe that the verifier is quite reliable but can make mistakes. Since our major goal is not to benchmark the verifier, we do not have quantitative results on its effectiveness. However, we have used this verifier for quite a while (starting from well before IMO 2025). We have been keeping an eye on its performance and below is our qualitative observation:
\begin{itemize}
    \item Critical errors are seldom missed by the verifier. This is consistent with the observations in Refs.~\cite{SLJ15,guo2025temporal}. In the unlikely event such errors are not caught, simply running the verifier a few more times would very likely catch it. This is good because we do not wish to miss critical errors.
    \item If the verifier reports a critical error, it may not always be critical, but it almost always needs some revision.
    \item The verifier may report some justification gaps which are only slightly beyond trivial statements and thus are not really gaps for mathematicians.
\end{itemize}
Indeed, our system is quite robust to errors made by the verifier. We iteratively use the verifier a sufficiently number of times. If it misses an error in one iteration, it still has some probability to catch it in the next iteration. Also, if it claims an error which is actually not an error, such a false negative may not go through the bug report review step (Step 4). Furthermore, we instruct the model (who generates the solution) to review each item in the bug report. If the model does not agree with a particular item, it is encouraged to revise its solution to minimize misunderstanding. This is analogues to the peer review process. If a referee makes a wrong judgment, the authors are encouraged to revise the paper. Ultimately, the presentation is improved.

At the time we plan to accept a solution, we do not wish the verifier to miss any issue; we run the verifier five times and accept a solution only if it passes every time.

\section{Experiment Setup}

For Gemini 2.5 Pro and Grok-4, we choose low temperature: $0.1$. For GPT-5, the temperature cannot be adjusted. For Gemini 2.5 Pro, we use the maximum thinking budget ($32768$ reasoning tokens). For Grok-4, the reasoning effort cannot be adjusted. For GPT-5, we use high reasoning effort. We disable web search (of course), code, and all other tools. We share the solver and verifier prompts below.

\subsection{Solver Prompt}
\label{prompt:step1}
\begin{lstlisting}
    ### Core Instructions ###

*   **Rigor is Paramount:** Your primary goal is to produce a complete and rigorously justified solution. Every step in your solution must be logically sound and clearly explained. A correct final answer derived from flawed or incomplete reasoning is considered a failure.
*   **Honesty About Completeness:** If you cannot find a complete solution, you must **not** guess or create a solution that appears correct but contains hidden flaws or justification gaps. Instead, you should present only significant partial results that you can rigorously prove. A partial result is considered significant if it represents a substantial advancement toward a full solution. Examples include:
    *   Proving a key lemma.
    *   Fully resolving one or more cases within a logically sound case-based proof.
    *   Establishing a critical property of the mathematical objects in the problem.
    *   For an optimization problem, proving an upper or lower bound without proving that this bound is achievable.
*   **Use TeX for All Mathematics:** All mathematical variables, expressions, and relations must be enclosed in TeX delimiters (e.g., `Let $n$ be an integer.`).

### Output Format ###

Your response MUST be structured into the following sections, in this exact order.

**1. Summary**

Provide a concise overview of your findings. This section must contain two parts:

*   **a. Verdict:** State clearly whether you have found a complete solution or a partial solution.
    *   **For a complete solution:** State the final answer, e.g., "I have successfully solved the problem. The final answer is..."
    *   **For a partial solution:** State the main rigorous conclusion(s) you were able to prove, e.g., "I have not found a complete solution, but I have rigorously proven that..."
*   **b. Method Sketch:** Present a high-level, conceptual outline of your solution. This sketch should allow an expert to understand the logical flow of your argument without reading the full detail. It should include:
    *   A narrative of your overall strategy.
    *   The full and precise mathematical statements of any key lemmas or major intermediate results.
    *   If applicable, describe any key constructions or case splits that form the backbone of your argument.

**2. Detailed Solution**

Present the full, step-by-step mathematical proof. Each step must be logically justified and clearly explained. The level of detail should be sufficient for an expert to verify the correctness of your reasoning without needing to fill in any gaps. This section must contain ONLY the complete, rigorous proof, free of any internal commentary, alternative approaches, or failed attempts.

### Self-Correction Instruction ###

Before finalizing your output, carefully review your "Method Sketch" and "Detailed Solution" to ensure they are clean, rigorous, and strictly adhere to all instructions provided above. Verify that every statement contributes directly to the final, coherent mathematical argument.
\end{lstlisting}

\subsection{Verifier Prompt}
\label{prompt:verification}
\begin{lstlisting}
    You are an expert mathematician and a meticulous grader for an International Mathematical Olympiad (IMO) level exam. Your primary task is to rigorously verify the provided mathematical solution. A solution is to be judged correct **only if every step is rigorously justified.** A solution that arrives at a correct final answer through flawed reasoning, educated guesses, or with gaps in its arguments must be flagged as incorrect or incomplete.

### Instructions ###

**1. Core Instructions**
*   Your sole task is to find and report all issues in the provided solution. You must act as a **verifier**, NOT a solver. **Do NOT attempt to correct the errors or fill the gaps you find.**
*   You must perform a **step-by-step** check of the entire solution. This analysis will be presented in a **Detailed Verification Log**, where you justify your assessment of each step: for correct steps, a brief justification suffices; for steps with errors or gaps, you must provide a detailed explanation.

**2. How to Handle Issues in the Solution**
When you identify an issue in a step, you MUST first classify it into one of the following two categories and then follow the specified procedure.

*   **a. Critical Error:**
    This is any error that breaks the logical chain of the proof. This includes both **logical fallacies** (e.g., claiming that `A>B, C>D` implies `A-C>B-D`) and **factual errors** (e.g., a calculation error like `2+3=6`).
    *   **Procedure:**
        *   Explain the specific error and state that it **invalidates the current line of reasoning**.
        *   Do NOT check any further steps that rely on this error.
        *   You MUST, however, scan the rest of the solution to identify and verify any fully independent parts. For example, if a proof is split into multiple cases, an error in one case does not prevent you from checking the other cases.

*   **b. Justification Gap:**
    This is for steps where the conclusion may be correct, but the provided argument is incomplete, hand-wavy, or lacks sufficient rigor.
    *   **Procedure:**
        *   Explain the gap in the justification.
        *   State that you will **assume the step's conclusion is true** for the sake of argument.
        *   Then, proceed to verify all subsequent steps to check if the remainder of the argument is sound.

**3. Output Format**
Your response MUST be structured into two main sections: a **Summary** followed by the **Detailed Verification Log**.

*   **a. Summary**
    This section MUST be at the very beginning of your response. It must contain two components:
    *   **Final Verdict**: A single, clear sentence declaring the overall validity of the solution. For example: "The solution is correct," "The solution contains a Critical Error and is therefore invalid," or "The solution's approach is viable but contains several Justification Gaps."
    *   **List of Findings**: A bulleted list that summarizes **every** issue you discovered. For each finding, you must provide:
        *   **Location:** A direct quote of the key phrase or equation where the issue occurs.
        *   **Issue:** A brief description of the problem and its classification (**Critical Error** or **Justification Gap**).

*   **b. Detailed Verification Log**
    Following the summary, provide the full, step-by-step verification log as defined in the Core Instructions. When you refer to a specific part of the solution, **quote the relevant text** to make your reference clear before providing your detailed analysis of that part.

**Example of the Required Summary Format**
*This is a generic example to illustrate the required format. Your findings must be based on the actual solution provided below.*

**Final Verdict:** The solution is **invalid** because it contains a Critical Error.

**List of Findings:**
*   **Location:** "By interchanging the limit and the integral, we get..."
    *   **Issue:** Justification Gap - The solution interchanges a limit and an integral without providing justification, such as proving uniform convergence.
*   **Location:** "From $A > B$ and $C > D$, it follows that $A-C > B-D$"
    *   **Issue:** Critical Error - This step is a logical fallacy. Subtracting inequalities in this manner is not a valid mathematical operation.

======================================================================
### Problem ###

[Paste the TeX for the problem statement here]

======================================================================
### Solution ###

[Paste the TeX for the solution to be verified here]

======================================================================
### Verification Task Reminder ###

Your task is to act as an IMO grader. Now, generate the **summary** and the **step-by-step verification log** for the solution above. In your log, justify each correct step and explain in detail any errors or justification gaps you find, as specified in the instructions above.
\end{lstlisting}

\section{Results and Discussion}

\subsection{Performance on IMO 2025}

Our model-agnostic pipeline demonstrated consistent success across three leading LLMs. When equipped with Gemini 2.5 Pro, Grok-4, or GPT-5, the pipeline successfully generated rigorous solutions for 5 out of the 6 problems from the IMO 2025. The full, verbatim proofs for each problem from each model, which constitute the primary evidence for this claim, are provided in Appendix \ref{appA}.

In our initial experiments with Gemini 2.5 Pro, solutions for Problems 1 and 2 were generated with hints. For Problem 1, the hint was to use mathematical induction; for Problem 2, it was to use analytic geometry. We hypothesized that these general hints do not provide problem-specific insight but rather serve to reduce the computational search space. This hypothesis found strong empirical validation in subsequent experiments, where we successfully generated hint-free solutions for both problems, albeit at a higher computational cost. These hint-free solutions were obtained by only removing the hints from the prompt, without any further modifications to our code. This success confirms that the hints primarily improve efficiency rather than enabling a fundamentally new capability. For archival purposes, the original hint-based solutions from our initial experiments are preserved in Appendix \ref{appB}.

Despite the high success rate, the pipeline failed to solve Problem 6, and this failure was consistent across all three base models. For example, the model's output for this problem using Gemini 2.5 Pro, provided in Appendix \ref{a:6}, correctly identifies the trivial upper bound of $4048$. However, its attempt to prove a matching lower bound is built on a flawed premise. The core error is the assertion that any tile must lie entirely in one of two disjoint regions: the cells to the left of the uncovered squares ($C_L$) or the cells to the right ($C_R$). This is incorrect, as a single tile can span columns that are to the left of one row's uncovered square and to the right of another's. This invalidates the subsequent proof. The consistent failure on this problem suggests that certain types of complex combinatorial reasoning remain a significant hurdle for current models, even within a verification-and-refinement framework.

Recent findings by MathArena \cite{matharena2025} highlight a key challenge: single-pass solution generation is often insufficient for complex tasks demanding mathematical rigor. This is evidenced by the baseline accuracies on the IMO 2025, where even a best-of-32 post-selection strategy yielded only 31.6\% for Gemini 2.5 Pro, 21.4\% for Grok-4, and 38.1\% for GPT-5. In sharp contrast, our pipeline achieved a consistent accuracy of approximately 85.7\% across all three models. This substantial improvement demonstrates that the iterative refinement process systematically overcomes the limitations of single-pass generation, such as finite reasoning budgets and the critical errors or justification gaps that often appear in initial drafts. The verifier-guided loop, in particular, proved essential for eliciting rigorous and trustworthy arguments, validating the central thesis of this work: such a pipeline is key to converting the latent capabilities of powerful LLMs into sound mathematical proofs.

\subsection{Generalization to Undergraduate Mathematics}

To assess the broader applicability and robustness of our pipeline, it is essential to evaluate its performance on benchmarks that differ significantly from the IMO. We therefore turn to the International Mathematics Competition for University Students (IMC), a prestigious annual contest held since 1994. Like the IMO, the IMC requires complete and rigorous proofs, making it an excellent testbed for our verification-focused methodology. The IMC problems are drawn from the fields of algebra, analysis (real and complex), geometry, and combinatorics, reflecting the core of a standard undergraduate mathematics curriculum. Thus, the IMC requires a more extensive and advanced knowledge base than the IMO, which is grounded in pre-university mathematics. Its long history and emphasis on rigorous proofs establish the IMC as a well-regarded benchmark for advanced mathematical reasoning.

MathArena independently evaluated our pipeline on the IMC 2025~\cite{imc}. They implemented our publicly available code with Gemini 2.5 Pro as the base model, referring to this implementation as the ``Gemini agent.'' The results demonstrated a substantial performance improvement attributable to our pipeline. The Gemini agent achieved an accuracy of 94.5\%, a score that would have placed it at rank \#3 among the 434 human participants in the official competition \cite{imc_res}. By contrast, the base Gemini 2.5 Pro model scored only 57.7\%, corresponding to rank \#92.

This independent evaluation provides strong external validation for the effectiveness and generalizability of our method. First, the success on the IMC demonstrates that our pipeline is not tailored to IMO-style problems but is robust enough to handle the more knowledge-intensive domain of undergraduate mathematics. Second, because the IMC 2025 took place after the public release of our code on GitHub, the competition serves as a pristine testbed, mitigating the concern of data contamination. Finally, the contrast between the agent's performance (rank \#3) and that of the base model alone (rank \#92) highlights how our verification-and-refinement pipeline translates the latent capabilities of a powerful base model into reliable, high-quality, and competitive mathematical reasoning.

\section{Outlook}

A direct avenue for enhancing our pipeline's capabilities involves leveraging the more powerful, albeit computationally intensive, variants of the base models used in this study. These include Gemini 2.5 Pro Deep Think, Grok-4 Heavy, and GPT-5 Pro. We did not use them in this work due to two significant constraints. First, none of these models are currently available via an API, which precludes their integration into our automated workflow. Second, their current interfaces do not allow for disabling web search. This restriction prevents a fair evaluation on any problem whose solution is available on the Internet. As these models become accessible with the necessary controls via an API, integrating them into our pipeline will be a natural and important next step in pushing the frontiers of automated mathematical reasoning.

While our pipeline is model-agnostic, its current implementation operates within a single-model paradigm, where one base LLM serves as both the solver and the verifier. A natural extension of this work is to develop a multi-model collaborative framework that leverages the strengths of different leading LLMs (Gemini 2.5 Pro, Grok-4, and GPT-5). In such a system, each step of our pipeline, from initial solution generation to iterative refinement and verification, would involve two sub-steps: first, each model would work independently to generate a solution or verification report; second, the models would engage in a collective review \cite{DLT+24}, comparing and critiquing all individual outputs to synthesize a single, consolidated output for that step. This collaborative approach is expected to yield significant benefits. For creative tasks like solution generation, it would pool the diverse reasoning pathways of different models, fostering a richer set of novel ideas. For verification, a subtle error or logic gap missed by one model may be caught by another. By combining the complementary strengths of different models, we believe that such a collaborative system would possess significantly stronger and more reliable mathematical reasoning capabilities.

\section*{Archival Note}

The research presented in this paper was conducted incrementally, and we document its evolution here for transparency across its arXiv versions.

Our initial experiments, detailed in arXiv versions 1 and 2, focused exclusively on the Gemini 2.5 Pro model. In those versions, we presented solutions for Problems 3, 4, and 5 generated without hints. For Problems 1 and 2, we provided solutions generated with general hints (mathematical induction for Problem 1 and analytic geometry for Problem 2).

In arXiv version 3, we successfully generated hint-free solutions for Problems 1 and 2 using Gemini 2.5 Pro, albeit at a higher computational cost. These solutions were obtained by only removing the hints from the prompt, without any further modifications to our code. This result confirmed that the pipeline could solve these problems from first principles.

The present version (version 4) significantly expands the scope of our work to demonstrate the model-agnostic nature of our pipeline. We applied the pipeline to two additional leading models, Grok-4 and GPT-5, and successfully generated hint-free solutions for all five problems (1 through 5). This generalization prompted the change in the paper's title to reflect the pipeline's broad applicability.

To maintain a clear record, the appendix is now organized into two parts: Appendix \ref{appA} contains the complete set of hint-free solutions from all three models, while Appendix \ref{appB} archives the original hint-based solutions for Problems 1 and 2 from our initial experiments with Gemini 2.5 Pro.

\section*{Acknowledgments}

We thank xAI for providing computational credits for using Grok-4. LY is also supported in part by NSF Grant 2221871 and an Amazon Faculty Award.

\bibliographystyle{abbrv}
\bibliography{ref}

\begin{thebibliography}{10}

\bibitem{imc_res}
{IMC2025} preliminary results.
\newblock \url{https://www.imc-math.org.uk/?act=results&by=sum&year=2025}.

\bibitem{imo_official}
{International Mathematical Olympiad}.
\newblock \url{https://www.imo-official.org}.

\bibitem{matharena2025}
{Not Even Bronze: Evaluating LLMs on 2025 International Math Olympiad}.
\newblock \url{https://matharena.ai/imo/},
  \url{https://matharena.ai/?comp=imo--imo_2025}.

\bibitem{IMOF}
Organisation of the {I}nternational {M}athematical {O}lympiad.
\newblock \url{https://imof.co/about-imo/activities/}.

\bibitem{imc}
With flying colors: Language models ace the {I}nternational {M}athematics
  {C}ompetition.
\newblock \url{https://matharena.ai/imc/}.

\bibitem{AG20}
R.~Agarwal and P.~Gaule.
\newblock Invisible geniuses: Could the knowledge frontier advance faster?
\newblock {\em American Economic Review: Insights}, 2(4):409–24, 2020.

\bibitem{ahn2024large}
J.~Ahn, R.~Verma, R.~Lou, D.~Liu, R.~Zhang, and W.~Yin.
\newblock Large language models for mathematical reasoning: Progresses and
  challenges.
\newblock In N.~Falk, S.~Papi, and M.~Zhang, editors, {\em Proceedings of the
  18th Conference of the European Chapter of the Association for Computational
  Linguistics: Student Research Workshop}, pages 225--237, St. Julian{'}s,
  Malta, Mar. 2024. Association for Computational Linguistics.

\bibitem{deepmind2024alphaproof}
{AlphaProof and AlphaGeometry teams}.
\newblock {AI achieves silver-medal standard solving International Mathematical
  Olympiad problems}.
\newblock
  \url{https://deepmind.google/discover/blog/ai-solves-imo-problems-at-silver-medal-level/},
  2024.

\bibitem{BDP+25}
M.~Balunović, J.~Dekoninck, I.~Petrov, N.~Jovanović, and M.~Vechev.
\newblock Math{A}rena: Evaluating {LLM}s on uncontaminated math competitions.
\newblock arXiv:2505.23281.

\bibitem{bytedance2025seed}
{ByteDance Seed AI4Math}.
\newblock Seed-prover: Deep and broad reasoning for automated theorem proving.
\newblock arXiv:2507.23726.

\bibitem{zhou2024dont}
Y.~Cheng, Y.~Chang, and Y.~Wu.
\newblock A survey on data contamination for large language models.
\newblock arXiv:2502.14425.

\bibitem{CTO+15}
Y.~Chervonyi, T.~H. Trinh, M.~Olšák, X.~Yang, H.~Nguyen, M.~Menegali,
  J.~Jung, V.~Verma, Q.~V. Le, and T.~Luong.
\newblock Gold-medalist performance in solving {O}lympiad geometry with
  {AlphaGeometry2}.
\newblock arXiv:2502.03544.

\bibitem{cobbe2021gsm8k}
K.~Cobbe, V.~Kosaraju, M.~Bavarian, M.~Chen, H.~Jun, L.~Kaiser, M.~Plappert,
  J.~Tworek, J.~Hilton, R.~Nakano, C.~Hesse, and J.~Schulman.
\newblock Training verifiers to solve math word problems.
\newblock arXiv:2110.14168.

\bibitem{DLT+24}
Y.~Du, S.~Li, A.~Torralba, J.~B. Tenenbaum, and I.~Mordatch.
\newblock Improving factuality and reasoning in language models through
  multiagent debate.
\newblock In {\em The Twelfth International Conference on Learning
  Representations}, 2024.

\bibitem{FRRB23}
E.~First, M.~N. Rabe, T.~Ringer, and Y.~Brun.
\newblock Baldur: Whole-proof generation and repair with large language models.
\newblock In {\em Proceedings of the 31st ACM Joint European Software
  Engineering Conference and Symposium on the Foundations of Software
  Engineering}, ESEC/FSE 2023, page 1229–1241, New York, NY, USA, 2023.
  Association for Computing Machinery.

\bibitem{gemini2025blog}
{Gemini Team, Google}.
\newblock Gemini 2.5: Pushing the frontier with advanced reasoning,
  multimodality, long context, and next generation agentic capabilities.
\newblock arXiv:2507.06261.

\bibitem{guo2025temporal}
J.~Guo, Y.~Wu, J.~Qiu, K.~Huang, X.~Juan, L.~Yang, and M.~Wang.
\newblock Temporal consistency for {LLM} reasoning process error
  identification.
\newblock arXiv:2503.14495.

\bibitem{he2024olympiadbench}
C.~He, R.~Luo, Y.~Bai, S.~Hu, Z.~Thai, J.~Shen, J.~Hu, X.~Han, Y.~Huang,
  Y.~Zhang, J.~Liu, L.~Qi, Z.~Liu, and M.~Sun.
\newblock {O}lympiad{B}ench: A challenging benchmark for promoting {AGI} with
  {O}lympiad-level bilingual multimodal scientific problems.
\newblock In L.-W. Ku, A.~Martins, and V.~Srikumar, editors, {\em Proceedings
  of the 62nd Annual Meeting of the Association for Computational Linguistics
  (Volume 1: Long Papers)}, pages 3828--3850, Bangkok, Thailand, Aug. 2024.
  Association for Computational Linguistics.

\bibitem{hendrycks2021math}
D.~Hendrycks, C.~Burns, S.~Kadavath, A.~Arora, S.~Basart, E.~Tang, D.~Song, and
  J.~Steinhardt.
\newblock Measuring mathematical problem solving with the math dataset.
\newblock In J.~Vanschoren and S.~Yeung, editors, {\em Proceedings of the
  Neural Information Processing Systems Track on Datasets and Benchmarks},
  volume~1, 2021.

\bibitem{KZZ+24}
R.~Kamoi, Y.~Zhang, N.~Zhang, J.~Han, and R.~Zhang.
\newblock When can {LLM}s actually correct their own mistakes? {A} critical
  survey of self-correction of {LLM}s.
\newblock {\em Transactions of the Association for Computational Linguistics},
  12:1417--1440, 2024.

\bibitem{Kim23}
D.~Kim.
\newblock Looking back on the problem selection committee.
\newblock \url{https://web.stanford.edu/~dkim04/blog/imo-psc-2023/}, 2023.

\bibitem{KBM23}
G.~Kim, P.~Baldi, and S.~McAleer.
\newblock Language models can solve computer tasks.
\newblock In A.~Oh, T.~Naumann, A.~Globerson, K.~Saenko, M.~Hardt, and
  S.~Levine, editors, {\em Advances in Neural Information Processing Systems},
  volume~36, pages 39648--39677. Curran Associates, Inc., 2023.

\bibitem{Deepmind}
T.~Luong, E.~Lockhart, et~al.
\newblock {Advanced version of Gemini with Deep Think officially achieves
  gold-medal standard at the International Mathematical Olympiad}.
\newblock
  \url{https://deepmind.google/discover/blog/advanced-version-of-gemini-with-deep-think-officially-achieves-gold-medal-standard-at-the-international-mathematical-olympiad/},
  2025.

\bibitem{MTG+23}
A.~Madaan, N.~Tandon, P.~Gupta, S.~Hallinan, L.~Gao, S.~Wiegreffe, U.~Alon,
  N.~Dziri, S.~Prabhumoye, Y.~Yang, S.~Gupta, B.~P. Majumder, K.~Hermann,
  S.~Welleck, A.~Yazdanbakhsh, and P.~Clark.
\newblock Self-refine: Iterative refinement with self-feedback.
\newblock In A.~Oh, T.~Naumann, A.~Globerson, K.~Saenko, M.~Hardt, and
  S.~Levine, editors, {\em Advances in Neural Information Processing Systems},
  volume~36, pages 46534--46594. Curran Associates, Inc., 2023.

\bibitem{mahdavi2025brains}
H.~Mahdavi, A.~Hashemi, M.~Daliri, P.~Mohammadipour, A.~Farhadi, S.~Malek,
  A.~K. Yekta~Yazdanifard, and V.~Honavar.
\newblock {Brains vs. Bytes: Evaluating LLM Proficiency in Olympiad
  Mathematics}.
\newblock arXiv:2504.01995.

\bibitem{PIP+24}
D.~Paul, M.~Ismayilzada, M.~Peyrard, B.~Borges, A.~Bosselut, R.~West, and
  B.~Faltings.
\newblock {REFINER}: Reasoning feedback on intermediate representations.
\newblock In Y.~Graham and M.~Purver, editors, {\em Proceedings of the 18th
  Conference of the European Chapter of the Association for Computational
  Linguistics (Volume 1: Long Papers)}, pages 1100--1126, St. Julian{'}s,
  Malta, Mar. 2024. Association for Computational Linguistics.

\bibitem{petrov2025proof}
I.~Petrov, J.~Dekoninck, M.~D. Lyuben~Baltadzhiev, K.~Minchev, M.~Balunović,
  N.~Jovanović, and M.~Vechev.
\newblock {Proof or Bluff? Evaluating LLMs on 2025 USA Math Olympiad}.
\newblock arXiv:2503.21934.

\bibitem{SLJ15}
J.~Sheng, L.~Lyu, J.~Jin, T.~Xia, A.~Gu, J.~Zou, and P.~Lu.
\newblock Solving inequality proofs with large language models.
\newblock arXiv:2506.07927.

\bibitem{trinh2024solving}
T.~H. Trinh, Y.~Wu, Q.~V. Le, H.~He, and T.~Luong.
\newblock Solving olympiad geometry without human demonstrations.
\newblock {\em Nature}, 625(7995):476--482, 2024.

\bibitem{WLS+24}
P.~Wang, L.~Li, Z.~Shao, R.~Xu, D.~Dai, Y.~Li, D.~Chen, Y.~Wu, and Z.~Sui.
\newblock Math-shepherd: Verify and reinforce {LLM}s step-by-step without human
  annotations.
\newblock In L.-W. Ku, A.~Martins, and V.~Srikumar, editors, {\em Proceedings
  of the 62nd Annual Meeting of the Association for Computational Linguistics
  (Volume 1: Long Papers)}, pages 9426--9439, Bangkok, Thailand, Aug. 2024.
  Association for Computational Linguistics.

\bibitem{wang2025survey}
P.-Y. Wang, T.-S. Liu, C.~Wang, Y.-D. Wang, S.~Yan, C.-X. Jia, X.-H. Liu, X.-W.
  Chen, J.-C. Xu, Z.~Li, and Y.~Yu.
\newblock A survey on large language models for mathematical reasoning.
\newblock arXiv:2506.08446.

\bibitem{wei2025openai}
A.~Wei, S.~Hsu, and N.~Brown.
\newblock \url{https://x.com/alexwei_/status/1946477742855532918}.

\bibitem{wei2022chain}
J.~Wei, X.~Wang, D.~Schuurmans, M.~Bosma, B.~Ichter, F.~Xia, E.~Chi, Q.~V. Le,
  and D.~Zhou.
\newblock Chain-of-thought prompting elicits reasoning in large language
  models.
\newblock In S.~Koyejo, S.~Mohamed, A.~Agarwal, D.~Belgrave, K.~Cho, and A.~Oh,
  editors, {\em Advances in Neural Information Processing Systems}, volume~35,
  pages 24824--24837. Curran Associates, Inc., 2022.

\bibitem{WZX+23}
Y.~Weng, M.~Zhu, F.~Xia, B.~Li, S.~He, S.~Liu, B.~Sun, K.~Liu, and J.~Zhao.
\newblock Large language models are better reasoners with self-verification.
\newblock In H.~Bouamor, J.~Pino, and K.~Bali, editors, {\em Findings of the
  Association for Computational Linguistics: EMNLP 2023}, pages 2550--2575,
  Singapore, Dec. 2023. Association for Computational Linguistics.

\end{thebibliography}

\appendix

\section{Verbatim Model Outputs for IMO 2025 Problems (Hint-Free)} \label{appA}

This appendix contains the solutions produced by the models. They are presented verbatim, with the sole exception of minor formatting corrections limited to adding missing TeX delimiters (e.g., ``\$'') or replacing unicode symbols with the corresponding LaTeX symbols where necessary for LaTeX compilation. No part of the original wording, mathematical steps, or logic has been altered.

\subsection{Problem 1}

\begin{problem}
A line in the plane is called \textit{sunny} if it is not parallel to any of the $x$-axis, the $y$-axis, and the line $x+y=0$.

Let $n \ge 3$ be a given integer. Determine all nonnegative integers $k$ such that there exist $n$ distinct lines in the plane satisfying both the following:
\begin{itemize}
    \item For all positive integers $a$ and $b$ with $a + b \le n + 1$, the point $(a,b)$ is on at least one of the lines; and
    \item Exactly $k$ of the lines are sunny.
\end{itemize}
\end{problem}

\subsubsection{Solution by Gemini 2.5 Pro}

Let $P_n = \{ (a,b) \in \mathbb{Z}^2 \mid a > 0, b > 0, a+b \le n+1 \}$ be the set of points to be covered. A line is called *sunny* if it is not parallel to the $x$-axis (slope 0), the $y$-axis (undefined slope), or the line $x+y=0$ (slope -1). Let $k$ be the number of sunny lines in a set of $n$ distinct lines covering $P_n$. Let $K_n$ be the set of all possible values of $k$ for a given integer $n \ge 3$.

**Part 1: The case $n=3$**

For $n=3$, the set of points is $P_3 = \{(1,1), (1,2), (2,1), (1,3), (2,2), (3,1)\}$. We show that $K_3 = \{0,1,3\}$.

*   **$k=0$ is possible:** The set of non-sunny lines $\mathcal{L} = \{x=1, x=2, x=3\}$ covers all points in $P_3$. For any $(a,b) \in P_3$, $a \in \{1,2,3\}$, so $(a,b)$ lies on the line $x=a$. Thus, $0 \in K_3$.
*   **$k=1$ is possible:** Consider $\mathcal{L} = \{x=1, y=1, y=x\}$. The lines $x=1$ and $y=1$ are not sunny. The line $y=x$ (slope 1) is sunny. This set of lines covers all points in $P_3$. Thus, $1 \in K_3$.
*   **$k=3$ is possible:** Consider $\mathcal{L} = \{y=x, x+2y=5, 2x+y=5\}$. The slopes are $1, -1/2, -2$. All three lines are sunny. The line $y=x$ covers $(1,1)$ and $(2,2)$. The line $x+2y=5$ covers $(1,2)$ and $(3,1)$. The line $2x+y=5$ covers $(1,3)$ and $(2,1)$. Together, these three lines cover all six points of $P_3$. Thus, $3 \in K_3$.

*   **$k=2$ is impossible:** We prove this by contradiction. Assume there exists a set of 3 lines with 2 sunny lines ($L_1, L_2$) and 1 non-sunny line ($L_3$) that covers $P_3$.

    **Lemma:** A sunny line contains at most two points of $P_3$.
    **Proof:** The sets of three or more collinear points in $P_3$ are $\{(1,1), (1,2), (1,3)\}$ (on line $x=1$), $\{(1,1), (2,1), (3,1)\}$ (on line $y=1$), and $\{(1,3), (2,2), (3,1)\}$ (on line $x+y=4$). All three of these lines are non-sunny. Any other line can intersect $P_3$ in at most two points. Thus, a sunny line contains at most two points of $P_3$.

    Now we proceed with a case analysis on the non-sunny line $L_3$.
    1.  **$L_3$ covers 3 points of $P_3$**: $L_3$ must be one of $x=1, y=1,$ or $x+y=4$.
        *   If $L_3=x=1$, it covers $\{(1,1),(1,2),(1,3)\}$. The remaining points are $R_1 = \{(2,1),(2,2),(3,1)\}$. These three points are not collinear. To cover them with two lines $L_1, L_2$, one line must cover two points. The lines defined by pairs of points in $R_1$ are $x=2$, $y=1$, and $x+y=4$. All are non-sunny, contradicting that $L_1, L_2$ are sunny.
        *   If $L_3=y=1$, it covers $\{(1,1),(2,1),(3,1)\}$. The remaining points are $R_2 = \{(1,2),(1,3),(2,2)\}$. The lines defined by pairs of points in $R_2$ are $x=1$, $y=2$, and $x+y=4$. All are non-sunny. Contradiction.
        *   If $L_3=x+y=4$, it covers $\{(1,3),(2,2),(3,1)\}$. The remaining points are $R_3 = \{(1,1),(1,2),(2,1)\}$. The lines defined by pairs of points in $R_3$ are $x=1$, $y=1$, and $x+y=3$. All are non-sunny. Contradiction.
    2.  **$L_3$ covers 2 points of $P_3$**: The remaining 4 points must be covered by two sunny lines $L_1, L_2$. By the lemma, each must cover at most 2 points. Thus, each must cover exactly 2 points. The set of 4 remaining points must be partitioned into two pairs, each defining a sunny line. The non-sunny lines covering exactly 2 points of $P_3$ are $x=2, y=2, x+y=3$.
        *   If $L_3=x=2$, it covers $\{(2,1),(2,2)\}$. The remaining points are $R_4 = \{(1,1),(1,2),(1,3),(3,1)\}$. The partitions of $R_4$ into two pairs are:
            - $\{(1,1),(1,2)\}$ (on non-sunny $x=1$) and $\{(1,3),(3,1)\}$ (on non-sunny $x+y=4$).
            - $\{(1,1),(1,3)\}$ (on non-sunny $x=1$) and $\{(1,2),(3,1)\}$ (on sunny $x+2y=5$).
            - $\{(1,1),(3,1)\}$ (on non-sunny $y=1$) and $\{(1,2),(1,3)\}$ (on non-sunny $x=1$).
            In no case do we get two sunny lines. Contradiction.
        *   If $L_3=y=2$, it covers $\{(1,2),(2,2)\}$. The remaining points are $R_5 = \{(1,1),(2,1),(1,3),(3,1)\}$. This case is symmetric to $L_3=x=2$ by reflection across $y=x$, which preserves sunniness. No partition yields two sunny lines. Contradiction.
        *   If $L_3=x+y=3$, it covers $\{(1,2),(2,1)\}$. The remaining points are $R_6 = \{(1,1),(1,3),(2,2),(3,1)\}$. The partitions of $R_6$ into two pairs are:
            - $\{(1,1),(2,2)\}$ (on sunny $y=x$) and $\{(1,3),(3,1)\}$ (on non-sunny $x+y=4$).
            - $\{(1,1),(1,3)\}$ (on non-sunny $x=1$) and $\{(2,2),(3,1)\}$ (on non-sunny $x+y=4$).
            - $\{(1,1),(3,1)\}$ (on non-sunny $y=1$) and $\{(1,3),(2,2)\}$ (on non-sunny $x+y=4$).
            In no case do we get two sunny lines. Contradiction.
    3.  **$L_3$ covers 1 or 0 points of $P_3$**: The remaining $\ge 5$ points must be covered by two sunny lines $L_1, L_2$. By the lemma, they can cover at most $2+2=4$ points. This is not enough. Contradiction.

Since all cases lead to a contradiction, $k=2$ is impossible for $n=3$. Thus, $K_3 = \{0,1,3\}$.

**Part 2: A Preliminary Lemma**

**Lemma:** For any integer $m \ge 2$, the set of points $P_m$ cannot be covered by $m-1$ lines.
**Proof:** We proceed by induction on $m$.
*   **Base Case ($m=2$):** $P_2 = \{(1,1), (1,2), (2,1)\}$. These three points are not collinear, so they cannot be covered by $2-1=1$ line.
*   **Inductive Step:** Assume for some $m \ge 3$ that $P_{m-1}$ cannot be covered by $m-2$ lines. Suppose for contradiction that $P_m$ can be covered by a set $\mathcal{L}$ of $m-1$ lines.
    The set $P_m$ contains the $m$ points $S = \{(1,1), (1,2), \dots, (1,m)\}$, which lie on the line $x=1$. To cover these $m$ points with $m-1$ lines from $\mathcal{L}$, by the Pigeonhole Principle, at least one line in $\mathcal{L}$ must contain at least two points from $S$. Such a line must be the line $x=1$.
    Let $L_1 = (x=1) \in \mathcal{L}$. The remaining $m-2$ lines in $\mathcal{L} \setminus \{L_1\}$ must cover the remaining points $P_m \setminus S = \{(a,b) \in P_m \mid a \ge 2\}$.
    The affine transformation $T(x,y) = (x-1, y)$ maps the set $P_m \setminus S$ bijectively onto $P_{m-1}$, since $T(P_m \setminus S) = \{(a-1,b) \mid a \ge 2, b>0, a+b \le m+1\} = \{(a',b) \mid a' \ge 1, b>0, a'+1+b \le m+1\} = \{(a',b) \mid a'>0, b>0, a'+b \le m\} = P_{m-1}$. The set of transformed lines $\{T(L) \mid L \in \mathcal{L} \setminus \{L_1\}\}$ consists of $m-2$ lines that cover $P_{m-1}$. This contradicts our induction hypothesis.
    Therefore, $P_m$ cannot be covered by $m-1$ lines for any $m \ge 2$.

**Part 3: Key Lemma for $n \ge 4$**

**Lemma:** For $n \ge 4$, any set $\mathcal{L}$ of $n$ lines covering $P_n$ must contain at least one of the lines $x=1$, $y=1$, or $x+y=n+1$.
**Proof:** Assume for contradiction that $\mathcal{L}$ covers $P_n$ but contains none of these three lines.
Let $S_V = \{(1,j) \mid 1 \le j \le n\}$, $S_H = \{(i,1) \mid 1 \le i \le n\}$, and $S_D = \{(l,n+1-l) \mid 1 \le l \le n\}$. Since $x=1 \notin \mathcal{L}$, each of the $n$ lines in $\mathcal{L}$ must intersect the line $x=1$ at exactly one point, so each line covers exactly one point from $S_V$. Similarly, each line covers exactly one point from $S_H$ and one from $S_D$.
This establishes a bijection between the lines in $\mathcal{L}$ and the points in each of the sets $S_V, S_H, S_D$.

Let $L_j$ be the line covering $(1,j)$ for $j \in \{1, \dots, n\}$. Let $L_j$ intersect $S_H$ at $(\sigma(j),1)$ and $S_D$ at $(\tau(j), n+1-\tau(j))$. Since each line covers one point from each set, $\sigma$ and $\tau$ must be permutations of $\{1, \dots, n\}$.
For $j=1$, $L_1$ covers $(1,1) \in S_V \cap S_H$, so $\sigma(1)=1$. For $j=n$, $L_n$ covers $(1,n) \in S_V \cap S_D$, so $\tau(n)=1$.
For $j \in \{2, \dots, n-1\}$, the points $(1,j)$, $(\sigma(j),1)$, and $(\tau(j), n+1-\tau(j))$ are distinct and collinear. The collinearity condition implies $\tau(j) = n+1-j + \frac{(n-j)(j-1)}{\sigma(j)-j}$.

1.  **Case $j=2$ (since $n \ge 4$):** Let $i_2=\sigma(2)$ and $l_2=\tau(2)$.
    $l_2 = n-1 + \frac{n-2}{i_2-2}$. Since $\sigma$ is a permutation and $\sigma(1)=1$, $i_2 \in \{2, \dots, n\}$.
    If $i_2=2$, the line is $x+y=3$. For this line to pass through a point in $S_D$, we need $n+1=3 \implies n=2$, which contradicts $n \ge 4$. So $i_2 \ne 2$.
    Let $d=i_2-2 \in \{1, \dots, n-2\}$. For $l_2$ to be an integer, $d$ must divide $n-2$. If $d < n-2$, then $\frac{n-2}{d} > 1$, so $l_2 = n-1 + \frac{n-2}{d} > n$, which contradicts $l_2 \in \{1, \dots, n\}$. Thus, $d=n-2$, which implies $i_2=n$. This gives $l_2 = n-1 + 1 = n$. So we must have $\sigma(2)=n$ and $\tau(2)=n$.

2.  **Case $j=3$ (since $n \ge 4$):** Let $i_3=\sigma(3)$ and $l_3=\tau(3)$.
    $l_3 = n-2 + \frac{2(n-3)}{i_3-3}$. Since $\sigma$ is a permutation with $\sigma(1)=1, \sigma(2)=n$, we have $i_3 \in \{2, \dots, n-1\}$.
    If $i_3=3$, the line is $x+y=4$. For this to pass through a point in $S_D$, we need $n+1=4 \implies n=3$, which contradicts $n \ge 4$. So $i_3 \ne 3$.
    Let $d=i_3-3$. Then $i_3 \in \{2, 4, \dots, n-1\}$, so $d \in \{-1, 1, 2, \dots, n-4\}$.
    If $d>0$, then $d \le n-4$. For $l_3 \le n$, we need $n-2 + \frac{2(n-3)}{d} \le n \implies \frac{2(n-3)}{d} \le 2 \implies n-3 \le d$. This requires $n-3 \le d \le n-4$, which is impossible.
    If $d<0$, the only possibility is $d=-1$, which means $i_3=2$. Then $l_3 = n-2 + \frac{2(n-3)}{-1} = 4-n$. For $l_3 \in \{1, \dots, n\}$, we need $1 \le 4-n$, which implies $n \le 3$. This contradicts $n \ge 4$.

Since all possibilities for $\sigma(3)$ lead to a contradiction, our initial assumption must be false. Thus, for $n \ge 4$, $\mathcal{L}$ must contain one of the lines $x=1, y=1, x+y=n+1$.

**Part 4: Recurrence Relation for $K_n$**

*   **$K_n \subseteq K_{n-1}$ for $n \ge 4$:**
    Let $\mathcal{L}$ be a valid configuration of $n$ lines for $P_n$ with $k$ sunny lines. By the Key Lemma, $\mathcal{L}$ must contain a line $L \in \{x=1, y=1, x+y=n+1\}$. All these lines are non-sunny. The number of sunny lines in $\mathcal{L} \setminus \{L\}$ is $k$.
    If $L=x+y=n+1$, it covers the points $\{(i, n+1-i)\}_{i=1}^n$. The remaining $n-1$ lines must cover $P_n \setminus L = P_{n-1}$. So $\mathcal{L} \setminus \{L\}$ is a valid configuration for $P_{n-1}$ with $k$ sunny lines. Thus $k \in K_{n-1}$.
    If $L=x=1$, the remaining $n-1$ lines cover $P_n \setminus \{(1,j)\}_{j=1}^n$. The transformation $T(x,y)=(x-1,y)$ maps these points to $P_{n-1}$. The set of transformed lines is a valid configuration for $P_{n-1}$. $T$ preserves slopes, so sunniness is preserved. The number of sunny lines is $k$, so $k \in K_{n-1}$. The case $L=y=1$ is symmetric.
    In all cases, if $k \in K_n$, then $k \in K_{n-1}$.

*   **$K_{n-1} \subseteq K_n$ for $n \ge 4$:**
    Let $\mathcal{L}'$ be a valid configuration for $P_{n-1}$ with $k$ sunny lines. Let $T^{-1}(x',y')=(x'+1, y')$. Let $\mathcal{L}'' = \{T^{-1}(L') \mid L' \in \mathcal{L}'\}$. $\mathcal{L}''$ is a set of $n-1$ lines covering $T^{-1}(P_{n-1}) = \{(a,b) \in P_n \mid a \ge 2\}$. The number of sunny lines in $\mathcal{L}''$ is $k$.
    Consider the set of $n$ lines $\mathcal{L} = \mathcal{L}'' \cup \{x=1\}$. The line $x=1$ is not sunny and covers all points in $P_n$ with $x$-coordinate 1. Thus $\mathcal{L}$ covers $P_n$.
    We must check that the lines in $\mathcal{L}$ are distinct. This requires $x=1 \notin \mathcal{L}''$, which is equivalent to showing that the line $x'=0$ is not in $\mathcal{L}'$. Suppose $x'=0$ is in $\mathcal{L}'$. Since $x'=0$ contains no points of $P_{n-1}$ (where $a>0$), the remaining $(n-1)-1=n-2$ lines in $\mathcal{L}'$ must cover all of $P_{n-1}$. But by our Preliminary Lemma, for $n \ge 4$, $P_{n-1}$ cannot be covered by $n-2$ lines. This is a contradiction.
    Thus, $x=1 \notin \mathcal{L}''$, and $\mathcal{L}$ is a valid configuration for $P_n$ with $k$ sunny lines. So if $k \in K_{n-1}$, then $k \in K_n$.

**Part 5: Final Conclusion**

From Part 4, we have $K_n = K_{n-1}$ for all $n \ge 4$. By induction, this means $K_n = K_3$ for all $n \ge 3$.
In Part 1, we established that $K_3 = \{0, 1, 3\}$.
Therefore, for any integer $n \ge 3$, the set of all possible values for $k$ is $\{0, 1, 3\}$.

\subsubsection{Solution by Grok-4}

Let $S_n = \{(a,b) \mid a,b \ge 1, a+b \le n+1 \text{ integers}\}$, a set of $\binom{n+1}{2}$ points. The boundary consists of three sides: left $a=1$ ($n$ points $(1,b)$ for $b=1$ to $n$), bottom $b=1$ ($n$ points $(a,1)$ for $a=1$ to $n$), anti-diagonal $a+b=n+1$ ($n$ points $(a,n+1-a)$ for $a=1$ to $n$), with corners $(1,1),(1,n),(n,1)$ shared, so total boundary lattice points: $3n - 3$.

We prove by induction on $n \ge 3$ that the possible $k$ are exactly $0,1,3$.

**Base case $n=3$:** $S_3 = \{(1,1),(1,2),(1,3),(2,1),(2,2),(3,1)\}$.

*Constructions:*
- $k=0$: Lines $x=1$ (covers $(1,1),(1,2),(1,3)$), $x=2$ (covers $(2,1),(2,2)$), $x=3$ (covers $(3,1)$); all vertical, non-sunny.
- $k=1$: Lines $x+y=3$ (covers $(1,2),(2,1)$), $x+y=4$ (covers $(1,3),(2,2),(3,1)$), $y=x$ (covers $(1,1)$; sunny slope 1). (Note: $x+y=3,4$ are anti-diagonals, non-sunny.)
- $k=3$: Lines $y=x$ (covers $(1,1),(2,2)$), $y= -x/2 + 5/2$ (covers $(1,2),(3,1)$; slope $-1/2$), $y= -2x + 5$ (covers $(1,3),(2,1)$; slope $-2$); all sunny.

*Impossibility of other $k$:* Since $n=3$, $k \le 3$; we show $k=2$ impossible (others trivial: $k>3$ impossible, $k<0$ invalid). Suppose 3 lines with exactly 2 sunny, 1 non-sunny. The non-sunny is one of: horizontal ($y=c$, $c=1,2,3$), vertical ($x=c$, $c=1,2,3$), anti-diagonal ($x+y=d$, $d=2,3,4$). Each covers: $y=1$: 3 pts $\{(1,1),(2,1),(3,1)\}$; $y=2$: 2 pts $\{(1,2),(2,2)\}$; $y=3$: 1 pt $(1,3)$; $x=1$: 3 pts $\{(1,1),(1,2),(1,3)\}$; $x=2$: 2 pts $\{(2,1),(2,2)\}$; $x=3$: 1 pt $(3,1)$; $x+y=2$: 1 pt $(1,1)$; $x+y=3$: 2 pts $\{(1,2),(2,1)\}$; $x+y=4$: 3 pts $\{(1,3),(2,2),(3,1)\}$. Thus max 3 pts, min 1.

If non-sunny covers 3 pts, remaining $R$ has 3 pts. Possible $R$: for $y=1$, $R=\{(1,2),(1,3),(2,2)\}$; pairs: $(1,2)-(1,3)$ vertical (non-sunny), $(1,2)-(2,2)$ horizontal (non-sunny), $(1,3)-(2,2)$ slope $-1$ (non-sunny); no sunny pair, so cannot cover with 2 sunny (each covers $\le 1$ from $R$, total $\le 2<3$). For $x=1$, $R=\{(2,1),(2,2),(3,1)\}$; pairs: $(2,1)-(2,2)$ vertical, $(2,1)-(3,1)$ horizontal, $(2,2)-(3,1)$ slope $-1$; again no sunny pair. For $x+y=4$, $R=\{(1,1),(1,2),(2,1)\}$; pairs analogously non-sunny. Thus impossible.

If covers 2 pts, $R$ has 4 pts; need to cover with 2 sunny. But 2 sunny cover at most 4 pts only if each covers exactly 2 from $R$ (and disjoint). Check each possible: e.g., for $y=2$, $R=\{(1,1),(1,3),(2,1),(3,1)\}$; possible pairs on sunny: e.g., $(1,1)-(2,1)$ horizontal no; $(1,1)-(1,3)$ vertical no; $(1,1)-(3,1)$ horizontal no; $(1,3)-(2,1)$ slope $-2$; $(1,3)-(3,1)$ slope $-1$ no; $(2,1)-(3,1)$ horizontal no. So sunny pairs: only $(1,3)-(2,1)$. Remaining $(1,1),(3,1)$ are on horizontal (non-sunny). No other sunny pair, cannot partition into two sunny pairs. Similarly for $x=2$, $R=\{(1,1),(1,2),(1,3),(3,1)\}$; sunny pairs: e.g., $(1,2)-(3,1)$ slope $-0.5$; $(1,3)-(3,1)$ slope $-1$ no; $(1,1)-(3,1)$ horizontal no; $(1,1)-(1,2),(1,1)-(1,3)$ vertical no; $(1,2)-(1,3)$ vertical no. Only one sunny pair, remaining two on vertical (non-sunny). For $x+y=3$, $R=\{(1,1),(1,3),(2,2),(3,1)\}$; sunny pairs: $(1,1)-(2,2)$ slope 1; $(1,3)-(3,1)$ slope $-1$ no; $(1,1)-(3,1)$ horizontal; $(1,1)-(1,3)$ vertical; $(2,2)-(3,1)$ slope $-1$ no; $(1,3)-(2,2)$ slope $-1$ no. Only $(1,1)-(2,2)$ sunny, remaining $(1,3),(3,1)$ on slope $-1$ no. Thus no partition into two sunny pairs for any size-2 non-sunny.

If covers 1 pt, $R$ has 5 pts; 2 sunny cover at most $2 \times 2=4<5$ (in $S_3$, no sunny through 3 pts: the only lines through 3 pts are the non-sunny full sides, and no sunny triple collinear, as checked by enumeration: e.g., $(1,1),(1,2),(1,3)$ vertical; etc.), impossible.

Thus only $k=0,1,3$ for $n=3$.

**Inductive step:** Assume for all $3 \le m < n$, possible $k$ for $S_m$ are $0,1,3$. For $S_n$, first prove any covering by $n$ distinct lines must include at least one full boundary side: $x=1$, $y=1$, or $x+y=n+1$ (all non-sunny).

*Lemma 1:* Any sunny line covers at most 2 boundary points. The continuous boundary is the triangle with vertices $(1,1),(1,n),(n,1)$. A line $L$ intersects this closed convex set in a (possibly degenerate) segment. Since $S_n$ is convex, $L \cap$ boundary has at most 2 points (endpoints of the segment). Moreover, since sunny $L$ is not parallel to any side (axes or slope $-1$), $L$ intersects each side in at most 1 point and cannot contain any positive-length segment of a side (would require parallelism). Thus $L \cap$ boundary consists of at most 2 isolated points. Since these intersection points are either at lattice points or between them, but the lemma concerns lattice points on the boundary covered by $L$, and because there is no segment overlap with the boundary, there are no additional lattice points on $L$ lying on the boundary beyond these at most 2 intersection points (if they coincide with lattice points). Hence at most 2 boundary lattice points.

*Lemma 2:* Any non-sunny line not a full boundary side covers at most 2 boundary points. Non-sunny are horizontal ($y=c$), vertical ($x=c$), or anti-diagonal ($x+y=d$). If not full side, then: for $y=c$ ($1<c<n$), intersects left at $(1,c)$, anti-diagonal at $(n+1-c,c)$ (if integer), no bottom; at most 2. For $y=1$ but not full (impossible, as it would be the side). Similarly for partial verticals/anti-diagonals: e.g., $x=c$ ($1<c<n$) intersects bottom at $(c,1)$, anti-diagonal at $(c,n+1-c)$; at most 2. For $x+y=d$ ($2<d<n+1$), intersects left at $(1,d-1)$ (if $d-1 \le n$), bottom at $(d-1,1)$ (if $d-1 \le n$); at most 2 (cannot hit anti-diagonal unless $d=n+1$ full).

Thus, if no full side included, each of $n$ lines covers $\le 2$ boundary points, total $\le 2n$. But boundary has $3n-3 > 2n$ for $n>3$, contradiction. Hence at least one full side, say $x=1$ (non-sunny), covering $n$ points on left. Remaining points $\{(a,b) \in S_n \mid a \ge 2\} \cong S_{n-1}$ via $(a-1,b)$. The other $n-1$ lines cover $S_{n-1}$, so by induction, number of sunny among them is $k' \in \{0,1,3\}$, total $k=k'$. Same if other side chosen first. Thus possible $k \in \{0,1,3\}$.

**Constructions for $n$:** By induction, assume for $m<n$.
- $k=0$: Include $x=1$ (non-sunny), recursively cover $S_{n-1}$ with $n-1$ non-sunny lines.
- $k=1$: Include full sides recursively $n-1$ times to reach $S_1=\{(1,1)\}$, cover with sunny line (e.g., arbitrary sunny through it); total 1 sunny, $n-1$ non-sunny.
- $k=3$: Include full sides recursively $n-3$ times to reach $S_3$, cover with 3 sunny as base; total 3 sunny, $n-3$ non-sunny.

Thus all $k=0,1,3$ achievable, completing the induction.

\subsubsection{Solution by GPT-5}

Let $n \geq 3$ be fixed and define
$S_n := \{ (a, b) \in \mathbb{Z}^2 : a \geq 1, b \geq 1, a + b \leq n + 1 \}$.
A line is called non-sunny if it is horizontal, vertical, or of slope $-1$; otherwise it is sunny.

Lemma 1 (Boundary lemma).
Let $m \geq 4$. Suppose $m$ distinct lines $L_1, \ldots, L_m$ cover $S_m$. Then one of the $L_i$ is exactly one of the three sides $x = 1$, $y = 1$, or $x + y = m + 1$.

Proof.
Let $B_m$ be the set of lattice points on the boundary of the (closed) lattice triangle with vertices $(1,1)$, $(1,m)$, $(m,1)$. The three sides contain $m$, $m$, and $m$ lattice points, and the three vertices are double-counted, so $|B_m| = 3m - 3$. A line that is not equal to any side of this triangle intersects its boundary in at most two points (the triangle is convex). Therefore, if none of the $L_i$ is a side, then together the $L_i$ can cover at most $2m$ boundary lattice points, which is $< 3m - 3$ for $m \geq 4$, a contradiction. Hence at least one $L_i$ is a boundary side. $\qed$

Lemma 2 (Peeling).
In any cover of $S_n$ by $n$ distinct lines, at least $n - 3$ of the lines are non-sunny. Moreover, after removing $n - 3$ boundary sides (all non-sunny) as provided by Lemma 1 iteratively while the current size $m \geq 4$, the remaining three lines from the original family cover a set that is a translate of $S_3$.

Proof.
Initialize $m := n$ and $U := S_m$. By Lemma 1, since $m \geq 4$, among the covering lines there is a boundary side of $U$; remove this line and all points of $U$ on it. The uncovered set is, up to translation, $S_{m-1}$. Indeed:
- Removing $y = 1$ yields $\{(a, b) : a \geq 1, b \geq 2, a + b \leq m + 1\}$, which is the translate by $(0, -1)$ of $S_{m-1}$.
- Removing $x = 1$ yields the translate by $(-1, 0)$ of $S_{m-1}$.
- Removing $x + y = m + 1$ yields $S_{m-1}$ itself.
In all cases the removed line is non-sunny. Replace $m$ by $m - 1$ and $U$ by the uncovered set, and repeat while $m \geq 4$. After exactly $n - 3$ iterations we obtain $m = 3$ and three of the original lines remain and cover a translate of $S_3$. Therefore at least $n - 3$ of the original lines are non-sunny. $\qed$

Remark on invariance.
Translations preserve parallelism and slopes, hence preserve the property of being sunny or non-sunny. Thus, in analyzing the last three lines covering a translate of $S_3$, we may translate the entire configuration so that the uncovered set is exactly $S_3$ without affecting the number of sunny lines.

We now analyze three-line covers of $S_3$. Let
$P := S_3 = \{ (1,1), (1,2), (1,3), (2,1), (2,2), (3,1) \}$.

Lemma 3 (Two-point sunny lines through $P$).
Among all pairs of distinct points in $P$, the only pairs that determine a sunny line are the three disjoint pairs
$\{(1,1),(2,2)\}$, $\{(1,2),(3,1)\}$, $\{(1,3),(2,1)\}$.
Each of these lines contains exactly the two indicated points of $P$. No sunny line contains three points of $P$.

Proof.
Compute slopes for all pairs in $P$. Horizontal pairs (slope $0$), vertical pairs (slope $\infty$), and pairs on lines $x + y = c$ (slope $-1$) are non-sunny. A direct check shows the only remaining pair slopes are $1$ for $(1,1)-(2,2)$, $-1/2$ for $(1,2)-(3,1)$, and $-2$ for $(1,3)-(2,1)$, which are sunny. Each of these three lines misses the other three points of $P$, so each contains exactly two points. Any line through three points of $P$ must be one of the three sides of the triangle, hence non-sunny. $\qed$

Lemma 4 (Non-sunny lines through $P$).
A non-sunny line contains at most three points of $P$. Those with three points are exactly the boundary sides $x = 1$, $y = 1$, $x + y = 4$. The only non-sunny lines with exactly two points of $P$ are $x = 2$, $y = 2$, and $x + y = 3$.

Proof.
Immediate by inspection of columns $x = 1,2,3$, rows $y = 1,2,3$, and diagonals $x + y = c$ with $c \in \{2,3,4\}$. $\qed$

Proposition 5 (Impossibility of exactly two sunny lines on $S_3$).
There is no cover of $P$ by three distinct lines with exactly two sunny lines.

Proof.
Suppose three lines $S_1$, $S_2$ (sunny) and $N$ (non-sunny) cover $P$.

Case 1: $N$ contains at most two points of $P$. By Lemma 3, each sunny line contains at most two points of $P$. To cover six points with three lines, we must have that each line contains exactly two points of $P$, and these three two-point sets form a partition of $P$. Then $N$ is one of the non-sunny two-point lines listed in Lemma 4.
- If $N$ is $x = 2$ with points $\{(2,1),(2,2)\}$, then the remaining four points are $\{(1,1),(1,2),(1,3),(3,1)\}$. By Lemma 3, among these four points the only sunny pair is $\{(1,2),(3,1)\}$. Hence two sunny lines cannot partition these four points into two sunny pairs, a contradiction.
- If $N$ is $y = 2$ with points $\{(1,2),(2,2)\}$, the remaining four points are $\{(1,1),(1,3),(2,1),(3,1)\}$. Among these, the only sunny pair is $\{(1,3),(2,1)\}$. Again impossible.
- If $N$ is $x + y = 3$ with points $\{(1,2),(2,1)\}$, the remaining four points are $\{(1,1),(1,3),(2,2),(3,1)\}$. Among these, the only sunny pair is $\{(1,1),(2,2)\}$. Again impossible.

Case 2: $N$ contains at least three points of $P$. By Lemma 4, $N$ is one of the three sides $x = 1$, $y = 1$, or $x + y = 4$. Consider the three remaining points $Q := P \setminus N$. In each of the three possibilities, $Q$ consists of three points such that any pair among them lies on a non-sunny line:
- If $N$ is $x = 1$, then $Q = \{(2,1),(2,2),(3,1)\}$; pairs are vertical, horizontal, and slope $-1$.
- If $N$ is $y = 1$, then $Q = \{(1,2),(1,3),(2,2)\}$; pairs are vertical, horizontal, and slope $-1$.
- If $N$ is $x + y = 4$, then $Q = \{(1,1),(1,2),(2,1)\}$; pairs are vertical, horizontal, and slope $-1$.
Thus no sunny line contains two points of $Q$. Two sunny lines can cover at most two of the three points in $Q$, contradiction.

In all cases we obtain a contradiction. Hence such a cover does not exist. $\qed$

Proposition 6 (Achievability on $S_3$).
There exist covers of $P$ by three lines with exactly $t$ sunny lines for each $t \in \{0, 1, 3\}$.
- $t = 0$: The three sides $x = 1$, $y = 1$, $x + y = 4$.
- $t = 1$: The lines $x = 1$, $y = 1$, and $y = x$.
- $t = 3$: The three sunny lines through the pairs in Lemma 3.

Combining Propositions 5 and 6, the possible numbers of sunny lines among three lines covering $S_3$ are exactly $t \in \{0, 1, 3\}$.

Conclusion for general $n$.
By Lemma 2, in any cover of $S_n$ by $n$ lines, at least $n - 3$ lines are non-sunny and the remaining three lines cover a translate of $S_3$. By translation invariance, the number $k$ of sunny lines among the original $n$ lines equals the number $t$ of sunny lines among three lines covering $S_3$, hence $k \in \{0, 1, 3\}$ and $k \neq 2$.

Finally, we construct configurations achieving $k \in \{0, 1, 3\}$ for each $n \geq 3$.

- $k = 0$: Take the $n$ lines
  $x = 1, x = 2, \ldots, x = n - 2, y = 1, x + y = n + 1$.
  These are $n$ distinct non-sunny lines. The verticals cover all points with $a \leq n - 2$; the two sides cover the remaining columns $a = n - 1$ and $a = n$.

- $k = 1$: Take the $n - 3$ verticals $x = 1, \ldots, x = n - 3$ and the three lines
  $x = n - 2, y = 1, y = x - (n - 3)$.
  The last is sunny (slope $1$), the others are non-sunny. The verticals cover columns $a \leq n - 3$. The three remaining lines cover the last three columns:
  $(n - 2, b)$ for $b = 1,2,3$ lie on $x = n - 2$; $(n - 1, 1)$ and $(n, 1)$ lie on $y = 1$; $(n - 1, 2)$ lies on $y = x - (n - 3)$.

- $k = 3$: Take the $n - 3$ verticals $x = 1, \ldots, x = n - 3$ and the three sunny lines
  $L_1$: through $(n - 2, 1)$ and $(n - 1, 2)$ (slope $1$),
  $L_2$: through $(n - 1, 1)$ and $(n - 2, 3)$ (slope $-2$),
  $L_3$: through $(n, 1)$ and $(n - 2, 2)$ (slope $-1/2$).
  These cover the last three columns:
  $(n - 2, b)$ for $b = 1,2,3$ lie on $L_1, L_3, L_2$ respectively; $(n - 1, 1)$ lies on $L_2$ and $(n - 1, 2)$ on $L_1$; $(n, 1)$ lies on $L_3$.

Therefore, for every $n \geq 3$, the admissible values of $k$ are exactly $\{0, 1, 3\}$.

\subsection{Problem 2}

\begin{problem}
Let $\Omega$ and $\Gamma$ be circles with centers $M$ and $N$, respectively, such that the radius of $\Omega$ is less than the radius of $\Gamma$. Suppose circles $\Omega$ and $\Gamma$ intersect at two distinct points $A$ and $B$. Let $MN$ intersect $\Omega$ at $C$ and $\Gamma$ at $D$, such that points $C$, $M$, $N$, and $D$ lie on the line in that order. Let $P$ be the circumcenter of triangle $ACD$. Line $AP$ intersects $\Omega$ again at $E \ne A$. Line $AP$ intersects $\Gamma$ again at $F \ne A$. Let $H$ be the orthocenter of triangle $PMN$.

Prove that the line through $H$ parallel to $AP$ is tangent to the circumcircle of triangle $BEF$.

(The orthocenter of a triangle is the point of intersection of its altitudes.)
\end{problem}

\subsubsection{Proof by Gemini 2.5 Pro}

Let the line containing the centers $M$ and $N$ be the x-axis. The radical axis of circles $\Omega$ and $\Gamma$ is the line $AB$, which is perpendicular to the line of centers. We place the origin $O(0,0)$ at the intersection of $MN$ and $AB$.

Let $M=(m,0)$ and $N=(n,0)$. The points $A$ and $B$ have coordinates $(0,a)$ and $(0,-a)$ for some $a>0$. Since $A$ lies on both circles:
$MA^2 = (0-m)^2 + (a-0)^2 = r^2 \implies m^2+a^2=r^2$.
$NA^2 = (0-n)^2 + (a-0)^2 = R^2 \implies n^2+a^2=R^2$.
This gives the key relation $a^2 = r^2-m^2 = R^2-n^2$, which implies $n^2-m^2=R^2-r^2$.
The problem states that $C, M, N, D$ lie on the line in that order, so $x_C < x_M < x_N < x_D$.
$C$ is on $\Omega$ and the x-axis, so its x-coordinate is $m-r$. Thus $C=(m-r,0)$.
$D$ is on $\Gamma$ and the x-axis, so its x-coordinate is $n+R$. Thus $D=(n+R,0)$.

**1. Coordinates of P and H**
$P(x_P, y_P)$ is the circumcenter of $\triangle ACD$. It lies on the perpendicular bisector of $CD$, so its x-coordinate is $x_P = \frac{(m-r)+(n+R)}{2}$.
The condition $PA^2=PC^2$ gives $x_P^2 + (y_P-a)^2 = (x_P-(m-r))^2 + y_P^2$, which simplifies to $x_P^2 + y_P^2 - 2ay_P + a^2 = x_P^2 - 2x_P(m-r) + (m-r)^2 + y_P^2$. This yields $2ay_P = 2x_P(m-r) - (m-r)^2 + a^2$. Substituting the expression for $x_P$:
$2ay_P = (m-r+n+R)(m-r) - (m-r)^2 + a^2 = (m-r)(n+R)+a^2$.

$H(x_H, y_H)$ is the orthocenter of $\triangle PMN$. The altitude from $P$ to $MN$ (the x-axis) is the vertical line $x=x_P$, so $x_H=x_P$.
The altitude from $M$ in $\triangle PMN$ is perpendicular to the line $PN$.

*   **Lemma:** The line $PN$ is perpendicular to the line $AD$.
*   **Proof:** $P$ is the circumcenter of $\triangle ACD$, so by definition, $P$ is equidistant from its vertices. Thus $PA=PD$. This means $P$ lies on the perpendicular bisector of segment $AD$.
    $N$ is the center of circle $\Gamma$. Points $A=(0,a)$ and $D=(n+R,0)$ are on $\Gamma$. The radius of $\Gamma$ is $R$. We have $NA^2 = (0-n)^2+(a-0)^2 = n^2+a^2=R^2$, so $NA=R$. The distance $ND$ is $|(n+R)-n|=R$. Thus $NA=ND=R$. This means $N$ also lies on the perpendicular bisector of segment $AD$.
    Since both $P$ and $N$ lie on the perpendicular bisector of $AD$, the line $PN$ is the perpendicular bisector of $AD$. Thus, $PN \perp AD$.

Since the altitude from $M$ in $\triangle PMN$ is perpendicular to $PN$, and we have shown $PN \perp AD$, this altitude must be parallel to $AD$.
The slope of $AD$ is $k_{AD} = \frac{a-0}{0-(n+R)} = \frac{-a}{n+R}$. The slope of the line $MH$ is $k_{MH} = \frac{y_H-0}{x_H-m} = \frac{y_H}{x_P-m}$.
Equating the slopes gives $y_H = k_{AD}(x_P-m) = -\frac{a(x_P-m)}{n+R}$.

**2. The Tangency Condition**
Let $\omega$ be the circumcircle of $\triangle BEF$ with center $O_\omega(x_c, y_c)$ and radius $R_\omega$. Let $k_{AP}$ be the slope of line $AP$. The line $\ell$ through $H$ parallel to $AP$ has equation $y-y_H = k_{AP}(x-x_P)$. The condition for $\ell$ to be tangent to $\omega$ is that the square of the distance from $O_\omega$ to $\ell$ equals $R_\omega^2$:
$$ \frac{(k_{AP}(x_c-x_P) - (y_c-y_H))^2}{1+k_{AP}^2} = R_\omega^2 $$
Since $B(0,-a)$ is on $\omega$, $R_\omega^2 = (x_c-0)^2+(y_c-(-a))^2 = x_c^2+(y_c+a)^2$.
Let $T_1 = k_{AP}(x_c-x_P) - (y_c-y_H)$ and $T_2 = (1+k_{AP}^2)(x_c^2 + (y_c+a)^2)$. We need to prove $T_1^2 = T_2$.

**3. Coordinates of $E, F$ and $O_\omega$**
The line $AP$ passes through $A(0,a)$ and has slope $k_{AP}$. Its equation is $y=k_{AP}x+a$.
$E$ is the second intersection of $y=k_{AP}x+a$ with $\Omega: (x-m)^2+y^2=r^2$. Substituting $y$ gives $(x-m)^2+(k_{AP}x+a)^2=r^2$, which simplifies to $(1+k_{AP}^2)x^2+2(ak_{AP}-m)x+(m^2+a^2-r^2)=0$. Since $m^2+a^2=r^2$, this becomes $x((1+k_{AP}^2)x+2(ak_{AP}-m))=0$. The solutions are $x=0$ (for point $A$) and $x_E = -\frac{2(ak_{AP}-m)}{1+k_{AP}^2}$.
Similarly, for $F$ on $\Gamma: (x-n)^2+y^2=R^2$, we get $x_F = -\frac{2(ak_{AP}-n)}{1+k_{AP}^2}$.

*   **Lemma:** The coordinates of $O_\omega(x_c, y_c)$ satisfy:
    a) $x_c+k_{AP}y_c = m+n-ak_{AP}$
    b) $y_c = -\frac{(ak_{AP}-m)(ak_{AP}-n)}{a(1+k_{AP}^2)}$
*   **Proof:**
    a) $O_\omega$ lies on the perpendicular bisector of $EF$. The midpoint of $EF$ is $(\frac{x_E+x_F}{2}, \frac{y_E+y_F}{2})$. The slope of $EF$ (line $AP$) is $k_{AP}$. The equation of the perpendicular bisector is $y-\frac{y_E+y_F}{2} = -\frac{1}{k_{AP}}(x-\frac{x_E+x_F}{2})$. Since $O_\omega(x_c, y_c)$ is on this line, we have $x_c+k_{AP}y_c = \frac{x_E+x_F}{2}+k_{AP}\frac{y_E+y_F}{2}$. Using $y_E=k_{AP}x_E+a$ and $y_F=k_{AP}x_F+a$, the right side is $\frac{1+k_{AP}^2}{2}(x_E+x_F)+ak_{AP}$. Substituting the expressions for $x_E$ and $x_F$:
    $\frac{1+k_{AP}^2}{2} \left(-\frac{2(ak_{AP}-m)+2(ak_{AP}-n)}{1+k_{AP}^2}\right) + ak_{AP} = -(2ak_{AP}-m-n)+ak_{AP} = m+n-ak_{AP}$.
    So, $x_c+k_{AP}y_c = m+n-ak_{AP}$.

    b) $O_\omega$ is equidistant from $B(0,-a)$ and $E(x_E, y_E)$. So $O_\omega B^2 = O_\omega E^2$.
    $x_c^2+(y_c+a)^2 = (x_c-x_E)^2+(y_c-y_E)^2$.
    $2ay_c + a^2 = -2x_cx_E + x_E^2 - 2y_cy_E + y_E^2$.
    Using $y_E=k_{AP}x_E+a$:
    $2ay_c + a^2 = -2x_cx_E + x_E^2 - 2y_c(k_{AP}x_E+a) + (k_{AP}x_E+a)^2$.
    $4ay_c = -2x_E(x_c+k_{AP}y_c) + (1+k_{AP}^2)x_E^2 + 2ak_{AP}x_E$.
    Dividing by $x_E \neq 0$: $\frac{4ay_c}{x_E} = -2(x_c+k_{AP}y_c) + (1+k_{AP}^2)x_E + 2ak_{AP}$.
    Using part (a) and the expression for $x_E$:
    $\frac{4ay_c}{x_E} = -2(m+n-ak_{AP}) -2(ak_{AP}-m) + 2ak_{AP} = -2m-2n+2ak_{AP} -2ak_{AP}+2m + 2ak_{AP} = 2(ak_{AP}-n)$.
    $2ay_c = x_E(ak_{AP}-n) = \left(-\frac{2(ak_{AP}-m)}{1+k_{AP}^2}\right)(ak_{AP}-n)$.
    $y_c = -\frac{(ak_{AP}-m)(ak_{AP}-n)}{a(1+k_{AP}^2)}$. This completes the proof of the lemma.

**4. Algebraic Verification**

*   **Lemma:** $T_1 = k_{AP}(x_c-x_P) - (y_c-y_H) = \frac{rR}{a}$.
*   **Proof:** $T_1 = (k_{AP}x_c-y_c) - (k_{AP}x_P-y_H)$.
    From part (a) of the previous lemma, $x_c = m+n-ak_{AP}-k_{AP}y_c$.
    $k_{AP}x_c-y_c = k_{AP}(m+n-ak_{AP})-(1+k_{AP}^2)y_c$. Substituting $y_c$:
    $k_{AP}(m+n-ak_{AP}) + \frac{(ak_{AP}-m)(ak_{AP}-n)}{a} = \frac{ak_{AP}(m+n)-a^2k_{AP}^2+a^2k_{AP}^2-ak_{AP}(m+n)+mn}{a} = \frac{mn}{a}$.
    The slope of $AP$ is $k_{AP} = \frac{y_P-a}{x_P}$, so $k_{AP}x_P = y_P-a$. Thus $k_{AP}x_P-y_H = y_P-a-y_H$.
    $T_1 = \frac{mn}{a} - (y_P-a-y_H) = \frac{mn - a(y_P-a-y_H)}{a}$.
    $2aT_1 = 2mn - 2ay_P + 2a^2 + 2ay_H$.
    Using $2ay_P = (m-r)(n+R)+a^2$ and $2ay_H = -2a^2\frac{x_P-m}{n+R} = -a^2\frac{n-m-r+R}{n+R}$:
    $2aT_1 = 2mn - (mn+mR-rn-rR+a^2) + 2a^2 - a^2\frac{n-m-r+R}{n+R}$.
    $2aT_1 = mn - mR + rn + rR + a^2 - (R^2-n^2)\frac{n-m-r+R}{n+R}$ since $a^2=R^2-n^2$.
    $2aT_1 = mn - mR + rn + rR + (R-n)(R+n) - (R-n)(n-m-r+R)$.
    $2aT_1 = mn - mR + rn + rR + (R-n)[(R+n)-(n-m-r+R)]$.
    The term in brackets is $R+n-n+m+r-R = m+r$.
    $2aT_1 = mn - mR + rn + rR + (R-n)(m+r) = mn - mR + rn + rR + Rm + Rr - nm - nr = 2rR$.
    Thus, $T_1 = \frac{rR}{a}$.

*   **Lemma:** $T_2 = (1+k_{AP}^2)(x_c^2 + (y_c+a)^2) = \frac{r^2R^2}{a^2}$.
*   **Proof:** From the relations for $x_c, y_c$, we have $x_c = \frac{a(m+n)+k_{AP}(mn-a^2)}{a(1+k_{AP}^2)}$ and $y_c+a = \frac{a^2-mn+ak_{AP}(m+n)}{a(1+k_{AP}^2)}$.
    Let $U=a(m+n)$ and $V=mn-a^2$. Then $a(1+k_{AP}^2)x_c = U+k_{AP}V$ and $a(1+k_{AP}^2)(y_c+a) = -V+k_{AP}U$.
    $a^2(1+k_{AP}^2)^2(x_c^2+(y_c+a)^2) = (U+k_{AP}V)^2 + (-V+k_{AP}U)^2 = (U^2+V^2)(1+k_{AP}^2)$.
    So, $T_2 = (1+k_{AP}^2)(x_c^2+(y_c+a)^2) = \frac{U^2+V^2}{a^2} = \frac{(a(m+n))^2+(mn-a^2)^2}{a^2}$.
    We verify that the numerator is $r^2R^2$:
    $r^2R^2 = (m^2+a^2)(n^2+a^2) = m^2n^2+a^2(m^2+n^2)+a^4$.
    $(a(m+n))^2+(mn-a^2)^2 = a^2(m^2+2mn+n^2) + (m^2n^2-2mna^2+a^4) = m^2n^2+a^2(m^2+n^2)+a^4$.
    The expressions are identical. So, $T_2 = \frac{r^2R^2}{a^2}$.

**Conclusion**
From our lemmas, $T_1^2 = (\frac{rR}{a})^2 = \frac{r^2R^2}{a^2}$ and $T_2 = \frac{r^2R^2}{a^2}$.
Since $T_1^2 = T_2$, the tangency condition is satisfied. The line through $H$ parallel to $AP$ is tangent to the circumcircle of $\triangle BEF$.

\subsubsection{Proof by Grok-4}

Place $M$ at $(0,0)$ with radius $r$, so $\Omega$: $x^2 + y^2 = r^2$. Place $N$ at $(d,0)$ with radius $R > r$, so $\Gamma$: $(x - d)^2 + y^2 = R^2$, with $|R - r| < d < R + r$.

The intersections are $A = (m, h)$, $B = (m, -h)$ where $m = (d^2 + r^2 - R^2)/(2d)$, $h = \sqrt{r^2 - m^2} > 0$.

The line $MN$ is the x-axis. It intersects $\Omega$ at $C = (-r, 0)$ and $\Gamma$ at $D = (d + R, 0)$, satisfying the order $C, M, N, D$.

The circumcenter $P = (p, y_p)$ of $\triangle ACD$ satisfies $(p + r)^2 + y_p^2 = (p - d - R)^2 + y_p^2 = (p - m)^2 + (y_p - h)^2$. From the first equality, $p = (d + R - r)/2$. Substituting into the second equality yields $(p - m)^2 + (y_p - h)^2 = (p + r)^2 + y_p^2$, which expands to $-2 y_p h = 2 p (m + r)$, so $y_p = -p (m + r)/h$.

The line $AP$ has slope $s = - [p (m + r) + h^2]/[h (p - m)]$. Let $N = p (m + r) + h^2$, $D = h (p - m)$, so $s = -N/D$.

The second intersection $E$ with $\Omega$ satisfies $u = -2 (m + s h)/(1 + s^2)$. Thus $E = (m + u, h + s u)$.

The second intersection $F$ with $\Gamma$ satisfies $v = -2 (m - d + s h)/(1 + s^2)$. Thus $F = (m + v, h + s v)$.

The orthocenter $H$ of $\triangle PMN$ is at $(p, y_h)$ where $y_h = -h (d - p)/(m + r)$.

The line through $H$ parallel to $AP$ is $s x - y - s p + y_h = 0$.

**Lemma 1:** $h (u + v) + s u v = 0$.

Let $g = 1 + s^2$, $q = m + s h$. Then $u = -2 q / g$, $v = -2 (q - d)/g$.

The condition expands to $2 h (d - 2 m) (1 - s^2) + 4 s (2 m^2 - m d - r^2) = 0$, with $a = 2 h (d - 2 m)$, $b = 4 (2 m^2 - m d - r^2)$.

Substituting $s = -N/D$ yields $E(p) = a (D^2 - N^2) - b N D$.

Expanding coefficients:
- $p^2$: $4 h (m + r) r^2$.
- $p$: $4 h (m + r) r^2 (r - d)$.
- Constant: $-2 d r^2 h^3$.

These match $\lambda$ times those of $Q(p) = 2 p^2 + 2 (r - d) p + d (m - r)$, with $\lambda = 2 h (m + r) r^2$, since $\lambda d (m - r) = -2 d r^2 h^3$.

Now, $Q(p) = 0$: Substitute $p = (d + R - r)/2$ into $Q(p) = \frac{1}{2} (d + R - r)^2 + (r - d) (d + R - r) + d (m - r)$.

Expanding yields $\frac{1}{2} d^2 + \frac{1}{2} R^2 + \frac{1}{2} r^2 + d R - d r - R r - d^2 - d R - r^2 + R r + 2 d r + \frac{1}{2} d^2 + \frac{1}{2} r^2 - \frac{1}{2} R^2 - d r = 0$, as all terms cancel ($d^2$ terms: $\frac{1}{2} - 1 + \frac{1}{2} = 0$; $R^2$ terms: $\frac{1}{2} - \frac{1}{2} = 0$; $r^2$ terms: $\frac{1}{2} - 1 + \frac{1}{2} = 0$; d R terms: $1 - 1 = 0$; R r terms: $-1 + 1 = 0$; d r terms: $-1 + 2 - 1 = 0$).

Thus $E(p) = \lambda Q(p) = 0$, proving the lemma.

**Lemma 2:** The circumcenter of $\triangle BEF$ has $x = d/2$.

The perpendicular bisectors of $BE$ and $BF$ intersect at $x = [p_1 x_1 - y_1 - p_2 x_2 + y_2] / (p_1 - p_2)$, where $x_1 = m + u/2$, $y_1 = s u /2$, $x_2 = m + v/2$, $y_2 = s v /2$, $p_1 = -u / (2 h + s u)$, $p_2 = -v / (2 h + s v)$.

The condition $x = d/2$ reduces to $I = (m - d/2) (p_1 - p_2) + (u/2) p_1 - (v/2) p_2 + (s/2) (v - u) = 0$.

Substituting yields numer = $(v - u) K$, where $K = h (\mathrm{sum} - d + 2 m) + (1/2) s \mathrm{prod} + 2 s h^2 + s^2 h \mathrm{sum} + (s^3 /2) \mathrm{prod}$ and $\mathrm{sum} = u + v$, $\mathrm{prod} = u v$.

Using $\mathrm{prod} = - h \mathrm{sum} / s$, $K = h [ (1/2) \mathrm{sum} - d + 2 m + 2 s h + (1/2) s^2 \mathrm{sum} ] = h [ (1/2) (1 + s^2) \mathrm{sum} + 2 (m + s h) - d ]$.

Substituting $\mathrm{sum} = -2 (2 (m + s h) - d)/(1 + s^2)$ gives $K = h \cdot 0 = 0$.

Thus numer = 0, so $I = 0$, proving $x = d/2$.

**Lemma 3:** $c_y = (d/2 - m)/s - h$.

The perpendicular bisector of $EF$ is $y - (h + s w) = (-1/s) (x - (m + w))$, where $w = (u + v)/2$.

At $x = d/2$, $c_y = h + s w - (1/s) (d/2 - m - w)$.

This equals $(d/2 - m)/s - h$ if $g w + 2 h s = d - 2 m$, where $g = 1 + s^2$.

From $\mathrm{sum} = -2 (2 (m + s h) - d)/g$, $w = (d - 2 m - 2 s h)/g$, so $g w + 2 h s = d - 2 m - 2 s h + 2 h s = d - 2 m$, as required.

**Lemma 4:** $s (d - p - m) + y_h + h = 0$.

Let $L = s (d - p - m) + y_h + h$. Then $I = (m + r) D L = - N (m + r) (d - p - m) - h^2 (d - p) (p - m) + h^2 (p - m) (m + r)$.

Expanding yields $p^2$ coeff: $2 r (m + r)$; $p$ coeff: $2 r (m + r) (r - d)$; constant: $- d r h^2$.

These match $r (m + r) Q(p)$, so $I = r (m + r) Q(p) = 0$, proving $L = 0$.

The circumradius $\rho$ satisfies $\rho^2 = (d/2 - m)^2 + (c_y + h)^2 = (d/2 - m)^2 + [(d/2 - m)/s]^2 = (d/2 - m)^2 (1 + 1/s^2)$, so $\rho = |d/2 - m| \sqrt{1 + s^2} / |s|$.

The distance from $O = (d/2, c_y)$ to $s x - y - s p + y_h = 0$ is $|s (d/2) - c_y - s p + y_h| / \sqrt{1 + s^2}$.

The numerator is $- (d/2 - m) (1 + s^2)/s$, so the distance is $|d/2 - m| \sqrt{1 + s^2} / |s| = \rho$, proving tangency.

the line through H parallel to AP is tangent to the circumcircle of $\triangle BEF$

\subsubsection{Proof by GPT-5}

Set a Cartesian coordinate system with $A=(0,0)$ and the line $MN$ horizontal, namely $y=s$ with $s\neq 0$. Write
$$
M=(m,s),\quad N=(n,s),\quad U:=n-m\neq 0.
$$
Let
$$
r_1:=|AM|=\sqrt{m^2+s^2},\quad r_2:=|AN|=\sqrt{n^2+s^2},\quad K:=M\cdot N=mn+s^2.
$$
The intersections of $MN$ with $\Omega$ and $\Gamma$ are
$$
C=(m-r_1,s),\qquad D=(n+r_2,s).
$$
Let $P$ denote the circumcenter of $\triangle ACD$, and put
$$
a:=\overrightarrow{AP}=P=(p_x,p_y),\qquad b:=R_{90^\circ}(a)=(-p_y,p_x).
$$

1) The points $B,E,F$ and nondegeneracy.
The second intersection of $\Omega$ and $\Gamma$ is the reflection of $A$ across $MN$, namely
$$
B=(0,2s).
$$
For any circle centered at $Q$ and any line through $A$ in the direction $a$, the second intersection $X$ satisfies
$$
X=\frac{2(Q\cdot a)}{|a|^2}a.
$$
Applying this to $\Omega$ and $\Gamma$ yields
$$
E=\frac{2(M\cdot a)}{|a|^2}a,\qquad F=\frac{2(N\cdot a)}{|a|^2}a.
$$
If $a_x=p_x=0$, then $M\cdot a=N\cdot a=s a_y$, whence $E=F=\dfrac{2s a_y}{|a|^2}a=B$; thus $B,E,F$ are collinear (indeed coincident), so their circumcircle does not exist. Conversely, if $a_x\neq 0$, then $M\cdot a\neq N\cdot a$ (since $(N-M)\cdot a=U a_x\neq 0$), so $E\neq F$, and $B\notin AP$ because $a_x\neq 0$; hence $B,E,F$ are noncollinear and $(BEF)$ exists. Therefore, the circumcircle $(BEF)$ exists if and only if $a_x\neq 0$. Since the problem speaks of ``the circumcircle of triangle $BEF$'', we henceforth work under the necessary and sufficient condition
$$
a_x\neq 0.
$$

2) The circumcircle $(BEF)$: its center and radius.
Let $O$ be the center and $R$ the radius of the circumcircle of $\triangle BEF$. In the orthogonal basis $\{a,b\}$ (with $|a|=|b|$), write
$$
O=xa+yb,\qquad E=e a,\qquad F=f a,
$$
where $e:=\dfrac{2(M\cdot a)}{|a|^2}$ and $f:=\dfrac{2(N\cdot a)}{|a|^2}$. Let
$$
\nu_B:=\frac{B\cdot a}{|a|^2}=\frac{2s a_y}{|a|^2},\qquad v_B:=\frac{B\cdot b}{|a|^2}=\frac{2s a_x}{|a|^2}.
$$

- Equal chords $OE=OF$ give
$$
((x-e)^2+y^2)|a|^2=((x-f)^2+y^2)|a|^2,
$$
hence $(x-e)^2=(x-f)^2$. Using $e\neq f$ (since $a_x\neq 0$), we obtain
$$
x=\frac{e+f}{2}=\frac{M\cdot a+N\cdot a}{|a|^2}.
$$

- Equalities $OE=OB$ give
$$
(x-e)^2+y^2=(x-u_B)^2+(y-v_B)^2.
$$
Substituting $x=(e+f)/2$ and expanding in the $\{a,b\}$-basis yields
$$
4s a_x\, y\,|a|^2=4K a_x^2,
$$
i.e.,
$$
b\cdot O=y|a|^2=\frac{K}{s}\,a_x.
$$

- The radius is
$$
R^2=|O-E|^2=((x-e)^2+y^2)|a|^2=\frac{\big((N-M)\cdot a\big)^2}{|a|^2}+y^2|a|^2.
$$
Multiplying by $|a|^2$ and using $b\cdot O=y|a|^2$ gives
$$
(|a|R)^2=\big((N-M)\cdot a\big)^2+(b\cdot O)^2=\big((N-M)\cdot a\big)^2+\Big(\frac{K}{s}a_x\Big)^2.
$$

3) The orthocenter $H$ of $\triangle PMN$.
Since $MN$ is horizontal, the altitude from $P$ is the vertical line $x=p_x$. The altitude through $M$ is perpendicular to $PN$, so there exists $t\in\mathbb{R}$ with
$$
H=M+t\,R_{90^\circ}(N-P)=(m,s)+t(s-p_y,\,-(n-p_x)).
$$
Define
$$
d:=s-p_y,\qquad T:=(n-p_x)(p_x-m).
$$
Imposing that $H$ lies on $x=p_x$ gives $t d=p_x-m$. Hence
$$
y_H-p_y=d-t(n-p_x),
$$
and therefore
$$
(y_H-p_y)d=d^2-T.
$$
Moreover,
$$
b\cdot H=b\cdot P+b\cdot(H-P)=0+p_x(y_H-p_y)=a_x(y_H-p_y).
$$
Finally, from a later relation (see (4) below) we will obtain $d\neq 0$.

4) Using that $P$ is the circumcenter of $\triangle ACD$.
The conditions $|PA|=|PC|$ and $|PA|=|PD|$ are equivalent to the linear equations
$$
C\cdot P=\frac{|C|^2}{2},\qquad D\cdot P=\frac{|D|^2}{2}.
$$
Now
$$
|C|^2=(m-r_1)^2+s^2=2r_1(r_1-m),\quad |D|^2=(n+r_2)^2+s^2=2r_2(r_2+n).
$$
Thus
$$
(m-r_1)p_x+s p_y=r_1(r_1-m),\qquad (n+r_2)p_x+s p_y=r_2(r_2+n).
$$
Subtracting these two equations yields
$$
p_x=\frac{m+n+r_2-r_1}{2}.
$$
From this,
$$
T=(n-p_x)(p_x-m)=-\frac{K-r_1r_2}{2}.
$$
Also, from $(m-r_1)p_x+s p_y=r_1(r_1-m)$, we obtain
$$
s(s-p_y)=(m-r_1)(p_x-m),
$$
i.e.,
$$
s\,d=(m-r_1)(p_x-m).
$$
Since $r_1>|m|$, one has $m-r_1\neq 0$. If $d=0$, then $p_x=m$ by the last display, so $P=(m,s)=M$, which is impossible because $P$ is equidistant from $A,C,D$ while $M$ is not equidistant from $A$ and $D$ (as $D\notin\Omega$). Hence
$$
d\neq 0.
$$

5) A key identity.
We claim
$$
d^2 - T - \frac{K}{s}d=-\frac{r_1r_2}{s}d.
$$
Equivalently,
$$
d^2 - T - \frac{K-r_1r_2}{s}d=0.
$$
Multiplying by $s^2$ and using $s d=(m-r_1)(p_x-m)$ and $T=(n-p_x)(p_x-m)$, we have
$$
s^2(d^2-T)-(K-r_1r_2)s d=(p_x-m)\Big((m-r_1)^2(p_x-m)-s^2(n-p_x)-(K-r_1r_2)(m-r_1)\Big).
$$
Since $s^2=r_1^2-m^2$, the bracket equals
$$
(m-r_1)\Big((m-r_1)(p_x-m)+(r_1+m)(n-p_x)-(K-r_1r_2)\Big).
$$
Using $2p_x=m+n+r_2-r_1$ to evaluate the expression in parentheses, one checks it vanishes identically. Therefore the claimed identity holds.

6) Tangency criterion and conclusion.
The line through $H$ parallel to $AP$ has direction $a$ and normal $b$; its equation is $b\cdot X=b\cdot H$. Its distance to $O$ is $\dfrac{|b\cdot H-b\cdot O|}{|b|}$. Hence this line is tangent to $(BEF)$ if and only if
$$
(b\cdot H-b\cdot O)^2=|b|^2R^2=(|a|R)^2.
$$
By the formulas above,
$$
\begin{aligned}
(b\cdot H-b\cdot O)^2
&=a_x^2\big((y_H-p_y)-K/s\big)^2\\
&=a_x^2\left(\frac{d^2-T}{d}-\frac{K}{s}\right)^2
=a_x^2\left(\frac{d^2-T-(K/s)d}{d}\right)^2.
\end{aligned}
$$
Using the key identity,
$$
(b\cdot H-b\cdot O)^2
=a_x^2\left(\frac{(r_1r_2/s)d}{d}\right)^2=a_x^2\left(\frac{r_1r_2}{s}\right)^2.
$$
On the other hand, from the radius formula,
$$
(|a|R)^2=\big((N-M)\cdot a\big)^2+\Big(\frac{K}{s}a_x\Big)^2
=a_x^2\left(U^2+\left(\frac{K}{s}\right)^2\right),
$$
since $(N-M)=(U,0)$ and $(N-M)\cdot a=U a_x$. Thus tangency is equivalent to
$$
\left(\frac{r_1r_2}{s}\right)^2=U^2+\left(\frac{K}{s}\right)^2,
$$
i.e.,
$$
r_1^2 r_2^2=K^2+s^2 U^2.
$$
Finally, note that
$$
|M|^2=r_1^2,\quad |N|^2=r_2^2,\quad M\cdot N=K,\quad \det(M,N)=m s-n s=-s U.
$$
Therefore the last equality is the identity
$$
|M|^2|N|^2=(M\cdot N)^2+\det(M,N)^2,
$$
which always holds. Hence the line through $H$ parallel to $AP$ is tangent to the circumcircle of $\triangle BEF$, as desired.

\subsection{Problem 3}

\begin{problem}
Let $\mathbb{N}$ denote the set of positive integers. A function $f : \mathbb{N} \to \mathbb{N}$ is said to be \textit{bonza} if
\[
f(a) \mid b^a - f(b)^{f(a)}
\]
for all positive integers $a$ and $b$.

Determine the smallest real constant $c$ such that $f(n) \le cn$ for all bonza functions $f$ and all positive integers $n$.
\end{problem}

\subsubsection{Solution by Gemini 2.5 Pro}

Let $P(a, b)$ be the assertion $f(a) \mid b^a - f(b)^{f(a)}$.

**Step 1: Prove that $f(1)=1$**
Let $C=f(1)$. Assume for contradiction that $C>1$. Let $p$ be a prime divisor of $C$.
The assertion $P(p, 1)$ states $f(p) \mid 1^p - f(1)^{f(p)}$, which is $f(p) \mid 1 - C^{f(p)}$.
Since $p \mid C$, we have $C \equiv 0 \pmod p$, so $1 - C^{f(p)} \equiv 1 \pmod p$.
If $p$ were a divisor of $f(p)$, then $p$ would have to divide $1 - C^{f(p)}$, which is impossible as $1 - C^{f(p)} \equiv 1 \pmod p$. Thus, $p \nmid f(p)$.

Now consider the assertion $P(1, p)$, which states $f(1) \mid p - f(p)^{f(1)}$, i.e., $C \mid p - f(p)^C$.
Since $p \mid C$, this implies $p \mid p - f(p)^C$, which in turn implies $p \mid f(p)^C$.
As $p$ is prime, this means $p \mid f(p)$.
We have derived a contradiction: $p \nmid f(p)$ and $p \mid f(p)$.
Therefore, the assumption $C>1$ must be false. Since $f(1) \in \mathbb N$, we must have $f(1)=1$.

**Step 2: Properties of bonza functions**
Let $S_f = \{p \text{ prime} \mid \exists n \in \mathbb N, p \mid f(n)\}$.

**Property 1:** For any $p \in S_f$ and $b \in \mathbb{N}$, if $p \mid f(b)$, then $p \mid b$.
*Proof:* Let $p \in S_f$. By definition, there exists an $a_0 \in \mathbb{N}$ such that $p \mid f(a_0)$. The bonza condition $P(a_0, b)$ is $f(a_0) \mid b^{a_0} - f(b)^{f(a_0)}$. As $p \mid f(a_0)$, we have $p \mid b^{a_0} - f(b)^{f(a_0)}$. If we assume $p \mid f(b)$, this implies $p \mid b^{a_0}$, and since $p$ is prime, $p \mid b$.

**Lemma A:** For any prime $p \in S_f$, we have $f(b) \equiv b \pmod p$ for all $b \in \mathbb N$.
*Proof:* Let $p \in S_f$. First, we show $f(p)$ is a power of $p$. If $f(p)=1$, then for any $a$ with $p \mid f(a)$, $P(a,p) \implies f(a) \mid p^a-f(p)^{f(a)} = p^a-1$, which contradicts $p \mid f(a)$. So $f(p)>1$. Let $q$ be a prime divisor of $f(p)$. Then $q \in S_f$. By Property 1, since $q \mid f(p)$, we must have $q \mid p$. As $p,q$ are primes, $q=p$. Thus $f(p)=p^k$ for some $k \ge 1$.
Now, $P(p,b) \implies f(p) \mid b^p - f(b)^{f(p)}$, so $p^k \mid b^p - f(b)^{p^k}$. This implies $b^p \equiv f(b)^{p^k} \pmod p$. By Fermat's Little Theorem, $b^p \equiv b \pmod p$ and $f(b)^{p^k} \equiv f(b) \pmod p$. Thus, $b \equiv f(b) \pmod p$.

**Lemma B:** For any prime $q \notin S_f$, we have $f(q)=1$.
*Proof:* Let $q$ be a prime with $q \notin S_f$. Suppose $f(q)>1$. Let $p$ be a prime divisor of $f(q)$. Then $p \in S_f$. By Property 1, since $p \mid f(q)$, we must have $p \mid q$. As $p,q$ are primes, $p=q$. This contradicts the fact that $p \in S_f$ and $q \notin S_f$. Thus, $f(q)=1$.

**Step 3: Classification of bonza functions**
For any $p \in S_f$ and any prime $q \notin S_f$, by Lemma A, $f(q) \equiv q \pmod p$. By Lemma B, $f(q)=1$. Thus, we have the crucial condition:
$$ \forall p \in S_f, \forall q \notin S_f \text{ (prime)}, \quad q \equiv 1 \pmod p \quad (*) $$
We analyze the possible structures for the set of primes $S_f$.
1.  $S_f = \emptyset$: Condition (*) is vacuously true. This is a possible case.
2.  $S_f$ is the set of all primes: Condition (*) is vacuously true as there are no primes $q \notin S_f$. This is a possible case.
3.  $S_f$ is a non-empty proper subset of the set of all primes.
    a) Suppose $S_f$ is an infinite proper subset. Let $q$ be a prime such that $q \notin S_f$. By (*), $q \equiv 1 \pmod p$ for all $p \in S_f$. This means every prime $p \in S_f$ is a divisor of $q-1$. Since $S_f$ is infinite, this implies the integer $q-1$ has infinitely many distinct prime divisors, which is impossible. Thus, $S_f$ cannot be an infinite proper subset.
    b) Suppose $S_f$ is a finite, non-empty proper subset. Let $S_f = \{p_1, \dots, p_r\}$. Let $P = p_1 p_2 \dots p_r$. Condition (*) implies that any prime $q \notin S_f$ must satisfy $q \equiv 1 \pmod P$.
    If $P > 2$, then Euler's totient function $\phi(P) > 1$. This means there exists an integer $a$ with $1 < a < P$ and $\gcd(a, P)=1$. By Dirichlet's theorem on arithmetic progressions, there are infinitely many primes $q$ of the form $kP+a$. We can choose such a prime $q$ that is not in the finite set $S_f$. For this prime $q$, we have $q \notin S_f$, so by our condition, it must satisfy $q \equiv 1 \pmod P$. But we chose $q$ such that $q \equiv a \pmod P$. This leads to $a \equiv 1 \pmod P$, which contradicts our choice of $a$.
    Therefore, the assumption $P>2$ must be false. We must have $P \le 2$.
    Since $S_f$ is non-empty, $P \ge 2$. Thus, $P=2$, which implies $S_f=\{2\}$.
Thus, the only possibilities for $S_f$ are $\emptyset$, $\{2\}$, or the set of all primes.

**Step 4: Analysis of $f(n)/n$ for each case**
*   **Case 1: $S_f = \emptyset$.** This implies that for all $n$, $f(n)$ has no prime factors, so $f(n)=1$. The function $f(n)=1$ for all $n$ is bonza, since $1 \mid b^a - 1^{1}$ is always true. For this function, $f(n)/n = 1/n \le 1$.
*   **Case 2: $S_f$ is the set of all primes.** For any prime $p$, Lemma A gives $f(n) \equiv n \pmod p$. This means $p \mid (f(n)-n)$ for all primes $p$. If $f(n) \ne n$, then $f(n)-n$ is a non-zero integer. A non-zero integer can only have a finite number of prime divisors. This forces $f(n)-n=0$, so $f(n)=n$. The function $f(n)=n$ is bonza, since $a \mid b^a - b^a$ is true. For this function, $f(n)/n = 1$.
*   **Case 3: $S_f = \{2\}$.** The range of $f$ is a subset of $\{1\} \cup \{2^k \mid k \in \mathbb N\}$. By Lemma A, $f(n) \equiv n \pmod 2$. This implies $f(n)=1$ for odd $n$, and $f(n)=2^{k_n}$ for some $k_n \ge 1$ for even $n$.
    Let $n$ be an even integer. Let $v_2(m)$ be the exponent of 2 in the prime factorization of $m$.
    1.  For any odd $b$, $P(n,b) \implies f(n) \mid b^n-f(b)^{f(n)}$. Since $b$ is odd, $f(b)=1$. So $2^{k_n} \mid b^n-1$. This must hold for all odd $b$. Thus, $k_n \le \min_{b \text{ odd}} v_2(b^n-1)$.
        Let $n=2^s t$ with $t$ odd, $s \ge 1$. For any odd $b$, let $x=b^t$, which is also odd. We have $v_2(b^n-1) = v_2(x^{2^s}-1)$. It is a known property that for an odd integer $x$ and $k \ge 1$, $v_2(x^{2^k}-1) = v_2(x^2-1)+k-1$. Applying this with $k=s$, we get $v_2(b^n-1) = v_2((b^t)^2-1)+s-1$.
        To find the minimum value, we must minimize $v_2((b^t)^2-1)$ over odd $b$. The map $\phi_t: (\mathbb Z/8\mathbb Z)^\times \to (\mathbb Z/8\mathbb Z)^\times$ given by $\phi_t(y)=y^t$ is a permutation for any odd $t$. This is because for any $y \in (\mathbb Z/8\mathbb Z)^\times$, $y^2 \equiv 1 \pmod 8$, so for $t=2k+1$, $y^t = y^{2k+1} = (y^2)^k y \equiv y \pmod 8$. Thus, as $b$ runs through odd integers, $b^t \pmod 8$ also takes on all values in $\{1,3,5,7\}$. We can choose $b$ such that $b^t \equiv 3 \pmod 8$. For such $b$, $v_2((b^t)^2-1) = v_2(3^2-1)=v_2(8)=3$. The minimum value of $v_2((b^t)^2-1)$ for $b^t \not\equiv 1 \pmod 8$ is 3.
        So, $\min_{b \text{ odd}} v_2(b^n-1) = 3+s-1=s+2=v_2(n)+2$. Thus, $k_n \le v_2(n)+2$.
    2.  For $b=2$, $P(n,2) \implies f(n) \mid 2^n - f(2)^{f(n)}$. Let $f(2)=2^{k_2}$ for some integer $k_2 \ge 1$. The condition is $2^{k_n} \mid 2^n - 2^{k_2 2^{k_n}}$. This requires $k_n \le v_2(2^n - 2^{k_2 2^{k_n}})$.
        If $n \ne k_2 2^{k_n}$, then $v_2(2^n - 2^{k_2 2^{k_n}}) = \min(n, k_2 2^{k_n})$. The condition becomes $k_n \le \min(n, k_2 2^{k_n})$, which implies $k_n \le n$.
        If $n = k_2 2^{k_n}$, the divisibility is on $0$, which holds for any $k_n$. However, $n = k_2 2^{k_n}$ with $k_2 \ge 1$ implies $n \ge 2^{k_n}$. Since $2^x > x$ for all $x \ge 1$, we have $n > k_n$.
        In all cases, we must have $k_n \le n$.
    Combining these constraints, for any bonza function in this class, $f(n) \le 2^{\min(n, v_2(n)+2)}$ for even $n$.

**Step 5: Construction of the maximal function and determination of $c$**
Let's define a function $f_0$ based on the derived upper bound:
$$ f_0(n) = \begin{cases} 1 & \text{if } n \text{ is odd} \\ 2^{\min(n, v_2(n)+2)} & \text{if } n \text{ is even} \end{cases} $$
Let $k_m = \min(m, v_2(m)+2)$ for any even $m$. We verify that $f_0$ is a bonza function.
-   If $a$ is odd, $f_0(a)=1$. The condition is $1 \mid b^a - f_0(b)^{1}$, which is always true.
-   If $a$ is even and $b$ is odd: $f_0(a)=2^{k_a}, f_0(b)=1$. We need $2^{k_a} \mid b^a-1$. By definition, $k_a \le v_2(a)+2$. As established in Step 4, for any even $a$ and odd $b$, $v_2(b^a-1) \ge v_2(a)+2$. Thus, $k_a \le v_2(a)+2 \le v_2(b^a-1)$, so the condition holds.
-   If $a, b$ are even: $f_0(a)=2^{k_a}, f_0(b)=2^{k_b}$. We need $2^{k_a} \mid b^a - (2^{k_b})^{2^{k_a}}$. Let $E = b^a - 2^{k_b 2^{k_a}}$. We need to show $v_2(E) \ge k_a$.
    Let $v_2(b^a) = a v_2(b)$ and the exponent of 2 in the second term is $k_b 2^{k_a}$.
    If $v_2(b^a) \ne k_b 2^{k_a}$, then $v_2(E) = \min(v_2(b^a), k_b 2^{k_a})$. We have $k_a \le a \le a v_2(b) = v_2(b^a)$ since $v_2(b) \ge 1$. Also $k_b \ge 1$, so $k_b 2^{k_a} \ge 2^{k_a} \ge k_a$ for $k_a \ge 1$. Thus $k_a \le v_2(E)$.
    If $v_2(b^a) = k_b 2^{k_a}$, let $v_b=v_2(b)$ and $m=b/2^{v_b}$. The condition is $a v_b = k_b 2^{k_a}$.
    $E = b^a - 2^{k_b 2^{k_a}} = (m 2^{v_b})^a - 2^{a v_b} = m^a 2^{a v_b} - 2^{a v_b} = 2^{a v_b}(m^a-1)$.
    If $m=1$, $b$ is a power of 2, then $E=0$ and $2^{k_a} \mid 0$ holds.
    If $m>1$, $v_2(E) = a v_b + v_2(m^a-1)$. Since $k_a \le a \le a v_b$, and $v_2(m^a-1) \ge 1$ (as $a$ is even, $m$ is odd), we have $k_a \le v_2(E)$.
    Thus, $f_0$ is a bonza function.

Now we find $\sup_{n} f_0(n)/n$. For odd $n$, $f_0(n)/n = 1/n \le 1$. For even $n=2^s t$ with $t$ odd, $s \ge 1$:
$\frac{f_0(n)}{n} = \frac{2^{\min(2^s t, s+2)}}{2^s t}$.
-   If $t \ge 3$, then $2^s t > s+2$ for $s \ge 1$. (Proof: $g(s)=3 \cdot 2^s - s - 2$ is increasing for $s \ge 1$ and $g(1)=3>0$. So $3 \cdot 2^s > s+2$. Then $t \cdot 2^s \ge 3 \cdot 2^s > s+2$.)
    So $\frac{f_0(n)}{n} = \frac{2^{s+2}}{2^s t} = \frac{4}{t} \le \frac{4}{3}$.
-   If $t=1$, $n=2^s$. $\frac{f_0(2^s)}{2^s} = \frac{2^{\min(2^s, s+2)}}{2^s}$.
    The inequality $2^s \ge s+2$ holds for all integers $s \ge 2$. (Proof by induction: for $s=2$, $4 \ge 4$. Assume $2^k \ge k+2$ for $k \ge 2$, then $2^{k+1}=2 \cdot 2^k \ge 2(k+2)=2k+4 = (k+3)+(k+1) > k+3$.)
    For $s=1$, $n=2$: $\frac{f_0(2)}{2} = \frac{2^{\min(2,3)}}{2} = \frac{2^2}{2} = 2$.
    For $s \ge 2$, since $2^s \ge s+2$, we have $\min(2^s, s+2) = s+2$. So $\frac{f_0(2^s)}{2^s} = \frac{2^{s+2}}{2^s} = 4$.
This value of 4 is achieved for $n=4, 8, 16, \dots$. For example, $f_0(4)=2^{\min(4, v_2(4)+2)} = 2^{\min(4,4)} = 2^4=16$, so $f_0(4)/4=4$.

The supremum of $f(n)/n$ over all bonza functions is 4. For any constant $c<4$, the function $f_0$ and $n=4$ gives $f_0(4)=16 > 4c$, so $f_0(4)>cn$. Thus, the smallest constant is $c=4$.

\subsubsection{Solution by Grok-4}           

First, prove that $f(1) = 1$. Suppose $f(1) = k > 1$. Setting $a = 1$, $b = 1$ gives $k \mid 1^1 - k^{k}$, so $k \mid 1 - k^{k}$. But $k^{k} \equiv 0 \pmod{k}$, so $k \mid 1$, contradicting $k > 1$. Thus $f(1) = 1$.

Next, prove $f(a) \mid a^a$ for all $a$. Setting $b = a$ gives $f(a) \mid a^a - f(a)^{f(a)}$. Clearly $f(a) \mid f(a)^{f(a)}$ since $f(a)^{f(a)} = f(a) \cdot f(a)^{f(a) - 1}$ and $f(a) \ge 1$. Therefore $f(a) \mid a^a$. It follows that the primes dividing $f(a)$ divide $a$, and $v_p(f(a)) \le a \cdot v_p(a)$ for each prime $p$.

Now, prove the tight valuation bounds, starting with odd primes.

**Lemma 1:** For any odd prime $p$ and bonza $f$, $v_p(f(n)) \le v_p(n)$.

Let $k = v_p(n)$, assume for contradiction that $g = v_p(f(n)) = k + 1 \ge 1$. Then $p^g \mid b^n - f(b)^{f(n)}$ for all $b$, so $b^n \equiv f(b)^{f(n)} \pmod{p^g}$.

For $b$ coprime to $p$, if $p \mid f(b)$, then right side $\equiv 0 \pmod{p}$, left side not (since $b$ invertible, $b^n$ invertible), contradiction. Thus $v_p(f(b)) = 0$, and both sides are units mod $p^g$.

Case 1: $\phi(p^g) \nmid n$. Then there exists $r \in (\mathbb{Z}/p^g \mathbb{Z})^*$ with $r^n \not\equiv 1 \pmod{p^g}$ (since the group is cyclic of order $\phi(p^g)$, and if all elements satisfied $x^n = 1$, then the exponent would divide $n$, so $\phi(p^g) \mid n$, contradiction). Choose odd prime $q \ne p$ such that $q \equiv r \pmod{p^g}$ (exists by Dirichlet), $f(n) \not\equiv n \pmod{q}$, and $\gcd(q, f(n)) = 1$ (possible since infinitely many such $q$, and finitely many dividing $f(n)-n$ or $f(n)$). Thus $q^n \not\equiv 1 \pmod{p^g}$, satisfying (i); (ii) and (iii) by choice.

Now, $f(q) \mid q^q$, so $f(q) = q^j$ for $0 \le j \le q$. Consider the condition for $a = q$, $b = n$: $q^j \mid n^q - f(n)^{q^j}$.

If $j = 0$, $f(q) = 1 \mid n^q - f(n)^{1}$, always true, but check the condition for $a = n$, $b = q$: $p^g \mid q^n - f(q)^{f(n)} = q^n - 1^{f(n)} = q^n - 1$. By choice (i), $p^g \nmid q^n - 1$, contradiction.

If $j \ge 1$, note $n^q \equiv n \pmod{q}$ by Fermat. Also, $f(n)^{q^j} \equiv f(n)^{q^j \mod (q-1)} \pmod{q}$ since $\gcd(f(n), q) = 1$. But $q^j \equiv 0 \pmod{q}$, and for the exponent modulo $q-1$, $q \equiv 1 \pmod{q-1}$, so $q^j \equiv 1^j =1 \pmod{q-1}$. Thus $f(n)^{q^j} \equiv f(n)^1 \equiv f(n) \pmod{q}$. By choice (ii), $f(n) \not\equiv n \pmod{q}$, so $n^q \not\equiv f(n)^{q^j} \pmod{q}$, hence $v_q(n^q - f(n)^{q^j}) = 0 < j$, so $q^j \nmid n^q - f(n)^{q^j}$, contradiction.

Thus no possible $f(q)$, contradicting the existence of $f$. 

Case 2: $\phi(p^g) \mid n$ (bad case). Choose odd prime $q \ne p$ such that $f(n) \not\equiv n \pmod{q}$, $\gcd(q, f(n)) = 1$ (exists as before). Additionally, since $f(p) = p^j$ with $j \ge 1$ (if $j=0$, then for $a=n$, $b=p$: $p^g \mid p^n - 1^{f(n)} = p^n - 1$, $v_p(p^n - 1) = 0 < g$, contradiction; hence $j \ge 1$ for any bonza $f$), the group $(\mathbb{Z}/p^j \mathbb{Z})^*$ has exponent $\phi(p^j) = p^{j-1}(p-1)$ not dividing $p$ (since $p-1 \nmid p$). Thus there exists $r \in (\mathbb{Z}/p^j \mathbb{Z})^*$ with $r^p \not\equiv 1 \pmod{p^j}$. Choose $q \equiv r \pmod{p^j}$ (compatible with previous conditions via CRT on finite moduli). 

For this $q$, $j \ge 1$ fail as in Case 1 (argument identical, independent of (i)). For $j=0$, the pair $(p, q)$ fails: $f(p) \mid q^p - 1^{f(p)} = q^p - 1$, but by choice $q^p \not\equiv 1 \pmod{p^j} = \pmod{f(p)}$, so $f(p) \nmid q^p - 1$, contradiction.

Thus no possible $f(q)$, contradicting the existence of $f$. Hence $v_p(f(n)) \le v_p(n)$.

**Lemma 2:** For any bonza $f$, $v_2(f(n)) \le v_2(n) + 2$.

Let $v = v_2(n)$, assume for contradiction $e = v_2(f(n)) \ge v + 3$. First, if $v = 0$ (i.e., $n$ odd), then $f(n) \mid n^n$ and $n^n$ odd imply $f(n)$ odd, so $e = 0$. But $0 \ge 3$ is false, contradiction. Thus $v \ge 1$ necessarily under the assumption.

Set $b=3$, then $2^e \mid 3^n - f(3)^{f(n)}$. Since $f(3) \mid 3^3 = 27$, $f(3) = 3^j$ for $0 \le j \le 3$.

If $j=0$, $f(3)=1$, $v_2(3^n -1)$. Since $v \ge 1$, $n$ even, lifting the exponent (LTE) for $p=2$ gives $v_2(x^n - y^n) = v_2(x-y) + v_2(x+y) + v_2(n) - 1$ when $x,y$ odd, $n$ even. Thus $v_2(3^n -1) = v_2(3-1) + v_2(3+1) + v -1 =1+2+v-1 = v+2 < e$.

If $j \ge 1$, compute $v_2(3^n - 3^{j f(n)})$. First, $j f(n) \ne n$: if $j f(n) = n$, then $f(n) = n/j$, $j \mid n$ (since $f(n)$ integer), $e = v_2(n/j) = v - v_2(j) \le v \le v+2 < v+3 \le e$, contradiction. Thus $n \ne j f(n)$. Let $m = |j f(n) - n| >0$. Since $e \ge v+3 > v$, $v_2(j f(n)) = v_2(j) + e \ge e > v = v_2(n)$, so $v_2(m) = v$. Without loss of generality assume $j f(n) > n$ (the other case symmetric), then $v_2(3^n - 3^{j f(n)}) = v_2(3^n (1 - 3^{m})) = v_2(3^n) + v_2(1 - 3^{m}) = 0 + v_2(3^m - 1)$ (negative sign irrelevant for $v_2$). Apply LTE to $3^m - 1^m$: $m$ even (since $j f(n)$ even, $n$ even for $v \ge 1$), so $v_2(3^m -1) =1+2 + v_2(m) -1 = v+2 <e$.

Hence $v_2(f(n)) \le v + 2$.

Now, combining: write $n = 2^{v} \prod_{p \ odd} p^{v_p(n)}$. Then $f(n) = 2^{e} \prod_{p \ odd} p^{f_p}$ with $e \le v + 2$, $f_p \le v_p(n)$ for odd $p$. The maximum occurs when $e = v+2$ and $f_p = v_p(n)$ for all odd $p$, giving $f(n) \le 2^{v+2} \cdot (n / 2^{v}) = 4n$.

For the lower bound, define $f$ by: $f(1)=1$; for $n>1$ odd, $f(n)=1$; for $n=2^d$ with $d \ge 2$, $f(n)=2^{d+2}$; for $n=2$, $f(n)=4$; for $n$ even not a power of $2$, $f(n)=2$.

First, verify $f(n) \mid n^n$ for all $n$:
- If $n=1$, $1 \mid 1^1$.
- If $n>1$ odd, $1 \mid n^n$.
- If $n=2^d$, $d \ge 2$, $2^{d+2} \mid (2^d)^{2^d} = 2^{d \cdot 2^d}$, since $d \cdot 2^d \ge 2 \cdot 4 = 8 > 4 = d+2$ for $d=2$, and larger for $d>2$.
- If $n=2$, $4 \mid 2^2 =4$.
- If $n$ even not power of $2$, $2 \mid n^n$ since $n$ even.

Now, verify bonza: $f(a) \mid b^a - f(b)^{f(a)}$ for all $a,b$. Since $f(a)$ is always a power of $2$ (or $1$), it suffices to check $v_2(b^a - f(b)^{f(a)}) \ge v_2(f(a))$ when $f(a) >1$, or trivial when $=1$.

**Case 1: $a=1$.** $f(a)=1$, trivial.

**Case 2: $a>1$ odd.** $f(a)=1$, trivial.

**Case 3: $a$ even not a power of $2$.** $f(a)=2$, so need $v_2(b^a - f(b)^2) \ge 1$ (i.e., even).
- If $b$ even, $b^a$ even, $f(b) \ge 2$ even so $f(b)^2$ even, even-even=even.
- If $b=1$, $1^a -1^2=0$, even.
- If $b>1$ odd, $b^a$ odd (odd to even power=odd), $f(b)=1$, $1^2=1$ odd, odd-odd=even.

**Case 4: $a=2^d$, $d \ge 1$.** $f(a)=2^{d+2}$ if $d \ge 2$, $f(a)=4=2^2$ if $d=1$. Need $v_2(b^{2^d} - f(b)^{f(a)}) \ge v_2(f(a))$.

**Subcase 4.1: $b=1$.** $1^{2^d} - f(1)^{f(a)}=1-1=0$, $v_2=\infty \ge$ anything.

**Subcase 4.2: $b>1$ odd.** $f(b)=1$, so $v_2(b^{2^d} -1)$. By LTE ($x=b$ odd, $y=1$ odd, $n=2^d$ even): $v_2(b^{2^d}-1)=v_2(b-1)+v_2(b+1)+d-1$.
- Cannot have $v_2(b-1)=v_2(b+1)=1$ (would require $b \equiv 1 \pmod{4}$ and $b \equiv -1 \pmod{4}$, impossible).
- Minimal when $\min(v_2(b-1),v_2(b+1))=1$, other $=2$: $1+2+d-1=d+2$.
- For $d \ge 2$, $d+2 \ge v_2(f(a))=d+2$.
- For $d=1$, actual minimal is $3$ (e.g., $b=3$: $v_2(9-1)=v_2(8)=3$; $b=5$: $v_2(25-1)=v_2(24)=3$), $\ge 2=v_2(f(a))$.

**Subcase 4.3: $b=2^k$, $k \ge 1$.** Let $e_1 = k \cdot 2^d = v_2(b^{2^d})$. If $k=1$, $f(b)=4=2^2$; if $k \ge 2$, $f(b)=2^{k+2}$. Let $e_2 = v_2(f(b)^{f(a)})$.
- If $k=1$, $e_2 = 2 \cdot f(a)$.
- If $k \ge 2$, $e_2 = (k+2) \cdot f(a)$.
Now compare $e_1$ and $e_2$.
- For $d \ge 2$, $f(a)=2^{d+2}$. If $k=1$, $e_1=2^d$, $e_2=2 \cdot 2^{d+2}=2^{d+3} > 2^d$ (since $d \ge 2$). If $k \ge 2$, $e_2=(k+2)2^{d+2} \ge 4 \cdot 2^{d+2}=2^{d+4} > k 2^d =e_1 ((k+2)/k \cdot 4 >4 >1)$.
- For $d=1$, $f(a)=4$. If $k=1$, $e_1=2$, $e_2=2\cdot4=8>2$. If $k \ge 2$, $e_2=(k+2)4 > k \cdot 2 =e_1/2$? Wait, earlier: $2(k+2)/k >2>1$, yes.
Thus always $e_2 > e_1$, so $v_2= e_1 + v_2(1 - 2^{e_2 - e_1})$. Since $e_2 - e_1 >0$, $2^{...}$ even, $1$-even=odd, $v_2=0$, total $v_2=e_1 = k 2^d \ge 2^d$. For $d \ge 2$, $2^d \ge d+2$ (e.g., $d=2$: $4=4$; $d=3$: $8>5$). For $d=1$, $k \cdot 2 \ge 2 \ge 2$.

**Subcase 4.4: $b$ even not power of $2$.** $f(b)=2$, $f(a)=2^{d+2}$ ($d \ge 2$) or $4$ ($d=1$), so $v_2(b^{2^d} - 2^{f(a)})$. Let $v_2(b)=s \ge 1$, $b=2^s m$, $m>1$ odd. Then $v_2(b^{2^d})=s 2^d$, $v_2(2^{f(a)})=f(a)=2^{d+2}$ or $4=2^2$.
- If $s 2^d < f(a)$, $v_2= s 2^d + v_2(m^{2^d} - 2^{f(a) - s 2^d})$. Since $f(a) - s 2^d >0$, $2^{...}$ even, $m^{2^d}$ odd, odd-even=odd, $v_2=0$, total $s 2^d$. For $d \ge 2$, $s \ge 1$, $2^d \ge d+2$; for maximal $s<4$, $3 \cdot 2^d \ge 12 >4$ for $d=2$, larger later. For $d=1$, $2s \ge 2$.
- If $s 2^d = f(a)$, $v_2= f(a) + v_2(m^{2^d} -1) \ge f(a) + d+2 > f(a)$.
- If $s 2^d > f(a)$, $v_2= f(a) + v_2(2^{s 2^d - f(a)} m^{2^d} -1)$. Since $s 2^d - f(a)>0$, even $\cdot$ odd = even, even-1=odd, $v_2=0$, total $f(a) \ge v_2(f(a))$.

Thus $f$ is bonza, achieving $f(2^d)/2^d =4$ for $d \ge 2$.

Hence the smallest $c=4$. 
\boxed{4}

\subsubsection{Solution by GPT-5}

Let $f: \mathbb{N} \to \mathbb{N}$ be bonza, i.e., for all positive integers $a,b$,
\begin{align}f(a) \mid \bigl(b^a - f(b)^{f(a)}\bigr). \tag{1
}\end{align}

Lemma 1. We have $f(1) = 1$.
Proof. Set $(a,b) = (1,1)$ in $(1)$. Then $f(1) | 1 - f(1)^{f(1)}$. Reducing modulo $f(1)$ gives $0 \equiv 1 \pmod{f(1)}$, hence $f(1) = 1$. \qed

Lemma 2. For every $n$, each prime divisor of $f(n)$ divides $n$.
Proof. Put $a = b = n$ in $(1)$. Then $f(n) | (n^n - f(n)^{f(n)})$. If $p | f(n)$ is prime, then reducing modulo $p$ yields $n^n \equiv 0 \pmod p$, hence $p | n$. \qed

Lemma 3. If $p$ is prime and $p | f(a)$, then $p | f(b)$ for every $b$ with $p | b$.
Proof. From $(1)$ and $p | f(a)$, reduce modulo $p$: for all $b$, $b^a \equiv f(b)^{f(a)} \pmod p$. If $p | b$, then $b^a \equiv 0 \pmod p$, so $f(b)^{f(a)} \equiv 0 \pmod p$, whence $p | f(b)$. \qed

Lemma 4 (rigidity modulo an odd prime value). Let $p$ be an odd prime with $f(p) > 1$. Then for all $b$,
\begin{align}f(b) \equiv b \pmod p. \tag{2}\end{align}
Proof. By Lemma 2, $f(p) = p^e$ with $e \ge 1$. Taking $a = p$ in $(1)$ and reducing modulo $p$, we get for all $b$,
\begin{align}
    b^p \equiv f(b)^{p^e} \pmod p. \tag{3}
\end{align}
If $p \nmid b$, then $b \in \mathbb{F}_p^\times$. Since $p^e \equiv 1 \pmod{p-1}$, we have $y^{p^e} \equiv y$ in $\mathbb{F}_p^\times$ for all $y$, and by Fermat $b^p \equiv b \pmod p$. Hence $f(b) \equiv b \pmod p$. If $p | b$, then both sides of $(3)$ are $0$ modulo $p$, so $f(b) \equiv b \equiv 0 \pmod p$. Thus $(2)$ holds for all $b$. \qed

Define the set of "odd prime values":
$$P := \{\, p \text{ odd prime} : f(p) > 1 \,\}.$$

Proposition 1. If $P \neq \emptyset$, then $f(n) = n$ for all $n \in \mathbb{N}$.
Proof. Step 1: $P$ is infinite. Pick $p_0 \in P$. We construct inductively a sequence of distinct odd primes in $P$. Start with $S_0 = \emptyset$. Given a finite set $S_k$ of odd primes already obtained in $P$ (possibly empty), set
$$M_k := \prod_{r \in S_k} r \quad (\text{with } M_k = 1 \text{ if } S_k = \emptyset), \quad b_k := 2 + p_0 M_k.$$
Then $b_k$ is odd and $b_k \equiv 2 \pmod{p_0}$. By Lemma 4 (applied to $p_0$), $f(b_k) \equiv b_k \equiv 2 \pmod{p_0}$, hence $f(b_k) \neq 1$. Let $r_k$ be an odd prime dividing $f(b_k)$. By Lemma 2, $r_k | b_k$. For any $r \in S_k$, $b_k \equiv 2 \pmod r$, so $r \nmid b_k$, hence $r_k \notin S_k$. By Lemma 3 with $a = b_k$ and $b = r_k$ we get $r_k | f(r_k)$, so $f(r_k) > 1$ and $r_k \in P$. Define $S_{k+1} := S_k \cup \{r_k\}$. Thus we can construct infinitely many distinct $r_k \in P$, proving $P$ is infinite.

Step 2: $f$ is the identity. Fix $n \in \mathbb{N}$. For each $p \in P$ we have $f(n) \equiv n \pmod p$ by Lemma 4 (applied to $p$, which satisfies $p | f(p)$ by Lemma 3 since $p \in P$). Since $P$ is infinite, we can choose infinitely many $p \in P$ with $p \nmid n$ (there are only finitely many primes dividing $n$). For all these primes $p$ we have $p | (f(n) - n)$. Hence $f(n) - n$ is divisible by infinitely many distinct primes. Therefore $f(n) - n = 0$, i.e., $f(n) = n$. As $n$ was arbitrary, $f$ is the identity. \qed

Corollary 1. Either $f(n) = n$ for all $n$, or else $f$ takes no odd prime value (equivalently, $P = \emptyset$).

We now analyze the complementary case.

Proposition 2 (the 2-adic branch). Suppose $P = \emptyset$. Then:
(i) $f(n) = 1$ for all odd $n$;
(ii) every $f(n)$ is a power of $2$;
(iii) for even $a$ with $s = v_2(a)$, we have $v_2(f(a)) \le s + 2$;
Consequently, for all $n$ we have $f(n) \le 2^{v_2(n)+2} \le 4n$.

Proof. (ii) If an odd prime $q$ divided some $f(n)$, then by Lemma 3 applied with that $q$ and $b = q$ we would get $q | f(q)$, i.e., $q \in P$, contradicting $P = \emptyset$. Thus all $f(n)$ are powers of $2$.

(i) If $n$ is odd and $f(n)$ were even, then the prime $2$ would divide $n$ by Lemma 2, a contradiction. Thus $f(n) = 1$ for odd $n$.

(iii) Let $a$ be even and $s = v_2(a)$. For every odd $b$ we have $f(b) = 1$ by (i), so $(1)$ yields
$$f(a) \mid (b^a - 1) \quad \text{for all odd } b.$$
Taking 2-adic valuations and using the LTE formula for $2$ (valid for odd $b$ and even $a$),
$$v_2(b^a - 1) = v_2(b - 1) + v_2(b + 1) + v_2(a) - 1.$$
Among odd $b$, one of $b-1$ and $b+1$ is exactly divisible by $2$ and the other by $4$, so the sum $v_2(b-1) + v_2(b+1)$ is at least $3$, and this minimum $3$ is attained (e.g. $b \equiv 3$ or $5 \pmod 8$). Hence
$$\min_{b \text{ odd}} v_2(b^a - 1) = s + 2,$$
so $v_2(f(a)) \le s + 2$. Consequently,
$$f(n) \le 2^{v_2(n)+2} = 4 \cdot 2^{v_2(n)} \le 4n.$$ \qed

Combining Proposition 1 and Proposition 2, we conclude that for every bonza $f$ and every $n$,
\begin{align}f(n) \le 4n. \tag{4}\end{align}

We now prove sharpness.

Construction. Define $F: \mathbb{N} \to \mathbb{N}$ by
$$F(n) = \begin{cases}
1, & n \text{ odd},\\
4, & n = 2,\\
2^{v_2(n)+2}, & n \ge 4 \text{ even}.
\end{cases}$$
We claim $F$ is bonza and attains $F(2^s) = 2^{s+2} = 4\cdot 2^s$ for all $s \ge 2$.

Verification that $F$ is bonza. Fix $a,b \in \mathbb{N}$. We must show
\begin{align}F(a) \mid \bigl(b^a - F(b)^{F(a)}\bigr). \tag{5}\end{align}
If $a$ is odd then $F(a) = 1$ and $(5)$ is trivial. Assume $a$ is even and write $s := v_2(a)$. Set $k := v_2(F(a))$; explicitly, $k = 2$ if $a = 2$, and $k = s + 2$ if $a \ge 4$.

Case 1: $b$ odd. Then $F(b) = 1$, so we need $2^k \mid (b^a - 1)$. By the LTE formula quoted above,
$$v_2(b^a - 1) = v_2(b - 1) + v_2(b + 1) + s - 1 \ge s + 2.$$
Hence $v_2(b^a - 1) \ge k$ in both subcases ($k = 2$ when $a = 2$; $k = s + 2$ when $a \ge 4$), proving $(5)$.

Case 2: $b$ even. Write $t := v_2(b) \ge 1$ and write $b = 2^t u$ with $u$ odd. Then
$$v_2(b^a) = a t.$$
Also,
$$v_2(F(b)) = \begin{cases}
2, & b = 2 \ (t = 1),\\
t + 2, & b \ge 4 \ (t \ge 2),
\end{cases}$$
hence
$$v_2\bigl(F(b)^{F(a)}\bigr) = v_2(F(b)) \cdot 2^k \ge 2 \cdot 2^k \ge k+1 \quad (\text{since } k \ge 2).$$
Therefore
\begin{align}v_2\bigl(b^a - F(b)^{F(a)}\bigr) \ge \min\bigl(a t, \ v_2(F(b)^{F(a)})\bigr). \tag{6}\end{align}
We claim $a t \ge k$. Indeed:
- If $a = 2$, then $k = 2$ and $a t = 2t \ge 2$.
- If $a \ge 4$, then with $s = v_2(a)$ we have $a \ge 2^s$. If $s = 1$ then $a \ge 4 > 3 = s+2$; if $s \ge 2$ then $2^s \ge s + 2$, hence $a \ge s + 2$. Thus in all cases $a \ge s + 2 = k$, and so $a t \ge a \ge k$.
Combining with $v_2(F(b)^{F(a)}) \ge k+1$, $(6)$ gives $v_2(b^a - F(b)^{F(a)}) \ge k$, i.e., $2^k | (b^a - F(b)^{F(a)})$. This proves $(5)$.

Thus $F$ is bonza. Finally, for $n = 2^s$ with $s \ge 2$ we have $F(n) = 2^{s+2} = 4 \cdot 2^s$, so equality is attained in the bound $f(n) \le 4n$ infinitely often. Hence the optimal constant is $c = 4$.

Conclusion. For every bonza function $f$, one has $f(n) \le 4n$ for all $n$, and the constant $4$ is best possible. \qed

\subsection{Problem 4}

\begin{problem}
A proper divisor of a positive integer $N$ is a positive divisor of $N$ other than $N$ itself.

The infinite sequence $a_1, a_2, \ldots$ consists of positive integers, each of which has at least three proper divisors. For each $n \ge 1$, the integer $a_{n+1}$ is the sum of the three largest proper divisors of $a_n$.

Determine all possible values of $a_1$.
\end{problem}

\subsubsection{Solution by Gemini 2.5 Pro}

Let $f(N)$ be the sum of the three largest proper divisors of a positive integer $N$. The sequence is defined by $a_{n+1} = f(a_n)$ for $n \ge 1$. The problem requires that each $a_n$ has at least three proper divisors. A proper divisor of $N$ is any positive divisor of $N$ except $N$ itself. The condition that $N$ has at least three proper divisors means that the number of divisors of $N$, denoted by $d(N)$, must satisfy $d(N)-1 \ge 3$, so $d(N) \ge 4$.

Let the divisors of $N$ in increasing order be $1=d_1 < d_2 < d_3 < \dots < d_k=N$, where $k=d(N) \ge 4$. The proper divisors of $N$ are $\{d_1, d_2, \dots, d_{k-1}\}$. The three largest proper divisors are $d_{k-1}, d_{k-2}, d_{k-3}$.
The map $d \mapsto N/d$ is a bijection on the set of divisors of $N$ that reverses the order. Therefore, the largest proper divisor is $d_{k-1} = N/d_2$, the second largest is $d_{k-2} = N/d_3$, and the third largest is $d_{k-3} = N/d_4$. This holds for any $N$ with $d(N) \ge 4$.
Thus, the sum of the three largest proper divisors is given by:
$f(N) = d_{k-1} + d_{k-2} + d_{k-3} = \frac{N}{d_2} + \frac{N}{d_3} + \frac{N}{d_4} = N\left(\frac{1}{d_2}+\frac{1}{d_3}+\frac{1}{d_4}\right)$.

**Lemma 1:** For any valid sequence, $a_n$ must be even for all $n \ge 1$.
*   **Proof:** Suppose $a_k$ is an odd integer for some $k \ge 1$. All its divisors are odd, so its three largest proper divisors are odd. Their sum, $a_{k+1} = f(a_k)$, is also odd. Therefore, if any term is odd, all subsequent terms are odd.
    Let $N$ be an odd integer with $d(N) \ge 4$. Its smallest divisors greater than 1 are $d_2, d_3, d_4$. Since $N$ is odd, all its divisors are odd. The smallest possible value for $d_2$ is 3. Thus, $d_2 \ge 3$. Since $d_3 > d_2$, $d_3 \ge 5$. Similarly, $d_4 > d_3$, so $d_4 \ge 7$.
    The sum $\frac{1}{d_2}+\frac{1}{d_3}+\frac{1}{d_4}$ is maximized when $d_2, d_3, d_4$ are minimized. Thus, $\frac{1}{d_2}+\frac{1}{d_3}+\frac{1}{d_4} \le \frac{1}{3}+\frac{1}{5}+\frac{1}{7} = \frac{71}{105}$.
    Therefore, $f(N) \le \frac{71}{105}N < N$.
    So, if $a_k$ is odd, the sequence $(a_n)_{n \ge k}$ is a strictly decreasing sequence of positive integers. Such a sequence must terminate, meaning it must produce a term which does not satisfy $d(a_n) \ge 4$. Therefore, an infinite sequence is not possible if any term is odd.

**Lemma 2:** For any valid sequence, $a_n$ must be divisible by 3 for all $n \ge 1$.
*   **Proof:** Suppose there is a term $a_k$ not divisible by 3. By Lemma 1, $a_k$ is even.
    Let $a_m$ be any term with $v_3(a_m)=0$. Its smallest divisor is $d_2=2$. Its divisors $d_3, d_4$ are not multiples of 3. Thus $d_3 \ge 4$ and $d_4 \ge 5$.
    $a_{m+1} = a_m(\frac{1}{2}+\frac{1}{d_3}+\frac{1}{d_4}) \le a_m(\frac{1}{2}+\frac{1}{4}+\frac{1}{5}) = \frac{19}{20}a_m < a_m$.
    So, if a term is not divisible by 3, the next term is strictly smaller.

    Now, consider the sequence $(a_n)_{n \ge k}$.
    If for all $n \ge k$, $v_3(a_n)=0$, then the sequence is strictly decreasing. A strictly decreasing sequence of positive integers must terminate. This is a failure.
    Therefore, for the sequence to be infinite, there must be a first term $a_m$ (with $m \ge k$) such that $v_3(a_m)=0$ and $v_3(a_{m+1})>0$.
    For $a_{m+1}$ to gain a factor of 3, the numerator of the fraction in $a_{m+1} = a_m(\frac{1}{d_2}+\frac{1}{d_3}+\frac{1}{d_4})$ must be divisible by 3. With $d_2=2$, this means $d_3d_4+2d_4+2d_3 \equiv 0 \pmod 3$, which implies $(d_3-1)(d_4-1) \equiv 1 \pmod 3$. This holds if and only if $d_3 \equiv 2 \pmod 3$ and $d_4 \equiv 2 \pmod 3$.
    If $v_2(a_m) \ge 2$, then $d_3=4 \equiv 1 \pmod 3$. The condition is not met.
    So, for the transition to happen, the term $a_m$ must have $v_2(a_m)=1$.
    In this case, $d_3=p$, the smallest odd prime factor of $a_m$. We need $p \equiv 2 \pmod 3$.
    $d_4$ is the smallest divisor of $a_m$ greater than $p$. We also need $d_4 \equiv 2 \pmod 3$.
    Let the prime factorization of $a_m$ be $2 \cdot p^{e_p} \cdot q^{e_q} \cdots$, where $p < q < \dots$ are odd primes.
    The candidates for $d_4$ are the smallest divisors of $a_m$ greater than $p$. These are $p^2$ (if $e_p \ge 2$), $2p$, and $q$ (if $a_m$ has a second odd prime factor $q$). Any other divisor of $a_m$ greater than $p$ is larger than one of these three.
    We check the congruences modulo 3, given $p \equiv 2 \pmod 3$:
    - $p^2 \equiv 2^2 = 4 \equiv 1 \pmod 3$.
    - $2p \equiv 2(2) = 4 \equiv 1 \pmod 3$.
    So neither $p^2$ nor $2p$ can be $d_4$.
    This implies that $d_4$ must be $q$, the second smallest odd prime factor of $a_m$. For this to be the case, we must have $q < p^2$ and $q < 2p$. And for the condition to be met, we must have $q \equiv 2 \pmod 3$.
    So, if a term $a_m$ with $v_2=1, v_3=0$ gains a factor of 3, its smallest two odd prime factors, $p$ and $q$, must both be congruent to $2 \pmod 3$.
    In this case, $d_3=p$ and $d_4=q$. Both are odd primes.
    Let's check $v_2(a_{m+1}) = v_2(a_m) + v_2(pq+2(p+q)) - v_2(2pq)$.
    We have $v_2(a_m)=1$. Since $p, q$ are odd, $pq$ is odd. $p+q$ is a sum of two odd numbers, so it's even. $2(p+q)$ is a multiple of 4. So $pq+2(p+q)$ is odd + (multiple of 4), which is odd. Thus $v_2(pq+2(p+q)) = 0$. Also, $v_2(2pq) = 1$ since $p,q$ are odd.
    $v_2(a_{m+1}) = 1 + 0 - 1 = 0$.
    So $a_{m+1}$ is odd. By Lemma 1, this leads to failure.
    In summary, if any term is not divisible by 3, the sequence must fail.

**Lemma 3:** For any valid sequence, no term $a_n$ can be divisible by 5.
*   **Proof:** By Lemmas 1 and 2, any term $a_n$ must be divisible by 2 and 3.
    First, suppose $v_5(a_k)=0$ for some $k \ge 1$. We show that $v_5(a_{k+1})=0$.
    The smallest divisors of $a_k$ are $d_2=2, d_3=3$. $d_4$ is the smallest divisor of $a_k$ greater than 3. Let $p$ be the smallest prime factor of $a_k$ other than 2 or 3. Since $v_5(a_k)=0$, $p \ge 7$. The candidates for $d_4$ are divisors of $a_k$ smaller than $p$, which can only be composed of primes 2 and 3. The smallest such divisor greater than 3 is 4.
    - If $v_2(a_k) \ge 2$, then 4 is a divisor of $a_k$. Since $3 < 4 < p$, we have $d_4=4$.
    - If $v_2(a_k) = 1$, then 4 is not a divisor of $a_k$. The smallest divisor greater than 3 must be $2 \cdot 3 = 6$ or $3^2=9$. Since $6 < 9$ and $6 < p$, $d_4=6$.
    In either case, $d_4$ is not a multiple of 5.
    If $d_4=4$, $a_{k+1} = \frac{13}{12}a_k$. If $d_4=6$, $a_{k+1}=a_k$. Neither operation introduces a factor of 5. So if $v_5(a_k)=0$, then $v_5(a_{k+1})=0$.

    Now, assume for contradiction that some term is divisible by 5. Let $a_n$ be such a term.
    If $v_2(a_n)=1$, then $d_2=2, d_3=3$. Since $v_5(a_n)\ge 1$, 5 is a divisor. $d_4=\min(6,5)=5$.
    Then $a_{n+1} = a_n\left(\frac{1}{2}+\frac{1}{3}+\frac{1}{5}\right) = \frac{31}{30}a_n$.
    $v_2(a_{n+1}) = v_2(a_n) - v_2(30) = 1-1=0$. So $a_{n+1}$ is odd, which fails by Lemma 1.
    Therefore, for a sequence with a term divisible by 5 to be valid, every term $a_n$ must satisfy $v_2(a_n) \ge 2$.
    This implies $d_4=4$ for all $n$, so $a_{n+1} = \frac{13}{12}a_n$. This means $v_2(a_{n+1}) = v_2(a_n)-2$ and $v_3(a_{n+1}) = v_3(a_n)-1$. This cannot continue indefinitely. The sequence must eventually produce a term $a_m$ with $v_2(a_m)<2$ or $v_3(a_m)<1$. If $v_2(a_m)=1$, the next term is odd. If $v_2(a_m)=0$, the term is odd. If $v_3(a_m)=0$, the term is not divisible by 3. All these cases lead to failure.

**Main Analysis**
From the lemmas, any term $a_n$ in a valid sequence must be of the form $N=2^a 3^b M$, where $a,b \ge 1$ and all prime factors of $M$ are $\ge 7$. We analyze the sequence based on the value of $a=v_2(N)$.

**Case 1: $v_2(N)=1$**
Let $N = 2^1 3^b M$ with $b \ge 1$ and prime factors of $M$ being at least 7. The smallest divisors of $N$ are $d_1=1, d_2=2, d_3=3$. The next smallest divisor is $d_4=\min(2^2, 2\cdot3, 3^2, p)$, where $p$ is the smallest prime factor of $M$. Since $v_2(N)=1$, $4$ is not a divisor. $d_4=\min(6,9,p)$. As $p \ge 7$, $d_4=6$.
Then $f(N) = N\left(\frac{1}{2}+\frac{1}{3}+\frac{1}{6}\right) = N(1) = N$.
Such numbers are fixed points. For $a_1$ to be one of these values, we must check that $d(a_1) \ge 4$.
$d(a_1) = d(2^1 3^b M) = 2(b+1)d(M)$. Since $b \ge 1$, $b+1 \ge 2$. $d(M) \ge 1$.
So $d(a_1) \ge 2(2)(1) = 4$. The condition is satisfied.
Thus, any integer $N=2^1 3^b M$ with $b \ge 1$ and prime factors of $M$ being at least 7 is a possible value for $a_1$.

**Case 2: $v_2(N) \ge 2$**
Let $N = 2^a 3^b M$ with $a \ge 2, b \ge 1$ and prime factors of $M$ being at least 7. The smallest divisors are $d_1=1, d_2=2, d_3=3, d_4=4$.
Then $f(N) = N\left(\frac{1}{2}+\frac{1}{3}+\frac{1}{4}\right) = \frac{13}{12}N$.
Let $a_1$ be such a number. The sequence starts with $a_{s+1} = \frac{13}{12}a_s$ as long as $v_2(a_s) \ge 2$.
This recurrence implies $v_2(a_{s+1})=v_2(a_s)-2$ and $v_3(a_{s+1})=v_3(a_s)-1$. For the sequence to be infinite, it must transition to a fixed point, which requires an iterate $a_k$ to have $v_2(a_k)=1$.
Let $v_2(a_1)=a$. The sequence of 2-adic valuations is $a, a-2, a-4, \dots$. For this sequence to attain the value 1, $a$ must be odd. If $a$ were even, it would eventually become 0, making the term odd and causing failure. So, $a$ must be an odd integer, $a \ge 3$.
The transition to $v_2=1$ occurs at step $s_0 = (a-1)/2$. The term is $a_{s_0+1} = a_{(a+1)/2}$.
$a_{s+1} = \left(\frac{13}{12}\right)^s a_1 = 2^{a-2s} 3^{b-s} 13^s M$.
For $s=s_0=(a-1)/2$, we get $v_2(a_{s_0+1}) = a-2\frac{a-1}{2}=1$.
For this term to be a fixed point, its 3-adic valuation must be at least 1.
$v_3(a_{s_0+1}) = b-s_0 = b-\frac{a-1}{2}$. We need $b-\frac{a-1}{2} \ge 1 \iff b \ge 1+\frac{a-1}{2} = \frac{a+1}{2}$.
This condition also ensures $v_3(a_s) \ge 1$ for all $s \le s_0$.

Finally, we verify that all terms in such a sequence satisfy $d(a_n) \ge 4$.
For $s \in \{0, 1, \dots, \frac{a-3}{2}\}$, the term is $a_{s+1} = 2^{a-2s} 3^{b-s} 13^s M$.
$v_2(a_{s+1}) = a-2s \ge a-2(\frac{a-3}{2}) = 3$.
$v_3(a_{s+1}) = b-s \ge b-\frac{a-3}{2} \ge \frac{a+1}{2}-\frac{a-3}{2} = 2$.
So $d(a_{s+1}) = (v_2(a_{s+1})+1)(v_3(a_{s+1})+1)d(13^s M) \ge (3+1)(2+1)(1) = 12 \ge 4$.
The term $a_{(a+1)/2}$ is the fixed point $2^1 \cdot 3^{b-(a-1)/2} \cdot 13^{(a-1)/2} M$.
$v_2=1$, $v_3 = b-\frac{a-1}{2} \ge 1$.
$d(a_{(a+1)/2}) = (1+1)(b-\frac{a-1}{2}+1)d(13^{(a-1)/2}M) \ge 2(1+1)(1) = 4$.
All terms satisfy the condition.

**Conclusion**
The possible values for $a_1$ are integers $N=2^a 3^b M$ where $a,b \ge 1$ and prime factors of $M$ are $\ge 7$, satisfying one of:
1.  $a=1, b \ge 1$ (fixed points).
2.  $a$ is an odd integer, $a \ge 3$, and $b \ge \frac{a+1}{2}$ (converging to a fixed point).

\subsubsection{Solution by Grok-4}

Let \(f(m)\) denote the sum of the three largest proper divisors of a positive integer \(m\) with at least three proper divisors (undefined otherwise). The problem requires finding all positive integers \(a_1\) such that the sequence defined by \(a_{n+1} = f(a_n)\) is infinite, i.e., every \(a_n\) has at least three proper divisors.

First, characterize when \(f(m) = m\), \(f(m) > m\), and \(f(m) < m\).

Let \(S\) be the set of positive integers divisible by 6 but not by 4 or 5.

For \(n \in S\), \(n\) is divisible by 2 and 3 with \(v_2(n) = 1\) and not divisible by 5. The proper divisors include \(n/2\), \(n/3\), \(n/6\), all integers. To show these are the three largest, note that any proper divisor is \(d = n/k\) for some integer \(k \geq 2\) dividing \(n\) with \(d < n\). The possible \(d\) correspond to the largest being \(n/2\) (for \(k=2\)), next \(n/3\) (for \(k=3\)), next \(n/6\) (for \(k=6\)). There is no integer \(k\) with \(2 < k < 3\), so no divisor between \(n/3\) and \(n/2\). For \(3 < k < 6\), the integers are 4 and 5, but \(n\) not divisible by 4 or 5, so \(n/4\) and \(n/5\) are not integers. For any \(k \geq 7\) (prime or composite), \(n/k \leq n/7 < n/6\) (since \(7 > 6\)). For composite \(k\) between 6 and 7 (none) or larger, similarly \(n/k < n/6\). Thus, no proper divisor exceeds \(n/6\) except \(n/2\) and \(n/3\), so the three largest proper divisors are \(n/2 > n/3 > n/6\), and \(f(n) = n/2 + n/3 + n/6 = n(1/2 + 1/3 + 1/6) = n\).

Every \(n \in S\) has at least three proper divisors, since \(n\) is divisible by distinct primes 2 and 3, so the number of divisors is at least \((1+1)(1+1) = 4\), hence at least three proper divisors. (For \(n = 6\), divisors are 1, 2, 3, 6; proper 1, 2, 3.)

Now, show \(S\) consists exactly of the fixed points. Suppose \(f(m) = m\).

- If \(m\) is odd, the largest proper divisor is at most \(m/3\) (since not even, no \(m/2\)), and the next two at most \(m/5\), \(m/7\) (or smaller if fewer large divisors), summing to at most \(m(1/3 + 1/5 + 1/7) = 71m/105 < m\).
- If \(m\) is even but not divisible by 3, the three largest are at most \(m/2\), \(m/4\) (if divisible by 4), \(m/5\) (or smaller), summing to at most \(m(1/2 + 1/4 + 1/5) = 19m/20 < m\).
- If \(m\) is divisible by 6 and by 4 (hence by 12), the possible proper divisors are \(m/k\) for integers \(k \geq 2\) dividing \(m\). The largest is \(m/2\) (\(k=2\)), next \(m/3\) (\(k=3\)), next \(m/4\) (\(k=4\), integer since divisible by 4). There is no integer \(k\) with \(3 < k < 4\), so no divisor between \(m/4\) and \(m/3\). For \(k \geq 5\), \(m/k \leq m/5 < m/4\) (since \(5 > 4\)). For any composite \(k \geq 6\), \(m/k \leq m/6 < m/4\) (since \(6 > 4\)). Thus, the three largest are \(m/2 > m/3 > m/4\), summing to \(13m/12 > m\).
- If \(m\) is divisible by 6 and by 5 but not by 4 (so \(v_2(m) = 1\)), the possible proper divisors are \(m/k\) for \(k \geq 2\) dividing \(m\). The largest is \(m/2\) (\(k=2\)), next \(m/3\) (\(k=3\)), next \(m/5\) (\(k=5\)). There is no integer \(k\) with \(3 < k < 5\) except 4, but not divisible by 4, so \(m/4\) not integer. For \(k \geq 6\), \(m/k \leq m/6 < m/5\) (since \(6 > 5\)). For composites like \(k=10\) (if divisible by 10, which it is since by 2 and 5), \(m/10 = 0.1m < m/5 = 0.2m\). Similarly for others. Thus, the three largest are \(m/2 > m/3 > m/5\), summing to \(m(1/2 + 1/3 + 1/5) = 31m/30 > m\).

Thus, \(f(m) = m\) iff \(m \in S\); \(f(m) > m\) only in the cases divisible by 12 or \(v_2(m) = 1\) and divisible by 15 (equivalent to divisible by 30 with \(v_2 = 1\)); and \(f(m) < m\) otherwise.

Next, prove that if \(m\) is not divisible by 6, then \(f(m) < m\) (already shown) and \(f(m)\) is not divisible by 6.

- If \(m\) is odd, all proper divisors are odd, so \(f(m)\) is sum of three odds, hence odd, not divisible by 2, hence not by 6.
- If \(m\) is even but not divisible by 3, write \(m = 2^k s\) with \(s\) odd not divisible by 3, \(k \geq 1\).

For \(k \geq 2\), the two largest proper divisors are \(m/2 = 2^{k-1} s\) and \(m/4 = 2^{k-2} s\) (since any \(m/p\) for odd prime \(p \geq 5\) dividing \(s\) satisfies \(m/p \leq m/5 < m/4\) as \(2^k /5 < 2^{k-2}\) for \(k \geq 2\)). The sum of these two is congruent to 0 modulo 3: the coefficients \(2^{k-1}\) and \(2^{k-2}\) are congruent to 1 and 2 modulo 3 (in some order, as \(2 \equiv -1 \pmod{3}\) and powers alternate), so the sum is \((1 + 2)s \equiv 0 \pmod{3}\). The third largest proper divisor is some \(d < m/4\) with \(d \mid m\). Since \(m \not\equiv 0 \pmod{3}\), no divisor \(d\) of \(m\) is divisible by 3 (else \(3 \mid m\)), so \(d \not\equiv 0 \pmod{3}\), i.e., \(d \equiv 1\) or \(2 \pmod{3}\). Thus, \(f(m) \equiv 0 + (1 \text{ or } 2) \not\equiv 0 \pmod{3}\). Also, \(f(m)\) is even (sum includes two evens, and the third is at most \(m/5\) or smaller, but regardless, even + even + anything = even), so even but not divisible by 3 implies not divisible by 6.

For \(k = 1\), \(m = 2s\), \(s\) odd not divisible by 3. Let \(p\) be the minimal prime dividing \(s\) (\(\geq 5\)), \(u = s/p\) (largest proper divisor of \(s\)). The largest proper divisor of \(m\) is \(s\) (odd). The largest even proper divisor is \(2u\). Since \(p \geq 5 > 2\), \(2u = 2(s/p) \leq 2(s/5) = (2/5)s < s\). The second largest is \(2u\) (even, as \(2u > u = s/p \leq s/5 < (2/5)s \leq 2u\)). The third largest is the maximum of: (i) the largest odd proper divisor less than \(2u\) (includes \(u\) and smaller divisors of \(s\)); (ii) the largest even proper divisor less than \(2u\) (of form \(2v\) with \(v \mid s\), \(v < u\)). Thus, if there exists \(v \mid s\) with \(u/2 < v < u\) (so \(2v > u\) but \(2v < 2u\)), then the max such \(2v\) is even and \( > u\), and if it is the overall max below \(2u\), the third is even; otherwise, if no such \(v\) or if \(u\) exceeds all such \(2v\), the third is \(u\) (odd). These exhaust the possibilities for the third largest, as all proper divisors are either odd (divisors of \(s\)) or even (\(2\) times divisors of \(s\)).

Case 1: Third = \(u\) (odd). Then \(f(m) = s + 2u + u = s + 3u \equiv s \pmod{3}\) (since \(3u \equiv 0\)), and \(s \not\equiv 0 \pmod{3}\), so not divisible by 3. Also even (odd + even + odd = even), hence not by 6.

Case 2: Third even (say \(e = 2v > u\)). Then \(f(m) = s + 2u + e\) = odd + even + even = odd, not divisible by 2, hence not by 6.

There are no cycles of length greater than 1. Suppose there is a cycle of length \(k > 1\). If not all terms divisible by 6, some \(m_i\) not divisible by 6, then \(m_{i+1} = f(m_i) < m_i\) and not divisible by 6, continuing strictly decreasing, but positive integers have no infinite descending chains (well-ordering), so finite, contradiction to infinite cycle. If all divisible by 6, then for each \(m_i\), \(f(m_i) \geq m_i\) (with equality iff in \(S\)). For cycle length >1, must have some \(f(m_i) > m_i\), but then the cycle would require summing increases and decreases, impossible unless all equal (contradiction).

The only valid infinite sequences are those eventually reaching and staying at some fixed point in \(S\).

The possible \(a_1\) are those whose forward orbit under \(f\) reaches \(S\) without hitting numbers with fewer than three proper divisors (i.e., 1, primes, or prime powers \(p^k\) with \(k \leq 2\), which have at most two proper divisors).

From the classification, orbits reaching \(S\) are the increasing chains under the cases where \(f(m) > m\). Chains using \(f(m) = (31/30)m\) (i.e., starting at \(m\) with \(v_2(m)=1\), divisible by 15) yield \(f(m)\) odd \(v_2=0\), then strictly decreasing (since odd implies \(f < m\), not divisible by 6), staying odd (sum of odds is odd). Such decreasing sequences cannot be infinite in positive integers (no infinite descending chains), so must eventually hit a number with $<3$ proper divisors, making the sequence finite. There are no mixed chains: backward from \(S\) or \(13/12\)-chains yields even \(v_2 \geq 2\), but \(31/30\) requires \(v_2=1\) (odd), contradiction. Forward mixing (applying \(31/30\) after \(13/12\)-steps) requires \(v_2=1\) intermediate, but \(13/12\)-chains have odd \(v_2 >1\), and applying \(31/30\) yields $v_2=0$, then decreasing without returning to $v_2>0$.

Unmatched \(13/12\)-chains (not reaching \(S\)) hit forbidden numbers. For \(m\) divisible by 12 (\(v_2(m) = a \geq 2\), \(v_3(m) = b \geq 1\)), \(f(m) = (13/12)m\) only if 13 divides the result, but no: \(f\) is always defined as the sum, and for div by 12, it is \(13/12\)m regardless of 13 (but for backward, 13 matters). Forward: each step subtracts 2 from $v_2$ and 1 from $v_3$. If a even, after $a/2$ steps, $v_2=0$ (even - 2 repeatedly stays even), $v_3 = b - a/2$; if $v_3$ drops to 0 earlier ($b < j$ for some $j \le a/2$), then that term not div by 3, hence not by 6, strictly decreasing thereafter (no infinite descending chains, so hits $<3$ proper). If $v_3 \ge a/2 at v_2=0$, then reaches odd \(v_2=0\), possibly div by $3$, but then strictly decreasing, staying odd, hits $<3$ proper. Never reaches $v_2=1$ (odd) from even $v_2$. If a odd, after $(a-1)/2$ steps, $v_2=1, v_3 = b - (a-1)/2$; if this $<1$ (i.e., $a - 2b \ge 0$, noting $a-2b$ even or odd but condition for not reaching positive $v_3$ at $v_2=1$), or if $v_3=0$ earlier, then not div by 6, strictly decreasing, hits $<3$ proper.

The valid chains are: for \(f \in S\) with \(v_{13}(f) = c \geq 0\), write \(f = 2^1 \cdot 3^b \cdot 13^c \cdot h\) (\(b \geq 1\), \(h\) coprime to 2,3,5,13, primes \(\geq 7\)). For \(0 \leq j \leq c\), the number after \(j\) backward steps is \(n = 2^{1+2j} \cdot 3^{b+j} \cdot 13^{c-j} \cdot h\) (each step multiplies by $12/13$, adding $2$ to $v_2$, $1$ to $v_3$, subtracting 1 from $v_{13}$). Forward, these increase to \(f\) in \(S\) (via \(f(n) = (13/12)n\)), with all terms divisible by 6, $v_2$ odd positive, $v_3$ positive, at least four divisors (hence $\ge3$ proper).

Such \(n\) has $v_2(n) = a = 1 + 2j$ odd positive, $v_3(n) = b' = b + j$ positive, $a - 2b' = 1 + 2j - 2(b + j) = 1 - 2b$ negative odd ($b \ge1$ implies $\ge -1$, odd), not div by $5$, other primes $\ge7$.

Conversely, any such \(n\) arises this way: set \(j = (a-1)/2 \geq 0\) integer (a odd positive), \(b = b' - j = (1 - (a - 2b'))/2 > 0\) integer ($a - 2b'$ negative odd implies $1 - (a - 2b')$ positive even), factor out \(2^a 3^{b'}\), remaining \(13^{c'} \cdot h'\) ($c' \ge0$, $h'$ primes $\ge7$), set $c = c' + j, h = h'$, then \(f = 2^1 3^b 13^c h \in S\).

Thus, these are all possible \(a_1\).

all positive integers n divisible by 6 but not by 5, with odd positive 2-adic valuation a, positive 3-adic valuation b such that a-2b is negative odd, and all other prime factors $\geq7$

\subsubsection{Solution by GPT-5}

Let $f(N)$ denote the sum of the three largest proper divisors of $N$. Writing the positive divisors of $N$ in increasing order as $1=d_1<d_2\le d_3\le d_4\le \cdots \le d_{\tau(N)}=N$, the three largest proper divisors are $N/d_2$, $N/d_3$, $N/d_4$, hence
$$f(N)=N\left(\frac{1}{d_2}+\frac{1}{d_3}+\frac{1}{d_4}\right).$$

Lemma 1. If $6\mid N$, then
$$ f(N)= \begin{cases} 13N/12, & \text{if } 4\mid N; \\ 31N/30, & \text{if } 4\nmid N \text{ and } 5\mid N; \\ N, & \text{if } 4\nmid N \text{ and } 5\nmid N. \end{cases} $$
In particular, for $6\mid N$ we have $f(N)=N$ if and only if $4\nmid N$ and $5\nmid N$.

Proof.
When $6\mid N$, we have $d_2=2$ and $d_3=3$. The next divisor $d_4$ is the least of $4,5,6$ that divides $N$. Therefore
$$f(N)=N\left(\frac12+\frac13+\frac1{d_4}\right),$$
which yields the three stated values. \qed

Lemma 2. If $2\mid N$ and $3\nmid N$, then
$$f(N)\le \left(\frac12+\frac14+\frac15\right)N=\frac{19}{20}N.$$

Proof.
Let $1=d_1<d_2\le d_3\le d_4$ be the first four divisors. Since $2\mid N$, $d_2=2$.

Case 1: $4\mid N$. Then $d_3=4$. Because $3\nmid N$, the next divisor $d_4\ge 5$. Thus
$$f(N)=\frac{N}{2}+\frac{N}{4}+\frac{N}{d_4}\le \left(\frac12+\frac14+\frac15\right)N.$$

Case 2: $4\nmid N$. Then $v_2(N)=1$. Since $3\nmid N$, the least odd divisor is at least $5$, so $d_3\ge 5$. The next divisor $d_4$ is either an odd divisor $\ge 7$, or of the form $2p$ with odd $p\ge 5$, hence $d_4\ge \min\{7,10\}=7$. Therefore
$$f(N)\le \left(\frac12+\frac15+\frac17\right)N=\frac{59}{70}N<\frac{19}{20}N.$$ \qed

Lemma 3. Let $N$ be odd and suppose $N$ has at least three proper divisors. Then
$$f(N)\le \left(\frac13+\frac15+\frac17\right)N=\frac{71}{105}N<N,$$
and $f(N)$ is odd.

Proof.
All divisors are odd. The smallest divisor greater than $1$ is at least $3$, so $d_2\ge 3$. The next $d_3$ is at least $5$ (either the next odd prime or a square $\ge 9$). Among $\{3,5\}$ at most two appear among $d_2,d_3,d_4$, hence $d_4\ge 7$. Thus
$$\frac{1}{d_2}+\frac{1}{d_3}+\frac{1}{d_4}\le \frac13+\frac15+\frac17,$$
giving the bound on $f(N)$. Each of $N/d_2$, $N/d_3$, $N/d_4$ is odd, so $f(N)$ is odd. \qed

Lemma 4. Let $N$ be even with $3\nmid N$.
(i) If $4\mid N$, then $3\nmid f(N)$.
(ii) If $v_2(N)=1$, then either $3\nmid f(N)$ or $f(N)$ is odd.

Proof.
Write $N=2^tM$ with $t\ge 1$, $M$ odd, $3\nmid M$.

(i) If $t\ge 2$, then $d_2=2$, $d_3=4$, and $d_4$ is either $8$ or the least odd divisor $p\ge 5$ of $M$. Modulo $3$, any $d$ coprime to $3$ satisfies $d^{-1}\equiv d\pmod{3}$. Hence
$$\frac{f(N)}{N}\equiv \frac12+\frac14+\frac1{d_4}\equiv (-1)+1+(\pm 1)\in\{\pm 1\}\pmod{3},$$
so $3\nmid f(N)$.

(ii) If $t=1$, then $d_2=2$, and $d_3$ is the least odd divisor $p\ge 5$ of $M$. The next divisor $d_4$ is either $2p$ or an odd divisor $q$ of $M$. If $d_4=2p$, then
$$\frac{f(N)}{N}=\frac12+\frac1p+\frac1{2p}\equiv (-1)+\varepsilon+(-\varepsilon)\equiv -1\not\equiv 0\pmod{3},$$
with $\varepsilon\in\{\pm 1\}$ the residue of $p$ modulo $3$, so $3\nmid f(N)$. If $d_4$ is odd, then $N/2$ is odd while $N/d_3$ and $N/d_4$ are even, hence $f(N)$ is odd. \qed

Corollary A (no odd term).
If some $a_n$ were odd, then by Lemma 3, $a_{n+1}$ is odd and $a_{n+1}<a_n$, so $a_n,a_{n+1},\dots$ would be a strictly decreasing infinite sequence of positive integers, a contradiction. Hence no odd term occurs.

Corollary B (no even, $3$-free term).
Suppose some $a_k$ is even with $3\nmid a_k$. Let $a_k,\dots,a_\ell$ be the maximal contiguous block of such terms. By Lemma 2, for each $j$ in this block,
$$a_{j+1}\le \frac{19}{20}a_j,$$ so the block is finite. At its end $a_{\ell+1}$ is not even and $3$-free; by Lemma 4 it cannot be a multiple of $6$, hence it is odd, contradicting Corollary A. Therefore no even, $3$-free term occurs.

Consequently, every $a_n$ is divisible by $6$.

Proposition C’ (no term is divisible by $5$).
No term $a_n$ is divisible by $5$.

Proof.
Assume $5\mid a_n$ for some $n$.

If $3\nmid a_n$, then $a_n$ is either odd or even and $3$-free, both impossible (Corollaries A and B). If $2\nmid a_n$, then $a_n$ is odd, again impossible. Thus $6\mid a_n$.

Write $a_n=2^s3^ru$ with $s\ge 1$, $r\ge 1$, $5\mid u$, and $\gcd(u,6)=1$.

- If $s=1$ (i.e., $4\nmid a_n$), then by Lemma 1,
  $a_{n+1}=f(a_n)=\frac{31}{30}a_n$, whence $v_2(a_{n+1})=v_2(a_n)-1=0$, so $a_{n+1}$ is odd, contradicting Corollary A.

- If $s\ge 2$ (i.e., $12\mid a_n$), perform the iteration $N_{i+1}=\frac{13}{12}N_i$ as long as $12\mid N_i$. Let $N_0=a_n$ and let $k\ge 1$ be maximal with this property; put $N:=N_k$. Each step reduces $v_2$ by $2$ and $v_3$ by $1$ and preserves $5\mid N_i$. Maximality implies at $N$ at least one of $v_3(N)=0$ or $v_2(N)\le 1$ holds. The possible terminal cases are exactly:

  (A) $v_3(N)=0$ and $v_2(N)\ge 2$ (even, $3$-free), contradicting Corollary B.

  (A') $v_3(N)=0$ and $v_2(N)=1$ (even, $3$-free), also contradicting Corollary B.

  (B) $v_2(N)=1$ and $v_3(N)\ge 1$: then $4\nmid N$ and $5\mid N$, so by Lemma 1,
      $f(N)=\frac{31}{30}N$, which is odd (since $v_2(f(N))=0$), contradicting Corollary A.

  (C) $v_2(N)=0$ (so $N$ is odd), contradicting Corollary A.

  These cases are mutually exclusive and exhaustive. In every case we reach a contradiction.

Therefore $5\nmid a_n$ for all $n$. \qed

Proposition E (structure of steps).
For every $n$, either $a_{n+1}=a_n$ or $a_{n+1}=\frac{13}{12}a_n$, and the latter occurs precisely when $12\mid a_n$.

Proof.
By Corollaries A–B, $6\mid a_n$, and by Proposition C’, $5\nmid a_n$. Lemma 1 then yields:
if $4\nmid a_n$, $f(a_n)=a_n$; if $4\mid a_n$, $f(a_n)=\frac{13}{12}a_n$. The latter is equivalent to $12\mid a_n$. \qed

Proposition H (finiteness of non-fixed steps; stabilization).
Only finitely many indices $n$ satisfy $a_{n+1}=\frac{13}{12}a_n$. Consequently, the sequence eventually stabilizes at a fixed point $N$ with $6\mid N$, $4\nmid N$, $5\nmid N$.

Proof.
By Corollaries A–B, every term is a multiple of $6$, so $v_3(a_n)\ge 1$. By Proposition C’, $5\nmid a_n$ for all $n$. By Proposition E, the only non-fixed step is multiplication by $\frac{13}{12}$, which reduces $v_2$ by $2$ and $v_3$ by $1$. Infinitely many such steps would force $v_3$ to become $0$ along some tail, impossible. Hence only finitely many non-fixed steps occur. After the last such step, we have a term $N$ with $6\mid N$, $4\nmid N$, $5\nmid N$, and by Lemma 1 it is fixed: $f(N)=N$. \qed

Proposition F (necessary conditions on $a_1$).
Let $T$ be the number of indices $j$ with $a_{j+1}=\frac{13}{12}a_j$. Then
$$v_2(a_1)=2T+1, \quad v_3(a_1)\ge T+1, \quad \text{and} \quad 5\nmid a_1.$$

Proof.
By Proposition H the sequence stabilizes at a fixed multiple of $6$ not divisible by $5$. By Lemma 1 and Proposition E, a fixed multiple of $6$ not divisible by $5$ must satisfy $4\nmid N$, hence $v_2(N)=1$, and has $v_3(N)\ge 1$. Each non-fixed step multiplies by $\frac{13}{12}$, decreasing $v_2$ by $2$ and $v_3$ by $1$. Tracing back $T$ such steps gives
$$v_2(a_1)=1+2T=2T+1 \quad \text{and} \quad v_3(a_1)\ge 1+T=T+1.$$ Finally $5\nmid a_1$ by Proposition C’. \qed

Equivalently, $a_1$ must be of the form $a_1=2^{\,2t+1}3^{\,b}m$ with integers $t\ge 0$, $b\ge t+1$, and $\gcd(m,30)=1$.

Proposition G (sufficient conditions on $a_1$).
Assume $a_1=2^{\,2t+1}3^{\,b}m$ with $t\ge 0$, $b\ge t+1$, and $\gcd(m,30)=1$. Then the sequence is valid and evolves as follows:
- For $j=1,2,\dots,t$, $a_{j+1}=\frac{13}{12}a_j$.
- The term $a_{t+1}$ satisfies $6\mid a_{t+1}$, $4\nmid a_{t+1}$, $5\nmid a_{t+1}$, hence $a_{t+2}=a_{t+1}$.
- Every term is a multiple of $6$, and thus has at least three proper divisors.

Proof.
For $1\le j\le t$,
$$v_2(a_j)=2t+1-2(j-1)\ge 3 \quad \text{and} \quad v_3(a_j)=b-(j-1)\ge t+1-(j-1)\ge 2,$$
so $12\mid a_j$ and, by Proposition E, $a_{j+1}=\frac{13}{12}a_j$. After $t$ steps,
$$v_2(a_{t+1})=1 \quad \text{and} \quad v_3(a_{t+1})\ge 1.$$ Since $\gcd(m,30)=1$ and the multiplier $\frac{13}{12}$ introduces no factor $5$, we have $5\nmid a_{t+1}$. Lemma 1 gives $a_{t+2}=f(a_{t+1})=a_{t+1}$, and the sequence is constant thereafter. Finally, $6\mid a_j$ for all $j$, hence each term has proper divisors $1,2,3$. \qed

Conclusion.
Combining Propositions F and G yields the classification:
$a_1$ is valid if and only if $a_1=2^{\,2t+1}3^{\,b}m$ with $t\ge 0$, $b\ge t+1$, and $\gcd(m,30)=1$.
For such $a_1$, exactly $t$ steps of $a\mapsto \frac{13}{12}a$ occur, after which the sequence stabilizes at a fixed point $N$ with $6\mid N$, $4\nmid N$, $5\nmid N$.

\subsection{Problem 5}

\begin{problem}
Alice and Bazza are playing the \textit{inekoalaty} game, a two-player game whose rules depend on a positive real number $\lambda$ which is known to both players. On the $n$th turn of the game (starting with $n = 1$) the following happens:
\begin{itemize}
    \item If $n$ is odd, Alice chooses a nonnegative real number $x_n$ such that
    \[
    x_1 + x_2 + \cdots + x_n \le \lambda n.
    \]
    \item If $n$ is even, Bazza chooses a nonnegative real number $x_n$ such that 
    \[
    x_1^2 + x_2^2 + \cdots + x_n^2 \le n.
    \]
\end{itemize}

If a player cannot choose a suitable number $x_n$, the game ends and the other player wins. If the game goes forever, neither player wins. All chosen numbers are known to both players.

Determine all values of $\lambda$ for which Alice has a winning strategy and all those for which Bazza has a winning strategy.
\end{problem}

\subsubsection{Solution by Gemini 2.5 Pro}

Let $S_n = \sum_{i=1}^n x_i$ and $Q_n = \sum_{i=1}^n x_i^2$.
Alice plays on odd turns $n$, Bazza on even turns $n$.
Alice's move $x_n$ is valid if $S_{n-1}+x_n \le \lambda n$. If $S_{n-1} > \lambda n$, she loses.
Bazza's move $x_n$ is valid if $Q_{n-1}+x_n^2 \le n$. If $Q_{n-1} > n$, he loses.

**Case 1: $\lambda > \frac{\sqrt{2}}{2}$ (Alice has a winning strategy)**

Alice can devise a plan to win on a predetermined turn $2m-1$.
**Alice's Plan:**
1.  Alice chooses an integer $m$ large enough such that $\lambda > \frac{m\sqrt{2}}{2m-1}$. Such an $m$ exists because the function $g(m) = \frac{m\sqrt{2}}{2m-1}$ is strictly decreasing for $m \ge 1$ and $\lim_{m\to\infty} g(m) = \frac{\sqrt{2}}{2}$.
2.  For her turns $2k-1$ where $k=1, \dots, m-1$, Alice chooses $x_{2k-1}=0$.
3.  On turn $2m-1$, Alice will choose a value $x_{2m-1}$ that makes Bazza's next move impossible.

**Analysis of Alice's Plan:**
Alice wins on turn $2m-1$ if she can choose $x_{2m-1} \ge 0$ such that her move is valid and Bazza's next move is not. This is possible if and only if the interval of winning moves for $x_{2m-1}$, which is $(\sqrt{2m - Q_{2m-2}}, \lambda(2m-1) - S_{2m-2}]$, is non-empty. This requires the condition:
$$S_{2m-2} + \sqrt{2m - Q_{2m-2}} < \lambda(2m-1)$$
Bazza's goal is to prevent this. Given Alice's plan ($x_{2k-1}=0$ for $k<m$), Bazza controls the values $y_k = x_{2k}$ for $k=1, \dots, m-1$. These choices determine $S_{2m-2} = \sum_{k=1}^{m-1} y_k$ and $Q_{2m-2} = \sum_{k=1}^{m-1} y_k^2$. Bazza's best defense is to choose his moves to maximize the function $F = S_{2m-2} + \sqrt{2m - Q_{2m-2}}$.

**Lemma:** The maximum value of $F$ that Bazza can achieve is $m\sqrt{2}$.
**Proof:** Let $S = S_{2m-2}$ and $Q = Q_{2m-2}$. Bazza's moves are constrained by $\sum_{i=1}^j y_i^2 \le 2j$ for $j=1, \dots, m-1$. This implies $Q = \sum_{k=1}^{m-1} y_k^2 \le 2(m-1)$.
By the Cauchy-Schwarz inequality, $S^2 = (\sum_{k=1}^{m-1} y_k)^2 \le (m-1) \sum_{k=1}^{m-1} y_k^2 = (m-1)Q$. Thus, $S \le \sqrt{(m-1)Q}$.
So, $F \le \sqrt{(m-1)Q} + \sqrt{2m-Q}$. Let this upper bound be $h(Q)$. We maximize $h(Q)$ for $Q \in [0, 2(m-1)]$.
The derivative $h'(Q) = \frac{\sqrt{m-1}}{2\sqrt{Q}} - \frac{1}{2\sqrt{2m-Q}}$ is positive for $Q < 2(m-1)$. So $h(Q)$ is strictly increasing on its domain. The maximum is at $Q=2(m-1)$.
The maximum value of $h(Q)$ is $h(2(m-1)) = \sqrt{(m-1)2(m-1)} + \sqrt{2m-2(m-1)} = \sqrt{2}(m-1) + \sqrt{2} = m\sqrt{2}$.
This maximum is achieved when $Q=2(m-1)$ and the Cauchy-Schwarz inequality is an equality, which means all $y_k$ are equal. Let $y_k=c$. Then $Q=(m-1)c^2=2(m-1) \implies c=\sqrt{2}$. The sequence of moves $x_{2k}=\sqrt{2}$ for $k=1, \dots, m-1$ is valid for Bazza and it maximizes the defensive function $F$.

**Alice's Victory:**
Alice's strategy is guaranteed to work if her winning condition holds even against Bazza's best defense. This requires $\max(F) < \lambda(2m-1)$, which is $m\sqrt{2} < \lambda(2m-1)$, or $\lambda > \frac{m\sqrt{2}}{2m-1}$. By her initial choice of $m$, this condition is met. We must also check that Alice's moves $x_{2k-1}=0$ for $k<m$ are valid. This requires $S_{2k-2} \le \lambda(2k-1)$. Bazza's best defense maximizes $S_{2k-2}$ to $(k-1)\sqrt{2}$. The condition is $(k-1)\sqrt{2} \le \lambda(2k-1)$, or $\lambda \ge \frac{(k-1)\sqrt{2}}{2k-1}$. Since $\lambda > \frac{\sqrt{2}}{2}$ and $\frac{(k-1)\sqrt{2}}{2k-1}$ is an increasing function of $k$ with limit $\frac{\sqrt{2}}{2}$, this condition holds for all $k$. Thus, Alice has a winning strategy.

**Case 2: $\lambda < \frac{\sqrt{2}}{2}$ (Bazza has a winning strategy)**

Bazza's strategy is to always play $x_{2k} = \sqrt{2k - Q_{2k-1}}$ if possible. This sets $Q_{2k}=2k$. As shown in Case 1, Alice cannot win against this strategy because her winning condition $\lambda > \frac{m\sqrt{2}}{2m-1}$ can never be met if $\lambda < \frac{\sqrt{2}}{2}$. We now show that Bazza will win.

Alice loses on turn $2m-1$ if $S_{2m-2} > \lambda(2m-1)$. To survive, Alice must choose her moves to keep the sequence of sums $S_{2k-2}$ as small as possible for as long as possible.
With Bazza's strategy, Alice's choice of $x_{2k-1}$ (provided $x_{2k-1} \le \sqrt{2}$) determines Bazza's response $x_{2k}=\sqrt{2-x_{2k-1}^2}$. The sum grows by $C_k = x_{2k-1} + \sqrt{2-x_{2k-1}^2}$ over turns $2k-1$ and $2k$.
To minimize the sum $S_{2m-2} = \sum_{k=1}^{m-1} C_k$, Alice must choose each $x_{2k-1}$ to minimize $C_k$. The function $f(x)=x+\sqrt{2-x^2}$ on $[0, \sqrt{2}]$ has a minimum value of $\sqrt{2}$, achieved only at $x=0$ and $x=\sqrt{2}$. Any other choice would lead to a strictly larger sum $S_{2m-2}$ for all $m>k+1$, making survival strictly harder. Thus, an optimal survival strategy for Alice must consist only of moves $x_{2k-1} \in \{0, \sqrt{2}\}$.

Let's compare these two choices at turn $2k-1$. Suppose Alice has survived so far, with sum $S_{2k-2}$.
1.  If Alice chooses $x_{2k-1}=0$: This move is valid if $S_{2k-2} \le \lambda(2k-1)$. The resulting sum is $S_{2k} = S_{2k-2} + \sqrt{2}$.
2.  If Alice chooses $x_{2k-1}=\sqrt{2}$: This move is valid if $S_{2k-2}+\sqrt{2} \le \lambda(2k-1)$. The resulting sum is $S_{2k} = S_{2k-2} + \sqrt{2}$.

Both choices lead to the same future sums $S_{2j}$ for $j \ge k$, meaning the survival conditions for all subsequent turns are identical regardless of which of the two is chosen. However, the condition to be allowed to make the choice at turn $2k-1$ is strictly easier for $x_{2k-1}=0$. A strategy involving $x_{2k-1}=\sqrt{2}$ is only valid if the corresponding strategy with $x_{2k-1}=0$ is also valid, but the converse is not true. Therefore, the strategy of always choosing $x_{2k-1}=0$ is Alice's best hope for survival. If she cannot survive with this strategy, she cannot survive with any other.

We now analyze this specific line of play: Alice always plays $x_{2k-1}=0$, and Bazza responds with $x_{2k}=\sqrt{2}$.
1.  Let $h(k) = \frac{(k-1)\sqrt{2}}{2k-1}$. Since $h(k)$ is strictly increasing and approaches $\frac{\sqrt{2}}{2}$, and $\lambda < \frac{\sqrt{2}}{2}$, there exists a smallest integer $m \ge 2$ such that $\lambda < h(m)$.
2.  For any $k < m$, we have $\lambda \ge h(k)$. Alice's move $x_{2k-1}=0$ is valid, since $S_{2k-2}=(k-1)\sqrt{2}$ and the condition is $(k-1)\sqrt{2} \le \lambda(2k-1)$, which is equivalent to $\lambda \ge h(k)$.
3.  On turn $2m-1$, Alice has played according to her optimal survival strategy. The sum is $S_{2m-2}=(m-1)\sqrt{2}$. She must choose $x_{2m-1} \ge 0$ such that $(m-1)\sqrt{2} + x_{2m-1} \le \lambda(2m-1)$.
4.  By the choice of $m$, we have $\lambda < \frac{(m-1)\sqrt{2}}{2m-1}$, which is $\lambda(2m-1) < (m-1)\sqrt{2}$.
5.  The condition for Alice's move becomes $(m-1)\sqrt{2} + x_{2m-1} \le \lambda(2m-1) < (m-1)\sqrt{2}$. This implies $x_{2m-1} < 0$, which is impossible.
Alice cannot make a move, so Bazza wins.

**Case 3: $\lambda = \frac{\sqrt{2}}{2}$ (Draw)**

In this case, neither player has a winning strategy. A player has a winning strategy if they can force a win in a finite number of moves against any of the opponent's strategies.

**Alice does not have a winning strategy.**
To prove this, we show that Bazza has a defensive strategy that prevents Alice from ever winning. Let Bazza adopt the strategy of always choosing $x_{2k} = \sqrt{2k - Q_{2k-1}}$ (if possible).
Alice wins if she can play $x_{2m-1}$ on turn $2m-1$ such that $Q_{2m-1} > 2m$. With Bazza's strategy, $Q_{2m-2}=2(m-1)$ (assuming Alice has not won before). Alice's winning condition becomes $2(m-1)+x_{2m-1}^2 > 2m$, which simplifies to $x_{2m-1} > \sqrt{2}$.
To play such a move, her budget must allow it: $S_{2m-2} + x_{2m-1} \le \lambda(2m-1) = \frac{\sqrt{2}}{2}(2m-1)$.
This requires $S_{2m-2} + \sqrt{2} < \frac{\sqrt{2}}{2}(2m-1)$.
Against Bazza's strategy, $S_{2m-2} = \sum_{k=1}^{m-1} (x_{2k-1} + \sqrt{2-x_{2k-1}^2})$. The minimum value of each term is $\sqrt{2}$, so the minimum possible value of $S_{2m-2}$ is $(m-1)\sqrt{2}$.
Thus, for Alice to win, it must be that $(m-1)\sqrt{2} + \sqrt{2} < \frac{\sqrt{2}}{2}(2m-1)$, which simplifies to $m\sqrt{2} < \frac{\sqrt{2}}{2}(2m-1) \iff 2m < 2m-1 \iff 0 < -1$. This is impossible.
So, Alice can never satisfy her winning condition against this strategy from Bazza. Therefore, Alice does not have a winning strategy.

**Bazza does not have a winning strategy.**
To prove this, we show that Alice has a defensive strategy that prevents Bazza from ever winning. Let Alice adopt the strategy of always choosing $x_{2k-1}=0$.
Bazza wins if he can force $S_{2m-2} > \lambda(2m-1) = \frac{\sqrt{2}}{2}(2m-1)$ for some $m$.
With Alice's strategy, $S_{2m-2} = \sum_{k=1}^{m-1} x_{2k}$. Bazza's moves $x_{2k}$ are constrained by $Q_{2k} = \sum_{i=1}^k x_{2i}^2 \le 2k$.
To force a win, Bazza must maximize $S_{2m-2}$. As shown in the lemma in Case 1, the maximum possible value for $S_{2m-2}$ is $(m-1)\sqrt{2}$, achieved when Bazza plays $x_{2k}=\sqrt{2}$ for all $k<m$.
So, the largest sum Bazza can force is $S_{2m-2}=(m-1)\sqrt{2}$. Bazza wins if this sum exceeds Alice's budget limit, i.e., if $(m-1)\sqrt{2} > \frac{\sqrt{2}}{2}(2m-1)$. This simplifies to $2(m-1) > 2m-1 \iff 2m-2 > 2m-1 \iff -2 > -1$. This is impossible.
So, Bazza can never force Alice to lose. Therefore, Bazza does not have a winning strategy.

Since neither player can force a win, neither has a winning strategy. The game can continue indefinitely, for example with Alice always playing $x_{2k-1}=0$ and Bazza always playing $x_{2k}=\sqrt{2}$. This results in a draw.

\subsubsection{Solution by Grok-4}

We prove that Alice has a winning strategy if and only if $\lambda > \frac{\sqrt{2}}{2}$, Bazza has a winning strategy if and only if $\lambda < \frac{\sqrt{2}}{2}$, and neither has a winning strategy if $\lambda = \frac{\sqrt{2}}{2}$.

First, suppose $\lambda > \frac{\sqrt{2}}{2}$. Choose a fixed $k$ such that $\lambda (2k+1) - k \sqrt{2} > \sqrt{2}$; such $k$ exists because $\lambda > \sqrt{2} \cdot \frac{k+1}{2k+1}$ for sufficiently large $k$, as $\frac{k+1}{2k+1} \to \frac{1}{2}$. Alice employs the following strategy: choose $x_n = 0$ for all odd $n \leq 2k-1$, and on $n = 2k+1$ choose the maximum $x_{2k+1} = \lambda (2k+1) - S_{2k}$. Her goal is to ensure $Q_{2k+1} = Q_{2k} + x_{2k+1}^2 > 2k+2$, so Bazza cannot choose any $x_{2k+2} \geq 0$ satisfying $Q_{2k+1} + x_{2k+2}^2 \leq 2k+2$.

We first prove by induction on $k$ that if Alice chooses $x_n = 0$ on all odd $n \leq 2k-1$, then the maximum achievable $S_{2k}$ (over all possible Bazza plays) is $k \sqrt{2}$, achieved when Bazza chooses $x_{2j} = \sqrt{2}$ for each $j=1$ to $k$, yielding $Q_{2k} = 2k$; moreover, for any Bazza play, $S_{2k} \leq k \sqrt{2}$.

For the base case $k=1$: Bazza chooses $x_2 \leq \sqrt{2 - Q_1} = \sqrt{2}$ (since $Q_1 = 0$), so max $S_2 = \sqrt{2}$, achieved at $x_2 = \sqrt{2}$, with $Q_2 = 2$.

Assume true for $k-1$: The max $S_{2(k-1)} = (k-1) \sqrt{2}$, achieved by $x_{2j} = \sqrt{2}$ for $j=1$ to $k-1$, with $Q_{2(k-1)} = 2(k-1)$; for any play, $S_{2(k-1)} \leq (k-1) \sqrt{2}$. Since $x_{2k-1} = 0$, $Q_{2k-1} = Q_{2(k-1)}$. Bazza then chooses $x_{2k} \leq \sqrt{2k - Q_{2k-1}} = \sqrt{2k - Q_{2(k-1)}}$. Thus, $S_{2k} = S_{2(k-1)} + x_{2k} \leq S_{2(k-1)} + \sqrt{2k - Q_{2(k-1)}}$. Let $t = S_{2(k-1)}$, $q = Q_{2(k-1)}$, with $0 \leq t \leq (k-1) \sqrt{2}$. For fixed $t$, the expression $t + \sqrt{2k - q}$ is maximized when $q$ is minimized (subject to possible $q$ for that $t$). By the induction hypothesis applied to previous steps and the fact that deviations lead to smaller $t$ or larger $q$ for that $t$, the overall maximum is at most $\max_{0 \leq t \leq (k-1) \sqrt{2}} g(t)$, where $g(t) = t + \sqrt{2k - \frac{t^2}{k-1}}$ (using the minimal possible $q = \frac{t^2}{k-1}$ by Cauchy-Schwarz on the first $k-1$ even terms). We have $g'(t) = 1 - \frac{t/(k-1)}{\sqrt{2k - t^2/(k-1)}}$. Setting $g'(t) = 0$ yields $t = (k-1) \sqrt{2}$, and $g((k-1) \sqrt{2}) = (k-1) \sqrt{2} + \sqrt{2k - 2(k-1)} = (k-1) \sqrt{2} + \sqrt{2} = k \sqrt{2}$. At the boundaries, $g(0) = \sqrt{2k} < k \sqrt{2}$ for $k \geq 1$, and since the critical point is at the upper endpoint and $g'(t) > 0$ for $t < (k-1) \sqrt{2}$ (as verified by the condition for $g' > 0$ holding strictly there), the maximum is $k \sqrt{2}$. This is achievable when $t = (k-1) \sqrt{2}$ and $q = 2(k-1)$, which occurs under the greedy strategy. If Bazza deviates earlier, yielding $t < (k-1) \sqrt{2}$ or $q > \frac{t^2}{k-1}$, then $S_{2k} < k \sqrt{2}$. Finally, independently, by Cauchy-Schwarz applied to the $k$ even terms (odd terms are $0$), $Q_{2k} = \sum_{j=1}^k x_{2j}^2 \geq \frac{1}{k} \left( \sum_{j=1}^k x_{2j} \right)^2 = \frac{S_{2k}^2}{k}$, with equality if and only if all $x_{2j}$ are equal. This completes the induction.

Now, let $c = \lambda (2k+1) > k \sqrt{2} \geq S_{2k}$ (by choice of $k$), so Alice can choose $x_{2k+1} = c - S_{2k} \geq 0$. Then $Q_{2k+1} = Q_{2k} + (c - S_{2k})^2 \geq \frac{S_{2k}^2}{k} + (c - S_{2k})^2$. Let $h(s) = \frac{s^2}{k} + (c - s)^2$ for $0 \leq s \leq k \sqrt{2}$. This is a quadratic in $s$ with minimum at $s = \frac{c k}{k+1} > k \sqrt{2}$ (since $c > k \sqrt{2}$ implies $\frac{c k}{k+1} > \frac{k^2 \sqrt{2}}{k+1} \approx k \sqrt{2}$ for large $k$), so on $[0, k \sqrt{2}]$, the minimum is at $s = k \sqrt{2}$, where $h(k \sqrt{2}) = 2k + (c - k \sqrt{2})^2 > 2k + 2$ (since $c - k \sqrt{2} > \sqrt{2}$). Thus, $Q_{2k+1} > 2k + 2$ for all achievable $(S_{2k}, Q_{2k})$, forcing Bazza stuck on $2k+2$.

Next, suppose $\lambda < \frac{\sqrt{2}}{2}$. Bazza chooses $x_{2j} = \sqrt{2j - Q_{2j-1}}$ on each even $2j$ (or $0$ if negative, but we show it is always positive). We prove by induction on $m$ that under this strategy, (i) Bazza can always move ($Q_{2j-1} < 2j$ for all $j \leq m$), (ii) the minimal achievable $S_{2m}$ is $m \sqrt{2}$, attained iff Alice chooses $x_{2j-1} = 0$ for all $j=1$ to $m$, and (iii) any deviation by Alice yields $S_{2m} > m \sqrt{2}$.

For $m=1$: Alice chooses $0 \leq x_1 \leq \lambda < \frac{\sqrt{2}}{2} < \sqrt{2}$, so $Q_1 = x_1^2 < 0.5 < 2$. Bazza chooses $x_2 = \sqrt{2 - x_1^2} > \sqrt{2 - 0.5} > 1 > 0$. Let $h(y) = y + \sqrt{2 - y^2}$ for $0 \leq y \leq \lambda < \frac{\sqrt{2}}{2}$. Then $h'(y) = 1 - \frac{y}{\sqrt{2 - y^2}} > 0$ (since $y < 1$ implies $y < \sqrt{2 - y^2}$), so $h$ is increasing, min $h(0) = \sqrt{2}$, and $h(y) > \sqrt{2}$ for $y > 0$. Thus, min $S_2 = \sqrt{2}$ at $x_1 = 0$, and $> \sqrt{2}$ otherwise.

Assume true for $m-1$: Min $S_{2(m-1)} = (m-1) \sqrt{2}$ iff all prior odd $x = 0$, with $Q_{2(m-1)} \leq 2(m-1)$ always, and $Q_{2j-1} < 2j$ for $j \leq m-1$. Alice chooses $0 \leq x_{2m-1} \leq r_{2m-1} = \lambda (2m-1) - S_{2(m-1)} \leq \lambda (2m-1) - (m-1) \sqrt{2} < \frac{\sqrt{2}}{2} < 1$. Then $Q_{2m-1} = Q_{2(m-1)} + x_{2m-1}^2 < 2(m-1) + 0.5 < 2m$ (even if $S_{2(m-1)} > (m-1) \sqrt{2}$, $r_{2m-1} < \frac{\sqrt{2}}{2}$, so still $Q_{2m-1} < 2m$). Bazza chooses $x_{2m} = \sqrt{2m - Q_{2m-1}} > 0$. Now $S_{2m} = S_{2(m-1)} + x_{2m-1} + \sqrt{2 - x_{2m-1}^2} \geq (m-1) \sqrt{2} + \min h(y)$ over $0 \leq y \leq r_{2m-1} < \frac{\sqrt{2}}{2}$, so min $= (m-1) \sqrt{2} + \sqrt{2} = m \sqrt{2}$ at $y=0$ (requiring prior min), and $> m \sqrt{2}$ if $y > 0$ or prior $> (m-1) \sqrt{2}$. This completes the induction.

Thus, max $r_{2m+1} = \lambda (2m+1) - m \sqrt{2} < \frac{\sqrt{2}}{2} < \sqrt{2}$, so Alice cannot force $Q_{2m+1} > 2m+2$ (requires $x_{2m+1} > \sqrt{2}$). Moreover, since $2\lambda - \sqrt{2} < 0$, max $r_{2m+1} \to -\infty$, eventually $<0$; on other paths, $S_{2m} > m \sqrt{2}$, so $r_{2m+1} <0$ sooner. Hence Alice gets stuck on some odd turn, while Bazza always moves.

Finally, for $\lambda = \frac{\sqrt{2}}{2}$, neither has a winning strategy. First, Bazza has no winning strategy (cannot force termination on an odd turn) because Alice can choose $x_n = 0$ on all odd $n$. Then, by the lemma above (which holds independently of $\lambda$, as it assumes Alice chooses $0$ on odds), $S_{2m} \leq m \sqrt{2}$ for any Bazza play, so $r_{2m+1} = \frac{\sqrt{2}}{2} (2m+1) - S_{2m} \geq \frac{\sqrt{2}}{2} (2m+1) - m \sqrt{2} = \frac{\sqrt{2}}{2} > 0$. Thus, Alice can always choose $x_{2m+1} = 0 \leq r_{2m+1}$.

Alice has no winning strategy (cannot force termination on an even turn) because Bazza can choose the greedy maximization strategy $x_{2j} = \sqrt{2j - Q_{2j-1}}$. We prove by induction on $m$ that under this strategy, (i) Bazza can always move ($Q_{2j-1} \leq 2j$ for all $j \leq m$), (ii) the minimal achievable $S_{2m}$ is $m \sqrt{2}$, attained iff Alice chooses $x_{2j-1} = 0$ for all $j=1$ to $m$, and (iii) any deviation by Alice yields $S_{2m} > m \sqrt{2}$. This induction is explicitly adapted from the $\lambda < \frac{\sqrt{2}}{2}$ case; the only difference is that now $\lambda = \frac{\sqrt{2}}{2}$, but since the key bound $r_{2m-1} \leq \frac{\sqrt{2}}{2} < 1$ still holds on the minimizing path (as verified below), the derivative argument for $h(y) = y + \sqrt{2 - y^2}$ being increasing on $[0, \frac{\sqrt{2}}{2}]$ (with $h'(y) > 0$ for $y < 1$) applies identically.

For $m=1$: Alice chooses $0 \leq x_1 \leq \lambda = \frac{\sqrt{2}}{2} < \sqrt{2}$, so $Q_1 = x_1^2 \leq 0.5 < 2$. Bazza chooses $x_2 = \sqrt{2 - x_1^2} \geq \sqrt{2 - 0.5} > 1 \geq 0$. As before, $h(y) = y + \sqrt{2 - y^2}$ is increasing on $[0, \frac{\sqrt{2}}{2}]$, so min $S_2 = \sqrt{2}$ at $x_1 = 0$, and $> \sqrt{2}$ otherwise.

Assume true for $m-1$: Min $S_{2(m-1)} = (m-1) \sqrt{2}$ iff all prior odd $x = 0$, with $Q_{2(m-1)} \leq 2(m-1)$ always, and $Q_{2j-1} \leq 2j$ for $j \leq m-1$. Alice chooses $0 \leq x_{2m-1} \leq r_{2m-1} = \lambda (2m-1) - S_{2(m-1)} \leq \lambda (2m-1) - (m-1) \sqrt{2} = \frac{\sqrt{2}}{2} < 1$ (on the minimizing path; if $S_{2(m-1)} > (m-1) \sqrt{2}$, then $r_{2m-1} < \frac{\sqrt{2}}{2}$). Then $Q_{2m-1} = Q_{2(m-1)} + x_{2m-1}^2 \leq 2(m-1) + 0.5 < 2m$. Bazza chooses $x_{2m} = \sqrt{2m - Q_{2m-1}} \geq 0$ (since $Q_{2m-1} < 2m$). Now $S_{2m} = S_{2(m-1)} + x_{2m-1} + \sqrt{2 - x_{2m-1}^2} \geq (m-1) \sqrt{2} + \min h(y)$ over $0 \leq y \leq r_{2m-1} \leq \frac{\sqrt{2}}{2}$, so min $= (m-1) \sqrt{2} + \sqrt{2} = m \sqrt{2}$ at $y=0$ (requiring prior min), and $> m \sqrt{2}$ if $y > 0$ or prior $> (m-1) \sqrt{2}$. This completes the induction. Thus, $Q_{2j-1} \leq 2j$ always, so Bazza can always move (choosing $x_{2j} = 0$ if $Q_{2j-1} = 2j$). Under these strategies, both always move, and the game continues indefinitely.

Alice: $\lambda > \dfrac{\sqrt{2}}{2}$
Bazza: $\lambda < \dfrac{\sqrt{2}}{2}$
(neither: $\lambda = \dfrac{\sqrt{2}}{2}$)

\subsubsection{Solution by GPT-5}

Let $S_n := \sum_{j=1}^n x_j$ and $Q_n := \sum_{j=1}^n x_j^2$. By the rules, at odd turns $2k-1$ Alice must ensure $S_{2k-1} \le \lambda(2k-1)$, and at even turns $2k$ Bazza must ensure $Q_{2k} \le 2k$.

Lemma A (Even-turn response and pairwise lower bound). Fix $k \ge 1$. Suppose the game is ongoing at the start of turn $2k$; in particular, Bazza made a legal move at turn $2k-2$, hence
$$
Q_{2k-2} \le 2(k-1).
$$
Assume also that Bazza is not already stuck at turn $2k$, i.e.,
$$
Q_{2k-1} \le 2k.
$$
Define
$$x_{2k} := \sqrt{\max\{0,\,2 - x_{2k-1}^2\}}.
$$
Then $x_{2k}$ is legal (so $Q_{2k} \le 2k$) and
$$x_{2k-1} + x_{2k} \ge \sqrt{2}.
$$
Consequently, if Bazza uses this response at each even turn for which the game reaches that turn, then for every $k$ for which turn $2k$ occurs,
$$
S_{2k} \ge \sqrt{2}\,k.
$$

Proof. Write $t := x_{2k-1} \ge 0$. Since $Q_{2k-1} = Q_{2k-2} + t^2$ and $Q_{2k-2} \le 2(k-1)$,
$$
2k - Q_{2k-1} = 2k - Q_{2k-2} - t^2 \;\ge\; 2 - t^2.
$$
Therefore $x_{2k}^2 = \max\{0, 2 - t^2\} \le 2k - Q_{2k-1}$, so the move is legal and $Q_{2k} \le 2k$. For the pairwise sum, if $t \ge \sqrt{2}$ then $t + x_{2k} \ge \sqrt{2}$. If $0 \le t \le \sqrt{2}$, then $x_{2k} = \sqrt{2 - t^2}$ and
$$t + x_{2k} = t + \sqrt{2 - t^2} \ge \sqrt{2}.
$$
Summing these lower bounds over $k$ pairs yields $S_{2k} \ge \sqrt{2}\,k$. \qed

Lemma B (Bazza’s safety for $\lambda \le 1/\sqrt{2}$). Assume $\lambda \le 1/\sqrt{2}$ and that Bazza, whenever it is his turn and he is not already stuck, uses the move from Lemma A. Then, for every $k \ge 1$,
$$
Q_{2k-1} < 2k.
$$

Proof. By induction on $k$. For $k=1$, Alice must choose $x_1 \le \lambda \le 1/\sqrt{2}$, so $Q_1 = x_1^2 \le 1/2 < 2$. Suppose the claim holds for all indices less than $k$. Then, for each $j \le k-1$, Bazza had a legal move at turn $2j$ (since $Q_{2j-1} < 2j$ by the induction hypothesis) and played according to Lemma A; thus $x_{2j-1} + x_{2j} \ge \sqrt{2}$, and summing yields
$$
S_{2k-2} \ge \sqrt{2}\,(k-1).
$$
At turn $2k-1$, Alice must pick $x_{2k-1} \ge 0$ with $S_{2k-2} + x_{2k-1} \le \lambda(2k-1)$, so
$$x_{2k-1} \le \lambda(2k-1) - S_{2k-2} \le \frac{1}{\sqrt{2}}(2k-1) - \sqrt{2}\,(k-1) = \frac{1}{\sqrt{2}}.
$$
Therefore
$$
Q_{2k-1} \;=\; Q_{2k-2} + x_{2k-1}^2
\;\le\; 2(k-1) + \frac{1}{2}
\;<\; 2k.
$$
This completes the induction. \qed

Proposition 1 (Bazza wins for $\lambda < 1/\sqrt{2}$). If $\lambda < 1/\sqrt{2}$, then Bazza has a winning strategy.

Proof. Bazza follows Lemma A at every even turn. By Lemma B, he is never stuck at even turns. Lemma A also yields
$$
S_{2k} \ge \sqrt{2}\,k \quad \text{for all } k \ge 1.
$$
Let $\delta := \frac{1}{\sqrt{2}} - \lambda > 0$. Then
$$
\sqrt{2}\,k - \lambda(2k+1) = ( \sqrt{2} - 2\lambda )\,k - \lambda = 2\delta\,k - \lambda,
$$
which is positive for all sufficiently large $k$. Hence for some $k$,
$$
S_{2k} \;>\; \lambda(2k+1),
$$
so Alice has no legal move at turn $2k+1$. Thus Bazza wins. \qed

Lemma C (Alice’s decisive odd move after zeros). Fix $k \ge 1$. Suppose $x_{2i-1} = 0$ for $i=1,\dots,k-1$ and that the game is still ongoing at the start of turn $2k-1$, so $Q_{2k-2} \le 2k-2$. Let
$$s := x_2 + x_4 + \cdots + x_{2k-2} = S_{2k-2}, \quad q := x_2^2 + x_4^2 + \cdots + x_{2k-2}^2 \;=\; Q_{2k-2},
$$
and define
$$
\Delta := \lambda(2k-1) - s.
$$
Then, for every legal prior history,
$$
\Delta - \sqrt{2k - q} \;\ge\; \lambda(2k-1) - \sqrt{2}\,k.
$$
In particular, if $\lambda(2k-1) > \sqrt{2}\,k$, then Alice’s legal choice $x_{2k-1} := \Delta$ satisfies $x_{2k-1} > \sqrt{2k - q}$, whence
$$
Q_{2k-1} \;=\; q + x_{2k-1}^2 \;>\; 2k,
$$
so Bazza is stuck at turn $2k$.

Proof. By Cauchy–Schwarz, $s^2 \le (k-1)q$, so $q \ge s^2/(k-1)$. Also $q \le 2k-2$ by legality. Consider
$$f(s,q) := \lambda(2k-1) - s - \sqrt{2k - q}.
$$
For fixed $s$, $f$ is increasing in $q$ because $q \mapsto -\sqrt{2k-q}$ is increasing. Hence, for $q \ge s^2/(k-1)$,
$$f(s,q) \;\ge\; f\!(s, \frac{s^2}{k-1}).
$$
Define $g(s) := f\!(s, \frac{s^2}{k-1}) = \lambda(2k-1) - s - \sqrt{2k - s^2/(k-1)}$. From $q \le 2k-2$ and $s^2 \le (k-1)q$ we get $0 \le s \le \sqrt{2}\,(k-1)$, and on this interval
$$g'(s) \;=\; -1 + \frac{s}{(k-1)\sqrt{\,2k - s^2/(k-1)\,\,}}
\;\le\; -1 + \frac{s}{(k-1)\sqrt{2}} \;\le\; 0.
$$
Thus $g$ is decreasing on $[0,\sqrt{2}(k-1)]$, so
$$g(s) \;\ge\; g\!(\sqrt{2}\,(k-1))
\;=\; \lambda(2k-1) - \sqrt{2}\,(k-1) - \sqrt{2}
\;=\; \lambda(2k-1) - \sqrt{2}\,k.
$$
Therefore $f(s,q) \ge \lambda(2k-1) - \sqrt{2}k$, proving the claimed bound. If $\lambda(2k-1) > \sqrt{2}k$, then $x_{2k-1} := \Delta$ is legal and satisfies $x_{2k-1} > \sqrt{2k - q}$, so $Q_{2k-1} > 2k$. \qed

Proposition 2 (Alice wins for $\lambda > 1/\sqrt{2}$). If $\lambda > 1/\sqrt{2}$, then Alice has a winning strategy.

Proof. Alice plays $x_{2i-1} = 0$ for $i=1,2,\dots$, until the first $k$ with $\lambda(2k-1) > \sqrt{2}\,k$. Such a $k$ exists because the sequence $\sqrt{2}\,k/(2k-1)$ is strictly decreasing and converges to $1/\sqrt{2}$. These zero moves are always legal: for each $i \ge 1$, by Cauchy–Schwarz and even-turn legality,
$$
S_{2i-1} = S_{2i-2} \le \sqrt{i-1}\,\sqrt{Q_{2i-2}} \le \sqrt{i-1}\,\sqrt{2i-2} = \sqrt{2}\,(i-1).
$$
Moreover, for every integer $i \ge 1$,
$$
\sqrt{2}\,(i-1) \;<\; \frac{1}{\sqrt{2}}(2i-1) \;\le\; \lambda(2i-1),
$$
since $(i-1)/(2i-1) < 1/2$ and $\lambda > 1/\sqrt{2}$. Thus $x_{2i-1}=0$ is legal for all $i$ prior to $k$. At the special odd turn $2k-1$, Lemma C gives
$$
\Delta - \sqrt{2k - Q_{2k-2}} \;\ge\; \lambda(2k-1) - \sqrt{2}\,k \;>\; 0.
$$
Therefore Alice’s legal choice $x_{2k-1} := \Delta$ forces $Q_{2k-1} > 2k$, so Bazza has no legal move at turn $2k$. Hence Alice wins. \qed

Proposition 3 (Draw at $\lambda = 1/\sqrt{2}$). If $\lambda = 1/\sqrt{2}$, neither player has a winning strategy.

Proof. Alice avoids losing by always choosing $x_{2i-1} = 0$. Indeed, for each $i \ge 1$,
$$
S_{2i-1} = S_{2i-2} \le \sqrt{i-1}\,\sqrt{Q_{2i-2}} \le \sqrt{i-1}\,\sqrt{2i-2} = \sqrt{2}\,(i-1) \le \frac{1}{\sqrt{2}}(2i-1) = \lambda(2i-1),
$$
so the odd-turn constraint is satisfied. Bazza avoids losing by using the strategy of Lemma A; with $\lambda = 1/\sqrt{2}$, Lemma B gives $Q_{2k-1} < 2k$ for all $k$, so Bazza always has a legal even move. Thus the game continues indefinitely and is a draw. \qed

Conclusion. Combining Propositions 1, 2, and 3:
- Alice has a winning strategy if and only if $\lambda > 1/\sqrt{2}$.
- Bazza has a winning strategy if and only if $\lambda < 1/\sqrt{2}$.
- For $\lambda = 1/\sqrt{2}$, neither player has a winning strategy.

\subsection{Problem 6} \label{a:6}

\begin{problem}
Consider a $2025 \times 2025$ grid of unit squares. Matilda wishes to place on the grid some rectangular tiles, possibly of different sizes, such that each side of every tile lies on a grid line and every unit square is covered by at most one tile.

Determine the minimum number of tiles Matilda needs to place so that each row and each column of the grid has exactly one unit square that is not covered by any tile.
\end{problem}

\subsubsection{Flawed Solution by Gemini 2.5 Pro}

Let the grid be an $N \times N$ grid of unit squares, where $N=2025$. Let the square in row $i$ and column $j$ be denoted by $(i,j)$ for $1 \le i,j \le N$.

**1. The Structure of Uncovered Squares**

The condition is that each row and each column of the grid has exactly one uncovered square. Let $U$ be the set of uncovered squares.
For each row $i$, there is a unique column $j$ such that $(i,j) \in U$. This allows us to define a function $\sigma: \{1, \dots, N\} \to \{1, \dots, N\}$, where the uncovered square in row $i$ is $(i, \sigma(i))$.
For each column $j$, there is a unique row $i$ such that $(i,j) \in U$. This means that for any $j_0 \in \{1, \dots, N\}$, there is a unique $i_0$ such that $\sigma(i_0) = j_0$. This implies that the function $\sigma$ is a bijection, i.e., a permutation.
So, the set of uncovered squares is $U = \{(i, \sigma(i)) \mid i=1, \dots, N\}$ for some permutation $\sigma$ of $\{1, \dots, N\}$.

**2. Partitioning the Tiled Region**

Let $C$ be the set of all squares covered by tiles. $C$ is the complement of $U$ in the grid. We can partition $C$ into two disjoint sets:
*   $C_L = \{(i,j) \mid j < \sigma(i)\}$, the set of squares in $C$ that are to the left of the uncovered square in their respective rows.
*   $C_R = \{(i,j) \mid j > \sigma(i)\}$, the set of squares in $C$ that are to the right of the uncovered square in their respective rows.

It is clear that $C = C_L \cup C_R$ and $C_L \cap C_R = \emptyset$.

**3. Tiles are Confined to $C_L$ or $C_R$**

Let a rectangular tile $R$ cover the squares $(i,j)$ for $i \in [r_1, r_2]$ and $j \in [c_1, c_2]$, where $[a,b]$ denotes the set of integers $\{a, a+1, \dots, b\}$. For $R$ to be a valid tile, all squares it covers must be in $C$. This means for any $i \in [r_1, r_2]$, the uncovered square $(i, \sigma(i))$ is not in $R$. This implies that for any $i \in [r_1, r_2]$, $\sigma(i) \notin [c_1, c_2]$.

Let $S_{\sigma} = \{\sigma(i) \mid i \in [r_1, r_2]\}$. The condition for $R$ to be a valid tile is $S_{\sigma} \cap [c_1, c_2] = \emptyset$.
Let $m = \min(S_{\sigma})$ and $M = \max(S_{\sigma})$. The condition is equivalent to either $c_2 < m$ or $c_1 > M$.

*   If $c_2 < m$, then for any square $(i,j) \in R$, we have $j \le c_2 < m \le \sigma(i)$ for all $i \in [r_1, r_2]$. Thus $j < \sigma(i)$, which implies $(i,j) \in C_L$. Therefore, the entire tile $R$ is contained in $C_L$.
*   If $c_1 > M$, then for any square $(i,j) \in R$, we have $j \ge c_1 > M \ge \sigma(i)$ for all $i \in [r_1, r_2]$. Thus $j > \sigma(i)$, which implies $(i,j) \in C_R$. Therefore, the entire tile $R$ is contained in $C_R$.

This proves that any tile must be entirely in $C_L$ or entirely in $C_R$. Consequently, the regions $C_L$ and $C_R$ are tiled independently. The total minimum number of tiles, $k$, is the sum of the minimum number of tiles required to tile $C_L$ ($k_L$) and $C_R$ ($k_R$). So, $k = k_L + k_R$.

**4. Minimum Number of Tiles for $C_L$**

Let's determine $k_L$ for an arbitrary permutation $\sigma$.
Consider the set of squares $S_L = \{(i, \sigma(i)-1) \mid i \in \{1, \dots, N\} \text{ such that } \sigma(i) > 1\}$.
The size of this set is $|S_L| = N-1$, since there is exactly one $i_0$ for which $\sigma(i_0)=1$. Each square in $S_L$ is in $C_L$.

Let $s_i = (i, \sigma(i)-1)$ and $s_j = (j, \sigma(j)-1)$ be any two distinct squares in $S_L$. Assume without loss of generality that $i < j$. Any tile that covers both $s_i$ and $s_j$ must contain the smallest rectangle $R_{ij}$ that contains them. The rows of $R_{ij}$ are $[i,j]$ and the columns are $[\min(\sigma(i)-1, \sigma(j)-1), \max(\sigma(i)-1, \sigma(j)-1)]$.

We investigate if $R_{ij}$ contains any uncovered square from $U$.
*   **Case 1: $\sigma(i) < \sigma(j)$.** The column interval of $R_{ij}$ is $[\sigma(i)-1, \sigma(j)-1]$. Consider the uncovered square $u_i = (i, \sigma(i))$. Its row $i$ is in $[i,j]$. Its column $\sigma(i)$ is in $[\sigma(i)-1, \sigma(j)-1]$ because $\sigma(i)-1 \le \sigma(i)$ and $\sigma(i) < \sigma(j)$ implies $\sigma(i) \le \sigma(j)-1$ for integers. So, $u_i \in R_{ij}$.
*   **Case 2: $\sigma(i) > \sigma(j)$.** The column interval of $R_{ij}$ is $[\sigma(j)-1, \sigma(i)-1]$. Consider the uncovered square $u_j = (j, \sigma(j))$. Its row $j$ is in $[i,j]$. Its column $\sigma(j)$ is in $[\sigma(j)-1, \sigma(i)-1]$ because $\\sigma(j)-1 \le \sigma(j)$ and $\sigma(j) < \sigma(i)$ implies $\sigma(j) \le \sigma(i)-1$. So, $u_j \in R_{ij}$.

In both cases, the rectangle $R_{ij}$ contains an uncovered square. Therefore, $s_i$ and $s_j$ cannot be in the same tile. This holds for any pair of distinct squares in $S_L$. Thus, all $N-1$ squares in $S_L$ must belong to different tiles. This provides a lower bound on the number of tiles for $C_L$: $k_L \ge N-1$.

To show this bound is achievable, we can tile $C_L$ with exactly $N-1$ tiles. For each $i$ such that $\sigma(i)>1$, the region to be tiled in row $i$ is $\{(i,j) \mid 1 \le j \le \sigma(i)-1\}$. This is a single rectangular region, which can be covered by one tile $T_i = [i,i] \times [1, \sigma(i)-1]$. There are $N-1$ such values of $i$, so this construction uses $N-1$ tiles to cover all of $C_L$.

Thus, for any permutation $\sigma$, the minimum number of tiles for $C_L$ is exactly $k_L = N-1$.

**5. Minimum Number of Tiles for $C_R$**

We use a symmetry argument to determine $k_R$. Consider a 180-degree rotation of the grid. A square $(i,j)$ is mapped to $(i',j') = (N+1-i, N+1-j)$. An uncovered square $(i, \sigma(i))$ is mapped to $(N+1-i, N+1-\sigma(i))$.
Let's define a new permutation $\tau$ on $\{1, \dots, N\}$ by $\tau(i') = N+1-\sigma(N+1-i')$. The set of uncovered squares in the rotated grid is $\{(i', \tau(i')) \mid i'=1, \dots, N\}$.
The region $C_R(\sigma) = \{(i,j) \mid j > \sigma(i)\}$ in the original grid corresponds to the region of squares $(i',j')$ where $N+1-j' > \sigma(N+1-i')$. This inequality is equivalent to $j' < N+1-\sigma(N+1-i')$, which is $j' < \tau(i')$. This is precisely the region $C_L(\tau)$ for the permutation $\tau$.
The minimum number of tiles required to tile a region is invariant under rotation. Thus, the minimum number of tiles for $C_R(\sigma)$ is equal to the minimum number of tiles for $C_L(\tau)$.
Since we proved that $k_L(\pi) = N-1$ for *any* permutation $\pi$, it holds for $\tau$. So, $k_R(\sigma) = k_L(\tau) = N-1$.

**6. Conclusion**

For any permutation $\sigma$, the minimum number of tiles is $k = k_L + k_R = (N-1) + (N-1) = 2N-2$.
Since the number of tiles is independent of the choice of permutation, the minimum number of tiles Matilda needs is $2N-2$.

For $N=2025$, the minimum number of tiles is $2 \times (2025-1) = 2 \times 2024 = 4048$.

\subsubsection{Flawed Solution by Grok-4}

Let $n=2025$. Place the uncovered squares at positions $(i,i)$ for $i=1,\dots,n$. This ensures exactly one uncovered square per row and per column.

The covered squares are all $(i,j)$ with $i \neq j$. Define $L = \{(i,j) \mid 1 \le j < i \le n\}$ and $R = \{(i,j) \mid 1 \le i < j \le n\}$. The sets $L$ and $R$ are disjoint and their union is the set of all covered squares.

Consider any rectangle on the grid, spanning consecutive rows $[a,b]$ and consecutive columns $[c,d]$. For the rectangle to be valid (i.e., to contain no uncovered squares and to have all its unit squares available for covering), it must not contain any $(k,k)$ for $k \in [a,b]$, so $[a,b] \cap [c,d] = \emptyset$. Thus, either $b < c$ (whence for all $i \in [a,b]$, $j \in [c,d]$ satisfies $j \ge c > b \ge i$, so in $R$) or $d < a$ (whence $j \le d < a \le i$, so in $L$). Therefore, no valid rectangle contains squares from both $L$ and $R$.

The set $R$ consists of squares $(i,j)$ with $1 \le i < j \le n$. In column $j$ (for $j=2$ to $n$), the squares are in rows $1$ to $j-1$. Tile $R$ with $n-1$ rectangles: for each $j=2$ to $n$, place a $(j-1) \times 1$ rectangle in column $j$ spanning rows $1$ to $j-1$. These rectangles are disjoint, their union is $R$, and each is valid (contains no uncovered square, since for row $r \in [1,j-1]$, the uncovered square in row $r$ is $(r,r)$ with $r \neq j$).

The set $L$ consists of squares $(i,j)$ with $1 \le j < i \le n$. In column $j$ (for $j=1$ to $n-1$), the squares are in rows $j+1$ to $n$. Tile $L$ with $n-1$ rectangles: for each $j=1$ to $n-1$, place a $(n-j) \times 1$ rectangle in column $j$ spanning rows $j+1$ to $n$. These rectangles are disjoint, their union is $L$, and each is valid (contains no uncovered square, since for row $r \in [j+1,n]$, the uncovered square in row $r$ is $(r,r)$ with $r \neq j$).

This tiling uses $2(n-1)$ rectangles in total and covers all covered squares without overlap.

To show this is minimal, note that any tiling must tile $L$ and $R$ separately (no rectangle crosses). Consider $R$: the squares $s_i = (i,i+1)$ for $i=1$ to $n-1$. Suppose some rectangle contains two such squares $s_p$ and $s_q$ with $p<q$. This rectangle spans rows $[a,b] \supseteq [p,q]$ and columns $[c,d] \supseteq [p+1,q+1]$. Thus, it contains at least the sub-rectangle $[p,q] \times [p+1,q+1]$. This sub-rectangle includes $(p+1,p+1)$ (since $p \le p+1 \le q$ and $p+1 \le p+1 \le q+1$), but $(p+1,p+1) \notin R$. Therefore, no rectangle in any tiling of $R$ contains two of $\{s_i\}$. Each $s_i$ requires its own rectangle, so at least $n-1$ rectangles are needed for $R$. Similarly, for $L$, the squares $(i,i-1)$ for $i=2$ to $n$ require at least $n-1$ rectangles. Thus, at least $2(n-1)$ rectangles are needed in total.

For $n=2025$, the minimum is $2 \times 2024 = 4048$.

\boxed{4048}

\subsection{Flawed Solution by GPT-5}

Let \(n\ge 2\). We work on an \(n\times n\) grid of unit squares, indexed by pairs \((r,c)\) with \(r,c\in\{1,\dots,n\}\). A tile is a set of the form \(\{(r,c): r\in [r_1,r_2]\cap\mathbb{Z},\ c\in [c_1,c_2]\cap\mathbb{Z}\}\) for some integers \(1\le r_1\le r_2\le n\) and \(1\le c_1\le c_2\le n\). Tiles are pairwise interior-disjoint and cover all but exactly one unit square in each row and in each column.

Upper bound: \(t\le 2n-2\).

We construct a tiling with \(2n-2\) tiles.

- Place the \(n\) holes at \((i,n+1-i)\) for \(i=1,\dots,n\) (the anti-diagonal).

- For each \(i\in\{1,\dots,n-1\}\), define the horizontal rectangle
  \[
  R_i^h=\{(i,c): 1\le c\le n-i\}.
  \]
- For each \(j\in\{2,\dots,n\}\), define the vertical rectangle
  \[
  R_j^v=\{(r,j): n+2-j\le r\le n\}.
  \]

We verify the required properties.

1) Disjointness:
- If \(i\neq i'\), then \(R_i^h\) and \(R_{i'}^h\) lie in different rows and are disjoint.
- If \(j\neq j'\), then \(R_j^v\) and \(R_{j'}^v\) lie in different columns and are disjoint.
- To show \(R_i^h\cap R_j^v=\varnothing\) for all \(i\in\{1,\dots,n-1\}\) and \(j\in\{2,\dots,n\}\), suppose for contradiction that \((r,c)\in R_i^h\cap R_j^v\). From \((r,c)\in R_i^h\) we have \(r=i\) and \(1\le c\le n-i\). From \((r,c)\in R_j^v\) we have \(c=j\) and \(n+2-j\le r\le n\). Combining, \(j\le n-i\) and \(i\ge n+2-j\). Hence \(i+j\le n\) and \(i+j\ge n+2\), a contradiction.

2) Coverage of all non-hole cells and no coverage of holes:
- If \((r,c)\) is a hole, then \(c=n+1-r\). Such a cell lies in neither \(R_r^h\) (since its column exceeds \(n-r\)) nor \(R_c^v\) (since its row is less than \(n+2-c\)), hence is uncovered as required.
- If \(c<n+1-r\) (the cell lies to the left of the hole in row \(r\)), then \(c\le n-r\) and \((r,c)\in R_r^h\).
- If \(c>n+1-r\) (the cell lies to the right of the hole in row \(r\)), then \(r\ge n+2-c\) and \((r,c)\in R_c^v\).

Thus every covered cell lies in exactly one tile, and there are exactly \(2n-2\) tiles. Therefore \(t\le 2n-2\).

Lower bound: \(t\ge 2n-2\).

Let the unique hole in row \(i\) be at \((i,\pi(i))\), where \(\pi\) is a permutation of \([n]=\{1,\dots,n\}\). Define
\[
a_i=(i,\pi(i+1))\quad (1\le i\le n-1),\qquad b_i=(i+1,\pi(i))\quad (1\le i\le n-1),
\]
and
\[
S=\{a_i:1\le i\le n-1\}\ \cup\ \{b_i:1\le i\le n-1\}.
\]

We establish three facts.

Fact 1 (All points of \(S\) are covered). For \(1\le i\le n-1\), the hole in row \(i\) is at column \(\pi(i)\), so \(a_i=(i,\pi(i+1))\) is covered since \(\pi(i+1)\neq \pi(i)\). Similarly, the hole in row \(i+1\) is at column \(\pi(i+1)\), so \(b_i=(i+1,\pi(i))\) is covered since \(\pi(i)\neq \pi(i+1)\).

Fact 2 (The elements of \(S\) are pairwise distinct, so \(|S|=2n-2\)). If \(a_i=a_j\), then equality of rows gives \(i=j\). If \(b_i=b_j\), then equality of rows gives \(i=j\). If \(a_i=b_j\), then equality of rows gives \(i=j+1\), so \(j=i-1\). Equality of columns then gives \(\pi(i+1)=\pi(j)=\pi(i-1)\), which contradicts the injectivity of \(\pi\) because \(i+1\neq i-1\). Hence all elements are distinct and \(|S|=2n-2\).

Fact 3 (Geometric fooling set). For any two distinct points \(s,t\in S\), the axis-aligned bounding box of \(\{s,t\}\) contains a hole \((k,\pi(k))\). Consequently, no tile can contain two distinct points of \(S\).

Proof of Fact 3. Write \(a_i=(i,\pi(i+1))\) and \(b_i=(i+1,\pi(i))\). Consider three cases.

- Case A: \(s=a_i\) and \(t=a_j\) with \(i<j\). The bounding box has row interval \([i,j]\) and column interval with endpoints \(\pi(i+1)\) and \(\pi(j+1)\). Since \(i+1\in[i,j]\) and \(\pi(i+1)\) lies between the endpoints (indeed, it is one endpoint), the hole \((i+1,\pi(i+1))\) is inside the box.

- Case B: \(s=b_i\) and \(t=b_j\) with \(i<j\). The bounding box has row interval \([i+1,j+1]\) and column interval with endpoints \(\pi(i)\) and \(\pi(j)\). Since \(j\in[i+1,j+1]\) and \(\pi(j)\) lies between the endpoints (again, one endpoint), the hole \((j,\pi(j))\) is inside the box.

- Case C: \(s=a_i\) and \(t=b_j\). Without loss of generality assume \(i\le j\). The bounding box has row interval \([i,j+1]\) and column interval with endpoints \(\pi(i+1)\) and \(\pi(j)\). If \(\pi(i+1)\le \pi(j)\), then \((i+1,\pi(i+1))\) lies in the box since \(i+1\in[i,j+1]\). If \(\pi(i+1)>\pi(j)\), then \((j,\pi(j))\) lies in the box since \(j\in[i,j+1]\). Thus in either subcase the box contains a hole.

In all cases, the box contains a hole. Since any tile is a Cartesian product of an interval of rows and an interval of columns, if a tile contained both \(s\) and \(t\) then it would contain the entire bounding box of \(\{s,t\}\), hence it would cover a hole—impossible. Therefore each tile contains at most one element of \(S\).

Conclusion of the lower bound. Because there are \(|S|=2n-2\) distinct covered cells in \(S\) and each tile can contain at most one of them, any valid tiling uses at least \(2n-2\) tiles. Hence \(t\ge 2n-2\).

Combining the bounds \(t\le 2n-2\) and \(t\ge 2n-2\), we conclude \(t=2n-2\). For \(n=2025\), this yields
\[
t=2\cdot 2025-2=4048.
\]

\section{Hint-Based Solutions from Initial Experiments} \label{appB}

\subsection{Solution to Problem 1 by Gemini 2.5 Pro}

\textit{The following solution was generated by providing the model with the additional prompt: ``Let us try to solve the problem by induction.''}

Let $S_n = \{(a,b) \in \mathbb{Z}^2 \mid a \ge 1, b \ge 1, a+b \le n+1\}$. A line is *sunny* if its slope is not $0, \infty,$ or $-1$. We seek the set of non-negative integers $k$ such that for a given integer $n \ge 3$, there exist $n$ distinct lines covering $S_n$, exactly $k$ of which are sunny.

**Part 1: Reduction to a Core Problem**

Let $\mathcal{L}$ be a set of $n$ distinct lines covering $S_n$. The *principal boundary lines* of $S_n$ are the lines $x=1$, $y=1$, and $x+y=n+1$. These are all non-sunny.

**Lemma:** If a principal boundary line $L$ of $S_n$ is in $\mathcal{L}$, then there exists a set of $n-1$ distinct lines $\mathcal{L}'$ that covers $S_{n-1}$ and has the same number of sunny lines as $\mathcal{L}$.

*Proof.* Let $k$ be the number of sunny lines in $\mathcal{L}$. The line $L$ is not sunny.
1.  Suppose $L$ is the line $x=1$. This line covers the points $\{(1,b) \mid 1 \le b \le n\}$. The remaining $n-1$ lines in $\mathcal{L} \setminus \{L\}$ must cover the set $S'_n = \{(a,b) \in S_n \mid a \ge 2\}$. The affine transformation $\phi(a,b) = (a-1, b)$ is a bijection from $S'_n$ to $S_{n-1}$. Let $\mathcal{L}'$ be the set of lines obtained by transforming the lines in $\mathcal{L} \setminus \{L\}$. A point $(x',y')$ is on a transformed line $l'$ if its pre-image under $\phi$, which is $(x'+1, y')$, is on the original line $l$. If $l$ has equation $Ax+By+C=0$, the transformed line $l'$ has equation $A(x+1)+By+C=0$. This transformation preserves the slope of any line, so the number of sunny lines in $\mathcal{L}'$ is $k$. The lines in $\mathcal{L}'$ are distinct: if $l_1: A_1x+B_1y+C_1=0$ and $l_2: A_2x+B_2y+C_2=0$ are distinct lines in $\mathcal{L} \setminus \{L\}$, their images are $l'_1: A_1x+B_1y+(A_1+C_1)=0$ and $l'_2: A_2x+B_2y+(A_2+C_2)=0$. If $l'_1=l'_2$, their coefficients are proportional, so $(A_1, B_1, A_1+C_1) = \lambda(A_2, B_2, A_2+C_2)$ for some $\lambda \ne 0$. This implies $A_1=\lambda A_2$, $B_1=\lambda B_2$, and $A_1+C_1=\lambda(A_2+C_2)$. Substituting the first into the third gives $\lambda A_2+C_1=\lambda A_2+\lambda C_2$, so $C_1=\lambda C_2$. Thus $(A_1,B_1,C_1)=\lambda(A_2,B_2,C_2)$, contradicting the distinctness of $l_1, l_2$.
2.  If $L$ is $y=1$, a symmetric argument with the transformation $(a,b) \mapsto (a, b-1)$ applies.
3.  If $L$ is $x+y=n+1$, it covers the points $\{(a,b) \in S_n \mid a+b=n+1\}$. The other $n-1$ lines must cover the remaining points, which form the set $\{(a,b) \in S_n \mid a+b \le n\} = S_{n-1}$. Thus, $\mathcal{L}' = \mathcal{L} \setminus \{L\}$ is a set of $n-1$ distinct lines covering $S_{n-1}$ with $k$ sunny lines.

By repeatedly applying this lemma, any configuration for $S_n$ can be reduced. This process terminates when we obtain a set of $m$ lines covering $S_m$ (for some $m \le n$) where none of the lines are principal boundary lines of $S_m$. The number of sunny lines in this new configuration is still $k$.

Let $\mathcal{L}_m$ be such a set of $m$ lines covering $S_m$.
Let $C_{m,x} = \{(1,b) \mid 1 \le b \le m\}$. These $m$ points must be covered by $\mathcal{L}_m$. Since $x=1 \notin \mathcal{L}_m$, any vertical line in $\mathcal{L}_m$ (of the form $x=c, c \ne 1$) misses all points in $C_{m,x}$. Each of the non-vertical lines in $\mathcal{L}_m$ can cover at most one point of $C_{m,x}$. If there are $v$ vertical lines in $\mathcal{L}_m$, then the remaining $m-v$ lines must cover all $m$ points, so $m-v \ge m$, which implies $v=0$. Thus, $\mathcal{L}_m$ contains no vertical lines (slope $\infty$).
By a symmetric argument considering $C_{m,y} = \{(a,1) \mid 1 \le a \le m\}$, we deduce $\mathcal{L}_m$ contains no horizontal lines (slope $0$).
By considering $C_{m,sum} = \{(a,b) \in S_m \mid a+b=m+1\}$, we deduce $\mathcal{L}_m$ contains no lines of slope $-1$.
Therefore, all $m$ lines in $\mathcal{L}_m$ must be sunny. This implies $k=m$. The problem reduces to finding for which non-negative integers $k$ it is possible to cover $S_k$ with $k$ distinct sunny lines, none of which are principal boundary lines of $S_k$.

**Part 2: Analysis of the Core Problem**

*   **$k=0$:** $S_0 = \emptyset$. This is covered by 0 lines. So $k=0$ is possible.
*   **$k=1$:** $S_1 = \{(1,1)\}$. The line $y=x$ is sunny, covers $(1,1)$, and is not a principal boundary line of $S_1$ (which are $x=1, y=1, x+y=2$). So $k=1$ is possible.
*   **$k=2$:** $S_2 = \{(1,1), (1,2), (2,1)\}$. We need to cover these three points with two distinct sunny lines, neither of which can be a principal boundary line of $S_2$ ($x=1, y=1, x+y=3$). A line passing through any two of the points in $S_2$ must be one of these three principal boundary lines, which are non-sunny. Therefore, a sunny line can pass through at most one point of $S_2$. Consequently, two sunny lines can cover at most two points, leaving at least one point of $S_2$ uncovered. So $k=2$ is impossible.
*   **$k=3$:** $S_3 = \{(1,1), (1,2), (1,3), (2,1), (2,2), (3,1)\}$. The following three lines are sunny, distinct, not principal boundary lines of $S_3$, and cover $S_3$:
    $L_1: y = -2x+5$ (covers $(1,3), (2,1)$)
    $L_2: y = x$ (covers $(1,1), (2,2)$)
    $L_3: x+2y=5$ (covers $(1,2), (3,1)$)
    So $k=3$ is possible.
*   **$k \ge 4$:** Assume there exists a set $\mathcal{L}$ of $k$ distinct sunny lines covering $S_k$, with no principal boundary lines. As argued, each line in $\mathcal{L}$ must intersect each of the sets $C_y=\{(i,1)\}_{i=1}^k$, $C_x=\{(1,j)\}_{j=1}^k$, and $C_s=\{(p,k+1-p)\}_{p=1}^k$ at exactly one point.
    We can label the lines $L_i$ for $i \in \{1, \dots, k\}$ such that $L_i$ is the unique line in $\mathcal{L}$ passing through $(i,1)$. For each $i$, $L_i$ must also pass through a unique point $(1, \sigma(i)) \in C_x$ and a unique point $(\pi(i), k+1-\pi(i)) \in C_s$. The maps $\sigma, \pi: \{1, \dots, k\} \to \{1, \dots, k\}$ must be permutations.

    These permutations must satisfy several properties:
    1.  $L_1$ passes through $(1,1)$, so $\sigma(1)=1$.
    2.  $L_k$ passes through $(k,1)$, which is also in $C_s$. Thus, $(\pi(k), k+1-\pi(k))=(k,1)$, which implies $\pi(k)=k$.
    3.  There is a unique line $L_j$ passing through $(1,k) \in C_x$, so $\sigma(j)=k$. The point $(1,k)$ is also in $C_s$, so for $L_j$, $(\pi(j), k+1-\pi(j))=(1,k)$, which implies $\pi(j)=1$. Since $\sigma(1)=1 \ne k$ and $\pi(k)=k \ne 1$, we have $j \ne 1,k$, so $j \in \{2, \dots, k-1\}$.
    4.  For $i \in \{2, \dots, k\}$, if $\sigma(i)=i$, $L_i$ would pass through $(i,1)$ and $(1,i)$, giving it slope $-1$, which is not allowed. So $\sigma(i) \ne i$ for $i \ge 2$.
    5.  For $i \in \{1, \dots, k-1\}$, if $\pi(i)=i$, $L_i$ would pass through $(i,1)$ and $(i, k+1-i)$, making it a vertical line, which is not allowed. So $\pi(i) \ne i$ for $i \le k-1$.

    For any $i \in \{1, \dots, k\}$, the three points $P_1=(i,1)$, $P_2=(1, \sigma(i))$, and $P_3=(\pi(i), k+1-\pi(i))$ must be collinear. For $i \in \{2, \dots, k-1\} \setminus \{j\}$, these three points are distinct. To prove this:
    - $P_1=P_2 \implies i=1$, but $i \ge 2$.
    - $P_1=P_3 \implies \pi(i)=i$, but $\pi(i) \ne i$ for $i \le k-1$.
    - $P_2=P_3 \implies \pi(i)=1$ and $\sigma(i)=k$. By definition of $j$, this means $i=j$. But we consider $i \ne j$.
    Thus, for $i \in \{2, \dots, k-1\} \setminus \{j\}$, the points are distinct. Collinearity implies their slopes are equal: $\frac{\sigma(i)-1}{1-i} = \frac{k+1-\pi(i)-1}{\pi(i)-i}$. This gives $\sigma(i) = 1 + (i-1)\frac{k-\pi(i)}{i-\pi(i)}$.

    **Step 1: Show $j=k-1$.**
    Assume for contradiction that $j \ne k-1$. Then the formula for $\sigma(i)$ is valid for $i=k-1$.
    $\sigma(k-1) = 1 + (k-2)\frac{k-\pi(k-1)}{k-1-\pi(k-1)}$.
    From the properties of $\pi$: $\pi(k-1) \in \{1, \dots, k\}$, $\pi(k-1) \ne \pi(j)=1$, $\pi(k-1) \ne \pi(k)=k$, and $\pi(k-1) \ne k-1$. So $\pi(k-1) \in \{2, \dots, k-2\}$.
    Let $d = k-1-\pi(k-1)$. Then $d \in \{1, \dots, k-3\}$.
    $\sigma(k-1) = 1 + (k-2)\frac{k-(k-1-d)}{d} = 1 + (k-2)\frac{d+1}{d} = 1+(k-2)(1+\frac{1}{d}) = k-1 + \frac{k-2}{d}$.
    Since $k \ge 4$, we have $d \le k-3$, so $\frac{k-2}{d} \ge \frac{k-2}{k-3} = 1+\frac{1}{k-3} > 1$.
    So $\sigma(k-1) > k-1+1=k$. This contradicts $\sigma(k-1) \in \{1, \dots, k\}$.
    Thus, our assumption was false. We must have $j=k-1$.

    **Step 2: Derive a final contradiction.**
    We have established that for $k \ge 4$, it must be that $j=k-1$, which means $\sigma(k-1)=k$ and $\pi(k-1)=1$.
    Consider $i=2$. Since $k \ge 4$, $j=k-1 \ge 3$, so $j \ne 2$. The formula for $\sigma(i)$ is valid for $i=2$:
    $\sigma(2) = 1 + \frac{k-\pi(2)}{2-\pi(2)}$.
    Since $\pi$ is a permutation, its image on $\{1, \dots, k-2\}$ is $\{1, \dots, k\} \setminus \{\pi(k-1), \pi(k)\}$. With $\pi(k-1)=1$ and $\pi(k)=k$, we have $\pi(\{1, \dots, k-2\}) = \{2, \dots, k-1\}$.
    So $\pi(2) \in \{2, \dots, k-1\}$. By property 5, $\pi(2) \ne 2$. Thus $\pi(2) \in \{3, \dots, k-1\}$.
    Let $d' = \pi(2)-2$. Then $d' \in \{1, \dots, k-3\}$.
    The denominator is $2-\pi(2) = -d'$.
    $\sigma(2) = 1 + \frac{k-(d'+2)}{-d'} = 1 - \frac{k-d'-2}{d'} = 1 - \left(\frac{k-2}{d'}-1\right) = 2 - \frac{k-2}{d'}$.
    Since $d' \le k-3$, we have $\frac{k-2}{d'} \ge \frac{k-2}{k-3} = 1+\frac{1}{k-3} > 1$ for $k \ge 4$.
    So $\sigma(2) = 2 - \frac{k-2}{d'} < 2-1=1$.
    This contradicts $\sigma(2) \in \{1, \dots, k\}$.
    This final contradiction shows that no such configuration is possible for $k \ge 4$.

**Part 3: Constructions for general $n \ge 3$**

The set of possible values for $k$ is $\{0, 1, 3\}$. We now show these are all possible for any $n \ge 3$.

*   **$k=0$:** Let $\mathcal{L} = \{x=i \mid i=1, 2, \dots, n\}$. These are $n$ distinct non-sunny lines. Any point $(a,b) \in S_n$ has $1 \le a \le n$, so it lies on the line $x=a \in \mathcal{L}$.
*   **$k=1$:** Let $\mathcal{L} = \{x=i \mid i=1, \dots, n-1\} \cup \{y=x-(n-1)\}$. The first $n-1$ lines are non-sunny and cover all points $(a,b) \in S_n$ with $a \le n-1$. The only points in $S_n$ not covered are those with $a \ge n$. If $(a,b) \in S_n$ and $a \ge n$, then $n+b \le a+b \le n+1$, which implies $b \le 1$. Since $b \ge 1$, we must have $b=1$. Then $a+1 \le n+1 \implies a \le n$. The only uncovered point is $(n,1)$. The line $y=x-(n-1)$ has slope 1 (sunny), is distinct from the other lines, and passes through $(n,1)$ since $1 = n-(n-1)$.
*   **$k=3$:** Let $\mathcal{L}_{ns} = \{x+y=c \mid c=5, 6, \dots, n+1\}$. This is a set of $n-3$ distinct non-sunny lines (if $n=3$, this set is empty). These lines cover all points $(a,b) \in S_n$ with $a+b \ge 5$. The remaining points to be covered are $U = \{(a,b) \in S_n \mid a+b \le 4\}$. This set is precisely $S_3$. As shown in Part 2, $S_3$ can be covered by 3 distinct sunny lines. Let $\mathcal{L}_s$ be this set of 3 lines. Then $\mathcal{L} = \mathcal{L}_{ns} \cup \mathcal{L}_s$ is a set of $(n-3)+3=n$ lines covering $S_n$. The lines in $\mathcal{L}_s$ have slopes different from $-1$, so they are distinct from lines in $\mathcal{L}_{ns}$. This construction provides exactly 3 sunny lines.

Thus, for any integer $n \ge 3$, configurations with $k \in \{0, 1, 3\}$ sunny lines are possible.

\subsection{Solution to Problem 2 by Gemini 2.5 Pro}

\textit{The following proof was generated by providing the model with the additional prompt: ``Let us try to solve the problem by analytic geometry.''}

**1. Coordinate System and Geometric Setup**

We establish a Cartesian coordinate system. Let the line of centers $MN$ be the $x$-axis, and the common chord $AB$ be the $y$-axis. Let their intersection be the origin $O(0,0)$.

*   The points of intersection of the circles are $A(0,a)$ and $B(0,-a)$ for some $a>0$.
*   The centers of the circles are $M(m,0)$ and $N(n,0)$.
*   Circle $\Omega$ has center $M(m,0)$ and passes through $A(0,a)$. Its radius is $R_\Omega = \sqrt{m^2+a^2}$. The equation of $\Omega$ is $(x-m)^2+y^2 = m^2+a^2$, which simplifies to $x^2-2mx+y^2=a^2$.
*   Circle $\Gamma$ has center $N(n,0)$ and passes through $A(0,a)$. Its radius is $R_\Gamma = \sqrt{n^2+a^2}$. The equation of $\Gamma$ is $(x-n)^2+y^2 = n^2+a^2$, which simplifies to $x^2-2nx+y^2=a^2$.
*   The line $MN$ is the $x$-axis. The points $C$ and $D$ lie on the $x$-axis. Let their coordinates be $C(c,0)$ and $D(d,0)$.
*   The condition that $C, M, N, D$ lie on the line in that order implies their $x$-coordinates are ordered: $c < m < n < d$.
*   $C$ is an intersection of $\Omega$ with the $x$-axis. Its coordinate $c$ is a root of $x^2-2mx-a^2=0$. The roots are $m \pm \sqrt{m^2+a^2}$. Since $c<m$, we must have $c = m-\sqrt{m^2+a^2}$.
*   $D$ is an intersection of $\Gamma$ with the $x$-axis. Its coordinate $d$ is a root of $x^2-2nx-a^2=0$. The roots are $n \pm \sqrt{n^2+a^2}$. Since $d>n$, we must have $d=n+\sqrt{n^2+a^2}$.
*   From the equations for $c$ and $d$, we have $c^2-2mc-a^2=0$ and $d^2-2nd-a^2=0$. Since $c<0$ and $d>0$, $c,d \neq 0$. We can express $m$ and $n$ as:
    $m = \frac{c^2-a^2}{2c}$ and $n = \frac{d^2-a^2}{2d}$.

**2. Preliminary Geometric and Algebraic Properties**

We establish two key properties derived from the problem's conditions.

*   **Lemma 1:** $c+d \neq 0$.
    *   **Proof:** The condition $R_\Omega < R_\Gamma$ implies $m^2+a^2 < n^2+a^2$, so $m^2 < n^2$.
    *   The ordering $C,M,N,D$ implies $m<n$.
    *   From $m<n$ and $m^2<n^2$, we have $n^2-m^2 > 0$, so $(n-m)(n+m)>0$. Since $n-m>0$, we must have $n+m>0$.
    *   Also, $m^2 < n^2 \implies m^2+a^2 < n^2+a^2 \implies \sqrt{m^2+a^2} < \sqrt{n^2+a^2}$.
    *   Assume for contradiction that $c+d=0$. Using the expressions for $c$ and $d$:
        $(m-\sqrt{m^2+a^2}) + (n+\sqrt{n^2+a^2}) = 0 \implies m+n = \sqrt{m^2+a^2} - \sqrt{n^2+a^2}$.
    *   The left side, $m+n$, is positive. The right side, $\sqrt{m^2+a^2} - \sqrt{n^2+a^2}$, is negative. This is a contradiction.
    *   Thus, our assumption is false, and $c+d \neq 0$.

*   **Lemma 2:** $a^2+cd < 0$.
    *   **Proof:** The condition $m<n$ implies $\frac{c^2-a^2}{2c} < \frac{d^2-a^2}{2d}$.
    *   From their definitions, $c<0$ and $d>0$, so $cd<0$. Multiplying the inequality by $2cd$ (which is negative) reverses the inequality sign:
        $d(c^2-a^2) > c(d^2-a^2) \implies c^2d-a^2d > cd^2-a^2c \implies cd(c-d) > -a^2(c-d)$.
    *   Since $c<d$, we have $c-d<0$. Dividing by $c-d$ reverses the inequality sign again:
        $cd < -a^2 \implies a^2+cd < 0$.

**3. Coordinates of P and H**

*   Let $P(x_P, y_P)$ be the circumcenter of $\triangle ACD$ with vertices $A(0,a)$, $C(c,0)$, $D(d,0)$.
*   $P$ lies on the perpendicular bisector of segment $CD$, which is the line $x = \frac{c+d}{2}$. So, $x_P = \frac{c+d}{2}$.
*   The condition $PA^2 = PC^2$ gives $(x_P-0)^2 + (y_P-a)^2 = (x_P-c)^2 + (y_P-0)^2$.
    $x_P^2 + y_P^2 - 2ay_P + a^2 = x_P^2 - 2cx_P + c^2 + y_P^2 \implies -2ay_P + a^2 = -2cx_P + c^2$.
    Substituting $x_P = \frac{c+d}{2}$: $-2ay_P + a^2 = -c(c+d) + c^2 = -cd \implies y_P = \frac{a^2+cd}{2a}$.
    So, $P\left(\frac{c+d}{2}, \frac{a^2+cd}{2a}\right)$.

*   Let $H(x_H, y_H)$ be the orthocenter of $\triangle PMN$ with vertices $P(x_P, y_P)$, $M(m,0)$, $N(n,0)$.
*   The altitude from $P$ to $MN$ (on the $x$-axis) is the line $x=x_P$. Thus, $x_H = x_P = \frac{c+d}{2}$.
*   The altitude from $M$ is perpendicular to $PN$. The slope of this altitude is $-\frac{x_P-n}{y_P}$. The line is $y-0 = -\frac{x_P-n}{y_P}(x-m)$.
*   $H$ lies on this line, so $y_H = -\frac{x_P-n}{y_P}(x_H-m) = -\frac{(x_P-m)(x_P-n)}{y_P}$.
*   We express the numerator in terms of $a,c,d$:
    $x_P-m = \frac{c+d}{2} - \frac{c^2-a^2}{2c} = \frac{c(c+d)-(c^2-a^2)}{2c} = \frac{cd+a^2}{2c}$.
    $x_P-n = \frac{c+d}{2} - \frac{d^2-a^2}{2d} = \frac{d(c+d)-(d^2-a^2)}{2d} = \frac{cd+a^2}{2d}$.
*   Substituting these into the expression for $y_H$:
    $y_H = -\frac{1}{y_P} \left(\frac{cd+a^2}{2c}\right) \left(\frac{cd+a^2}{2d}\right) = -\frac{(cd+a^2)^2}{4cd y_P}$.
    Using $y_P = \frac{a^2+cd}{2a}$ and $a^2+cd \neq 0$ (from Lemma 2), we get:
    $y_H = -\frac{(cd+a^2)^2}{4cd} \frac{2a}{a^2+cd} = -\frac{a(a^2+cd)}{2cd}$.
*   So, the coordinates of the orthocenter are $H\left(\frac{c+d}{2}, -\frac{a(a^2+cd)}{2cd}\right)$.

**4. The Line AP and its Intersections**

*   Since $a \neq 0$ and $c+d \neq 0$ (Lemma 1), the slope of line $AP$, denoted $k_{AP}$, is well-defined:
    $k_{AP} = \frac{y_P-a}{x_P-0} = \frac{\frac{a^2+cd}{2a}-a}{\frac{c+d}{2}} = \frac{a^2+cd-2a^2}{a(c+d)} = \frac{cd-a^2}{a(c+d)}$.
*   The line $AP$ has equation $y=k_{AP}x+a$. $E(x_E, y_E)$ and $F(x_F, y_F)$ are the other intersection points of this line with $\Omega$ and $\Gamma$ respectively.
*   To find $E$, substitute $y=k_{AP}x+a$ into the equation of $\Omega$, $x^2-2mx+y^2=a^2$:
    $x^2-2mx+(k_{AP}x+a)^2=a^2 \implies x((1+k_{AP}^2)x - 2m + 2ak_{AP})=0$.
    The roots are $x=0$ (for point $A$) and $x_E = \frac{2(m-ak_{AP})}{1+k_{AP}^2}$.
*   Similarly, for $F$ on $\Gamma$ ($x^2-2nx+y^2=a^2$): $x_F = \frac{2(n-ak_{AP})}{1+k_{AP}^2}$.

**5. Auxiliary Algebraic Identities**

We derive identities that will simplify subsequent calculations.
*   **Identity 1:** $m-ak_{AP} = \frac{c^2-a^2}{2c} - a\left(\frac{cd-a^2}{a(c+d)}\right) = \frac{(c^2-a^2)(c+d) - 2c(cd-a^2)}{2c(c+d)} = \frac{(c^2+a^2)(c-d)}{2c(c+d)}$.
*   **Identity 2:** $n-ak_{AP} = \frac{d^2-a^2}{2d} - \frac{cd-a^2}{c+d} = \frac{(d^2-a^2)(c+d) - 2d(cd-a^2)}{2d(c+d)} = \frac{(d^2+a^2)(d-c)}{2d(c+d)}$.
*   **Identity 3:** $1+k_{AP}^2 = 1+\left(\frac{cd-a^2}{a(c+d)}\right)^2 = \frac{a^2(c+d)^2+(cd-a^2)^2}{a^2(c+d)^2} = \frac{(c^2+a^2)(d^2+a^2)}{a^2(c+d)^2}$.
*   **Identity 4:** $k_{AP}+\frac{2a}{x_E} = \frac{cd-a^2}{a(c+d)} + \frac{a(1+k_{AP}^2)}{m-ak_{AP}} = \frac{cd-a^2}{a(c+d)} + a \frac{(c^2+a^2)(d^2+a^2)}{a^2(c+d)^2} \left/ \frac{(c^2+a^2)(c-d)}{2c(c+d)} \right. = \frac{cd-a^2}{a(c+d)} + \frac{2c(d^2+a^2)}{a(c+d)(c-d)} = \frac{(cd-a^2)(c-d)+2c(d^2+a^2)}{a(c+d)(c-d)} = \frac{cd(c+d)+a^2(c+d)}{a(c+d)(c-d)} = \frac{cd+a^2}{a(c-d)}$.

**6. The Circumcircle of $\triangle BEF$**

Let $K(x_K, y_K)$ be the circumcenter of $\triangle BEF$. $K$ lies on the perpendicular bisectors of $BE$ and $BF$.
The perpendicular bisector of $BE$ is given by $KB^2=KE^2$, which simplifies to $2x_Kx_E+2y_K(y_E+a) = x_E^2+y_E^2-a^2+2a^2 = 2mx_E+2a^2$. Using $y_E=k_{AP}x_E+a$, we get $x_Kx_E+y_K(k_{AP}x_E+2a) = mx_E+a^2$. This is incorrect.
Let's re-derive. $KB^2=KE^2 \implies x_K^2+(y_K+a)^2 = (x_K-x_E)^2+(y_K-y_E)^2$.
$x_K^2+y_K^2+2ay_K+a^2 = x_K^2-2x_Kx_E+x_E^2+y_K^2-2y_Ky_E+y_E^2$.
$2ay_K+a^2 = -2x_Kx_E+x_E^2-2y_Ky_E+y_E^2$.
Since $E$ is on $\Omega$, $x_E^2+y_E^2=2mx_E+a^2$.
$2ay_K+a^2 = -2x_Kx_E+2mx_E+a^2-2y_Ky_E$.
$2ay_K = -2x_Kx_E+2mx_E-2y_Ky_E$.
$ay_K = -x_Kx_E+mx_E-y_Ky_E$.
$ay_K = -x_Kx_E+mx_E-y_K(k_{AP}x_E+a) \implies 2ay_K = -x_Kx_E+mx_E-y_Kk_{AP}x_E$.
$x_E(x_K+y_Kk_{AP}) + 2ay_K = mx_E$. Dividing by $x_E \neq 0$:
(1) $x_K + y_K(k_{AP}+\frac{2a}{x_E}) = m$.
A similar derivation for the perpendicular bisector of $BF$ (using point $F$ on $\Gamma$) yields:
(2) $x_K + y_K(k_{AP}+\frac{2a}{x_F}) = n$.
Subtracting (1) from (2): $y_K(\frac{2a}{x_F}-\frac{2a}{x_E}) = n-m \implies 2ay_K\frac{x_E-x_F}{x_Ex_F} = n-m$.
Using $x_E-x_F = \frac{2(m-n)}{1+k_{AP}^2}$, we get $2ay_K \frac{2(m-n)/(1+k_{AP}^2)}{x_Ex_F} = n-m$. Since $m \neq n$, we find $y_K = -\frac{x_Ex_F(1+k_{AP}^2)}{4a} = -\frac{(m-ak_{AP})(n-ak_{AP})}{a(1+k_{AP}^2)}$.
Using Identities 1, 2, 3:
$y_K = -\frac{1}{a} \left(\frac{(c^2+a^2)(c-d)}{2c(c+d)}\right) \left(\frac{(d^2+a^2)(d-c)}{2d(c+d)}\right) \left/ \frac{(c^2+a^2)(d^2+a^2)}{a^2(c+d)^2} \right. = \frac{a(c-d)^2}{4cd}$.
From (1) and Identity 4: $x_K = m - y_K(k_{AP}+\frac{2a}{x_E}) = m - y_K\frac{cd+a^2}{a(c-d)}$.
$x_K = \frac{c^2-a^2}{2c} - \frac{a(c-d)^2}{4cd} \frac{cd+a^2}{a(c-d)} = \frac{c^2-a^2}{2c} - \frac{(c-d)(cd+a^2)}{4cd} = \frac{2d(c^2-a^2) - (c-d)(cd+a^2)}{4cd} = \frac{cd(c+d)-a^2(c+d)}{4cd} = \frac{(cd-a^2)(c+d)}{4cd}$.
The radius squared of the circumcircle of $\triangle BEF$ is $R_K^2 = KB^2 = x_K^2+(y_K+a)^2$.
$y_K+a = \frac{a(c-d)^2}{4cd}+a = \frac{a(c+d)^2}{4cd}$.
$R_K^2 = \left(\frac{(cd-a^2)(c+d)}{4cd}\right)^2 + \left(\frac{a(c+d)^2}{4cd}\right)^2 = \frac{(c+d)^2}{(4cd)^2} \left((cd-a^2)^2 + a^2(c+d)^2\right) = \frac{(c+d)^2(c^2+a^2)(d^2+a^2)}{16c^2d^2}$.

**7. The Tangency Proof**

The line $\ell_H$ passes through $H(x_H, y_H)$ and is parallel to $AP$. Its equation is $y-y_H = k_{AP}(x-x_H)$, which can be written as $k_{AP}x - y - (k_{AP}x_H - y_H) = 0$.
This line is tangent to the circumcircle of $\triangle BEF$ (center $K$, radius $R_K$) if the square of the distance from $K$ to $\ell_H$ is $R_K^2$. The squared distance is $\frac{(k_{AP}x_K - y_K - (k_{AP}x_H - y_H))^2}{k_{AP}^2+1}$.
The condition for tangency is $(k_{AP}(x_K-x_H)-(y_K-y_H))^2 = R_K^2(1+k_{AP}^2)$.
Let's compute the terms on the left side (LHS).
$x_K-x_H = \frac{(cd-a^2)(c+d)}{4cd} - \frac{c+d}{2} = \frac{(c+d)(cd-a^2-2cd)}{4cd} = -\frac{(c+d)(cd+a^2)}{4cd}$.
$y_K-y_H = \frac{a(c-d)^2}{4cd} - \left(-\frac{a(a^2+cd)}{2cd}\right) = \frac{a(c^2-2cd+d^2)+2a(a^2+cd)}{4cd} = \frac{a(c^2+d^2+2a^2)}{4cd}$.
$k_{AP}(x_K-x_H) = \frac{cd-a^2}{a(c+d)} \left(-\frac{(c+d)(cd+a^2)}{4cd}\right) = -\frac{(cd-a^2)(cd+a^2)}{4acd} = -\frac{c^2d^2-a^4}{4acd}$.
$k_{AP}(x_K-x_H)-(y_K-y_H) = -\frac{c^2d^2-a^4}{4acd} - \frac{a^2(c^2+d^2+2a^2)}{4acd} = -\frac{c^2d^2-a^4+a^2c^2+a^2d^2+2a^4}{4acd} = -\frac{(c^2+a^2)(d^2+a^2)}{4acd}$.
LHS = $\left(-\frac{(c^2+a^2)(d^2+a^2)}{4acd}\right)^2 = \frac{(c^2+a^2)^2(d^2+a^2)^2}{16a^2c^2d^2}$.
Now we compute the right side (RHS) using $R_K^2$ and Identity 3:
RHS = $R_K^2(1+k_{AP}^2) = \left(\frac{(c+d)^2(c^2+a^2)(d^2+a^2)}{16c^2d^2}\right) \left(\frac{(c^2+a^2)(d^2+a^2)}{a^2(c+d)^2}\right) = \frac{(c^2+a^2)^2(d^2+a^2)^2}{16a^2c^2d^2}$.
Since LHS = RHS, the tangency condition is satisfied. This completes the proof.

\end{document}